\documentclass{article} % For LaTeX2e
\usepackage{iclr2027_conference,times,lila}
\preprint

\usepackage{amsmath,amsfonts,bm}

\def\eqref#1{equation~\ref{#1}}
\def\1{\bm{1}}

\DeclareMathAlphabet{\mathsfit}{\encodingdefault}{\sfdefault}{m}{sl}
\SetMathAlphabet{\mathsfit}{bold}{\encodingdefault}{\sfdefault}{bx}{n}

\usepackage{graphicx}
\usepackage{subcaption}
\usepackage{algorithm}
\usepackage{amsmath}
\usepackage{amssymb}
\usepackage{algorithmicx}
\usepackage{algpseudocode}
\usepackage{dsfont}
\usepackage{hyperref}
\usepackage{cleveref}
\usepackage{url}
\usepackage{tabularx}
\usepackage{booktabs}
\usepackage{enumitem}
\usepackage{makecell}
\usepackage{multirow}
\usepackage[most]{tcolorbox}
\usepackage{wrapfig}
\usepackage{listings}

\definecolor{brandRef}{HTML}{A0AAAE}
\definecolor{brandDGRPO}{HTML}{D29937}
\definecolor{brandDARLING}{HTML}{37997A}
\definecolor{brandCP}{HTML}{12588C}
\newcommand{\methodswatch}[1]{\textcolor{#1}{\rule[0ex]{0.75em}{0.75em}}\hspace{0.4em}}

\lstdefinestyle{prompttemplate}{
    basicstyle=\footnotesize\ttfamily,
    breaklines=true,
    columns=flexible,
    keepspaces=true,
    showstringspaces=false,
    literate=
        {—}{{---}}1
        {–}{{--}}1
        {‘}{{`}}1
        {’}{{'}}1
        {“}{{"}}1
        {”}{{"}}1
        {…}{{...}}1
        {é}{{\'e}}1
}

\newcommand{\cp}{\textsc{CP}}
\newcommand{\stat}[2]{#1\,{\scriptsize$\pm\,#2$}}

\title{Creative Priming: Decoupling Process from Product to Escape the Creative Paradox}

\author{Alon Albalak, Sebastian Gabriel, David B. D'Ambrosio, Joel Lehman, Kenneth O. Stanley\\
\texttt{aalbalak@lila.ai}\\
Lila Sciences
}

\begin{document}

\maketitle

\begin{abstract}
% Creativity is important for LLMs, and hasn't been properly addressed
Many of the most consequential advances promised by AI, from curing diseases and inventing new technologies to formulating new scientific theories, are fundamentally creative acts.
% Creative paradox
Yet creativity is uniquely difficult to learn because it faces a \emph{creative paradox}: learning to repeat past examples of creative success makes those specific examples more likely to recur, inhibiting the very creativity desired.
% How we view creativity training
This work argues that creativity is better modeled as an explicit process rather than as a property of individual artifacts, and proposes to learn the underlying generative principles behind the creative process.
% This is how we propose to do creativity training (introduce method, and high-level idea of how it works)
In particular, to escape the paradox, we introduce \emph{creative priming} (\cp{}), wherein a trained priming model generates a prompt-specific primer that assists a response model in being creative, explicitly decoupling the creative process from response generation.
The priming model learns through GRPO, rewarding primers that lead to responses that are novel, high-quality, and diverse, thereby reinforcing the process that yields creative responses without reinforcing any specific response.
% Headline results (vs. baselines, and transfer to API models)
Evaluating \cp{} on NoveltyBench and comparing against response models directly trained for creativity, we find that priming produces a 39--200\% relative increase in the number of responses that are simultaneously novel, high-quality, and semantically distinct.
Additionally, although the primer is only trained to improve creativity for a small open-source model (Qwen3-8B), the creative principles it learns transfer to frontier API models, increasing the number of semantically distinct responses they generate by 4.5--10$\times$ and diversity by 2.5--4.6$\times$, while maintaining or improving quality.
% Closing sentence
\cp{} thereby shows that decoupling the creative process from the responses themselves offers a concrete escape from the creative paradox.
\end{abstract}

\section{Introduction}

% PARA: Motivate why we should work on creativity
Even as benchmark after benchmark and test after test have fallen to LLMs in recent years~\citep{rein2023gpqagraduatelevelgoogleproofqa, OSWorld, chan2025mlebenchevaluatingmachinelearning, yue2025mmmuprorobustmultidisciplinemultimodal, patwardhan2025gdpvalevaluatingaimodel, arora2025healthbenchevaluatinglargelanguage, deng2025swebenchproaiagents, phan2025lastexam, merrill2026terminalbenchbenchmarkingagentshard, sajadieh2026artificialintelligenceindexreport}, creativity remains an unconquered frontier~\citep{ICLR2024_02dec887, NEURIPS2025_348f2e87, mizrahi-etal-2026-cooking, jiang2025artificialhivemindopenendedhomogeneity, hou2026creativityprism, bao2026contemporaryailacksimagination}. Perhaps that is why many of the most compelling motivational framings for the future of AI point to creative endeavors: we will \emph{discover} new cures, \emph{invent} new technologies, \emph{find} solutions to our most pressing problems, and \emph{imagine} new art forms~\citep{amodei2024machines, altman2024intelligence}.
All of these looming triumphs involve fundamentally creative acts.

% PARA: Give the reader the sense that models don't currently have creativity. Building up to the idea that we, as a community, are missing something
As AI continues to extend its dominance across an increasingly wide set of domains, the question of when this creative renaissance will begin in earnest becomes increasingly central.
The focus on creativity is already reflected in scientific domains, with efforts for models to generate hypotheses, formulate research plans, anticipate scientific insights, and act as autonomous researchers~\citep{liu2026hypobenchsystematicprincipledbenchmarking, goel2025training, heyueya2026giantsgenerativeinsightanticipation, liu2026mindflow, lupidi2026airsbenchsuitetasksfrontier, nathani2025mlgymnewframeworkbenchmark}.
To be sure, there are \emph{some} domain-specific signs of AI creativity, yet perhaps there should be more evidence after billions of dollars spent and seemingly superhuman capabilities in mathematics~\citep{imo_2026,openai2026tenadvances,OAI_math}.
Where are the cured diseases, the revolutionary new technologies, the solutions to our energy challenges, the new genres that change the world?

% PARA: (some background on creativity) Talk a bit about what creativity is, using previous works from both AI and humans
Margaret Boden's influential conception of creativity imagines a continuum from weaker to stronger forms, from the more prosaic combinatorial and exploratory creativities to the more profound transformational version~\citep{boden2004creative}.
One potential hope is that models are slowly climbing this ladder as they become more generally capable, and are just now bumping against the lower rungs of transformational creativity.
However, studies of scientific ideation find that model-generated ideas remain less original than human ideas and tend to stay close to familiar scientific directions~\citep{chen2026measuringgaphumanllm, bao2026contemporaryailacksimagination, antoniades2026heuresissearchstrategiesautonomous}.
These findings give reason to believe that creativity is unique within the intelligence landscape and cannot be learned through conventional means like other aspects of intelligence.

% PARA: Why is creativity so difficult to teach?
Consider that creativity is notoriously challenging to teach even to humans~\citep{10.7551/mitpress/2205.003.0008, KAUFMAN2010380, baird2012inspired}.
A key obstacle is that, unlike other skills, creativity is difficult to learn from examples. While any successful past creative output could in principle serve as a training example for an aspiring student of creativity, most approaches to learning would make similar outputs \emph{more likely} in the future~\citep{jansson1991design, smith1993constraining}. However, for creativity, reproducing past outputs is \emph{exactly opposite the goal}; rather, the goal is to learn generative principles for producing future creative outputs.

% PARA: Introduce the creative paradox
In this way, learning to be creative faces a vexing \textbf{creative paradox}: emulating examples of past creative success (even if they were your own successes) can actually prevent you from reaching your full creative potential. The drive to escape the past is not ideally served by reinforcing past behavior.
Part of the problem is the fleeting nature of novelty; as soon as a novel artifact appears in the world, even if it was your own original idea, it is no longer novel. 
After all, anything in the distribution of past events is by definition not novel, which is why learning to generate past instances of novelty runs into the creative paradox.

% PARA: Creativity as a process
Therefore, rather than asking \emph{what did I do that was creative},
what we really want to ask 
is, \emph{what lessons can I abstract from the successful generation of creativity that can be reapplied to generate further creativity?}
Answering this question is the path to circumventing the creative paradox. The problem is that training methods do not automatically \emph{decouple} the abstraction of general lessons from the reinforcement of specific past effective actions.

% PARA: Creative primer
To address the creative paradox, this work proposes \emph{creative priming} (\textbf{\cp{}}), a process in which a creative priming model generates a prompt-specific creative primer (i.e.\ a short text) that is passed as additional context to a frozen response generation model. 
The priming model learns through GRPO~\citep{shao2024deepseekmathpushinglimitsmathematical}, rewarding primers that lead to responses that are novel, high-quality, and diverse.
Thus the system can reinforce the priming \emph{process} that yields creative outputs without reinforcing any specific creative output, circumventing the creative paradox.

% PARA: results
We evaluate \cp{} on NoveltyBench~\citep{zhang2025noveltybenchevaluatinglanguagemodels} and compare against two methods that train models to directly respond creatively. Each method is evaluated by generating 64 responses per prompt and counting the number of responses that are simultaneously novel, high-quality, and semantically distinct (a metric that we argue a creative system should score highly on), finding that \cp{} leads to an average of 22.8 such responses per prompt, compared to 16.4 and 7.6 from direct generation methods (1.39--3.0$\times$ more).
Furthermore, scaling the number of on-policy training samples for a direct generation baseline fails to match \cp{} on measures of system creativity and can actually reduce performance, supporting the creative paradox's prediction that directly reinforcing successful responses is not expected to improve creativity indefinitely.
Because the primer learns general principles for creativity, it also transfers to models it was never trained with: across four API models, priming increases the number of distinct responses by 4.5--10$\times$ and diversity by 2.5--4.6$\times$, while modestly improving quality (0.5--4.6\%).

% PARA: Contributions statement
These results show that creative priming not only offers a way out of the creative paradox but can also potentially combine effectively with other techniques and even with models that were not part of the original creative priming training stack. More generally, we hope the practical gains from creative priming can help to spark deeper introspection in the field about what makes training explicitly for creativity unique compared to other kinds of learning.

\section{Background}
Before discussing how to improve creativity in language models, we first review what creativity is and how it can be measured. We then move on to a brief discussion of prior works on increasing creativity and why they are still vulnerable to the creative paradox.

\subsection{Defining and Measuring Creativity}
\label{sec:creativity}

% PARA: adopt a converged novelty+value definition of creativity; "new" needs a reference point, which motivates scoping the paper to P-creativity (two reasons)
Definitions of human creativity often center on similar themes. For example:
\begin{itemize}
    \item ideas that are simultaneously \emph{original} and \emph{effective}~\citep{Runco2012TheSD}, or
    \item the ability to produce ideas or artifacts that are new, surprising, and valuable~\citep{boden_creativity_nutshell}.
\end{itemize}
Most definitions agree that an idea must be novel and it must be good, and neither property alone is sufficient, to be called creative.
\emph{Novelty}, however, is nebulous without a reference point: novel to whom?
Therefore, \citet{boden_creativity_nutshell} also draws an essential distinction between two types of creativity: artifacts and ideas that are new to the individual who produced them (\emph{psychological} creativity, or P-creativity) and those that are new in the entire history of human thought (\emph{historical} creativity, or H-creativity).

Throughout this work, we study the P-creativity of language models for two reasons.
First, H-creativity is nearly impossible to validate, except on a narrow set of known problems with unknown solutions that can be verified (e.g., open conjectures in math), whereas P-creativity is empirically tractable, requiring only that we have a record of a model's previous behaviors.
Second, for language models, the distinction between the two is collapsing: as frontier language models absorb ever more information, the gap between what is new to the model and what is new to humanity shrinks.
For these reasons, moving forward, when we refer to creativity we are referring to P-creativity.

% PARA: map novelty->deviation and effectiveness+value->quality; define the individual creative artifact (deviating AND high-quality)
Mapping the requirements of~\citet{Runco2012TheSD} (originality and effectiveness) onto language models, originality becomes \emph{deviation}, defined as the distance of a given response from some reference set of responses, and effectiveness becomes \emph{quality}.
Thus, the \textbf{creativity of an individual response} can be defined as when the response \emph{simultaneously deviates from the reference set and is high-quality}. A response that departs from the reference set but is incoherent is not creative, and neither is a polished response that merely restates a common idea.

\begin{figure}[t!]
    \centering
    \includegraphics[width=\linewidth]{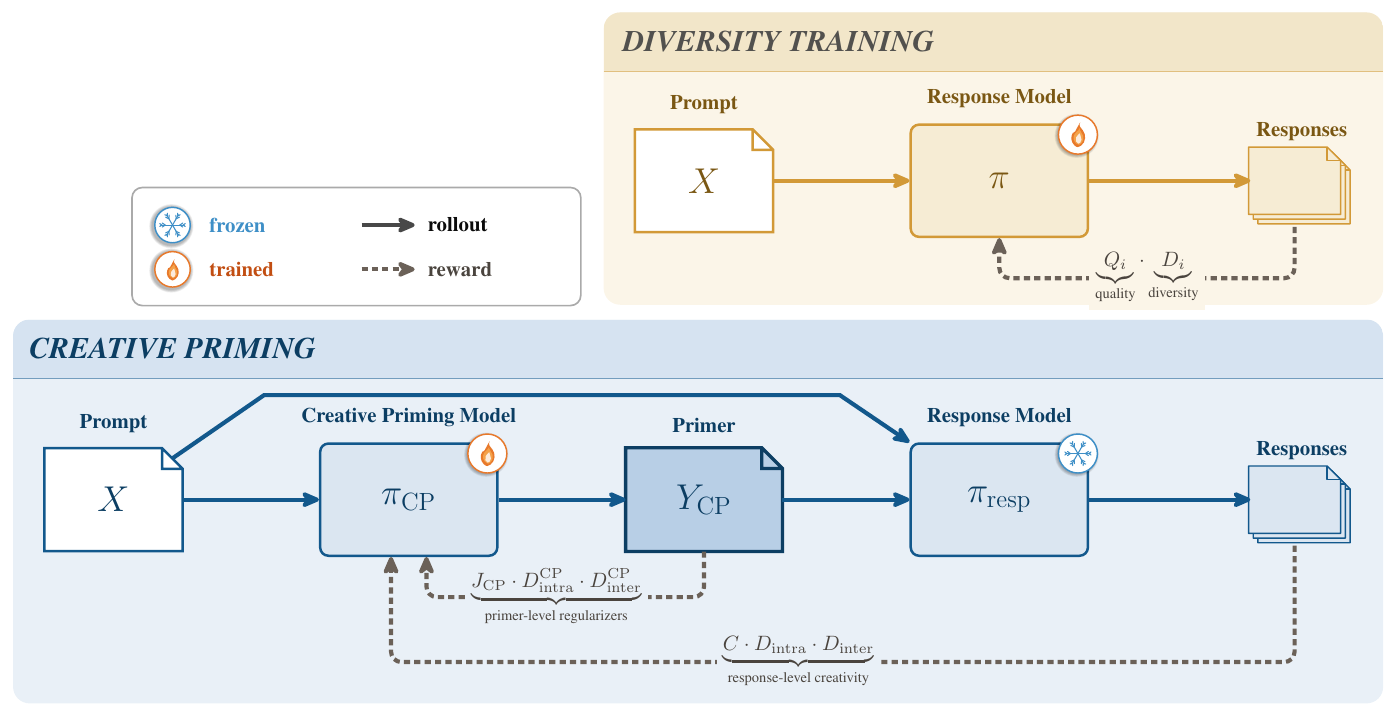}
    \caption{\textbf{The creative priming training procedure.} Diversity training (top right; Section~\ref{sec:prelim_baselines}) optimizes the response model $\pi$ directly, sampling responses from the prompt $X$ and rewarding each one for being simultaneously high-quality and diverse ($r_i = Q_i \cdot D_i$). Creative priming (bottom; Section~\ref{sec:cp}) instead trains a separate priming model $\pi_{\mathrm{CP}}$ over a \emph{frozen} response model $\pi_{\mathrm{resp}}$: $\pi_{\mathrm{CP}}$ produces a \emph{primer} $Y_{\mathrm{CP}}$ that is combined with $X$ and passed to $\pi_{\mathrm{resp}}$ to generate a response. The reward flows back into $\pi_{\mathrm{CP}}$ alone through two paths: response-level creativity signals ($C \cdot D_{\mathrm{intra}} \cdot D_{\mathrm{inter}}$) from the responses and primer-level regularizers ($J_{\mathrm{CP}} \cdot D^{\mathrm{CP}}_{\mathrm{intra}} \cdot D^{\mathrm{CP}}_{\mathrm{inter}}$) from the primer.
    The priming process decouples learning the creative process from reinforcing any particular responses, avoiding the creative paradox.
    }
    \label{fig:overview_pipeline}
\end{figure}

% PARA: lift from artifact to system ; a creative system reliably produces creative artifacts that are also diverse; motivates filter-then-measure-spread and a joint threshold sweep
The \textbf{creativity of a system} can then be defined as \emph{when the system reliably produces responses that are individually creative and are also varied amongst themselves}.
This definition offers a direct recipe for measuring system creativity (which we will later formalize): first, remove responses that are not individually creative (i.e.\ those below the deviation and quality thresholds), and then measure the spread of the set that survives, so that spread is credited only among responses that are already regarded as creative.

\subsection{Reinforcement Learning for Diversity}
\label{sec:preliminaries}

% PARA: homogeneity is the visible symptom; classical instruments and diversity methods target diversity alone, so optimizing it Goodharts creativity
The most visible symptom of the limited creativity in modern language models is their homogeneity: across many models and even across entire model families, outputs often collapse on the same handful of ideas with only minor variations in phrasing~\citep{jiang2025artificialhivemindopenendedhomogeneity, zhang2025noveltybenchevaluatinglanguagemodels}.
This homogeneity has motivated a wave of post-training methods to address the problem.

A number of preference-based methods formulate training data pairs to favor responses that are rare or marginally diverse while retaining quality~\citep{chung2025modifying, lanchantin2025diverse, samuel2026recovering}, while Creative Preference Optimization incorporates novelty, diversity, surprise, and quality directly into preference optimization~\citep{ismayilzada2025creative}.
Other methods instead construct rewards from collections of generated responses, using learned semantic partitions in DARLING~\citep{li2025jointlyreinforcingdiversityquality}, determinantal point processes in Diversity Quality Optimization~\citep{chen2026posttraining}, or specialization across a vector-valued reward space in Vector Policy Optimization~\citep{bahlousboldi2026vectorpolicyoptimizationtraining}.
CreativeInstruct takes a complementary instruction-tuning approach, teaching the response model when to enter creativity-biased spans~\citep{sahu2026creativeinstruct}.
In this work, two families of diversity-enhancing methods that differ only in how they measure diversity serve as baselines. Both methods train the response model to directly respond to prompts (top in Figure~\ref{fig:overview_pipeline}) with GRPO~\citep{shao2024deepseekmathpushinglimitsmathematical}.

\subsubsection{Group-Relative Policy Optimization}
\label{sec:prelim_grpo}

% PARA: GRPO backbone: motivate RL, give the group-relative advantage every method reuses (differing only in the reward), and fix the shared phi/d notation
In GRPO, for each prompt $X$, a group of $G$ responses $\{Y_1, \dots, Y_G\}$ is sampled from the model, $Y\sim \pi(\cdot \mid X)$, and each response is scored with a reward $r_i$, forming a group-relative advantage,
\begin{equation}
\label{eq:bg_grpo_adv}
A_i = \frac{r_i - \mathrm{mean}(\{r_1, \dots, r_G\})}{\mathrm{std}(\{r_1, \dots, r_G\})}.
\end{equation}
In this work, all models are trained with GRPO as the backbone and vary only in how the reward $r_i$ is defined.
Throughout, let $\phi(\cdot)$ be an embedding model and let $\mathrm{d}(a,b) = \min\!\big(1,\, 1 - \cos\theta_{ab}\big) \in [0,1]$ denote the cosine distance between two responses, where $\theta_{ab} = \angle(\phi(a), \phi(b))$ is the angle between their embeddings.

\subsubsection{Diversity-Enhancing Baselines}
\label{sec:prelim_baselines}

% PARA: frame the two baseline families that differ only in how they measure diversity
We consider two families of diversity-enhancing methods that differ only in how they measure diversity. Both methods train the response model to directly respond to prompts (top in Figure~\ref{fig:overview_pipeline}).

% PARA: family 1, Diversified GRPO (DGRPO): train the response model directly with a multiplicative reward, embedding diversity times a logistic-gated [0,1] quality; a PPO-style recast of DDPO/DORPO
\paragraph{Diversified GRPO (DGRPO).}
The first method, \emph{Diversified GRPO} (DGRPO), inspired by Diversified DPO and Diversified ORPO~\citep{chung2025modifying}, recasts their objective as a PPO-style algorithm, trading their labeled preference data (expensive and limiting to collect) for on-policy sampling. DGRPO rewards each response by the product of its quality $Q_i$ and its intra-group diversity $D_i$,
\begin{equation}
\label{eq:bg_rm_reward}
r_i = Q_i \cdot D_i,
\qquad
D_i = \frac{1}{G-1} \sum_{j \neq i} \mathrm{d}(Y_i, Y_j),
\qquad
Q_i = \sigma\!\big((r_{\mathrm{RM}}(X, Y_i) - T)/\tau\big).
\end{equation}
The quality term is a reward-model score passed through a logistic gate with threshold $T$ and temperature $\tau$, while diversity is the response's mean cosine distance from the other responses in its sampled group.
Exact definitions and the complete training loop are in Appendix~\ref{sec:rm_diversity}.

% PARA: family 2 (DARLING): same multiplicative shape, partition-based diversity instead of embedding distance, raw RM quality, no deviation; reimplemented under its original recipe
\paragraph{Partition-based diversity (DARLING).}
The second method, Diversity-Aware Reinforcement Learning (DARLING)~\citep{li2025jointlyreinforcingdiversityquality}, replaces $Q_{i}$ with the raw reward-model quality score, and $D_i$ with a partition-based signal.
For the parition-based diversity, a learned equivalence classifier groups the $G$ responses into semantic partitions, and $Y_i$ is rewarded in proportion to how many of its peers are distinct from it, $D_i = (G - |\mathrm{partition}(i)|)\,/\,(G-1)$, so that responses in rare partitions score highest. The potential benefit of DARLING over DGRPO lies in the learned equivalence classifier; it is trained to measure diversity beyond surface-level lexical variations.

\section{Approach: Creative Priming}
\label{sec:cp}

Rather than directly optimizing the response model to be creative, \emph{creative priming} (\textbf{\cp{}}) treats creativity as a process and decouples abstract thinking from response generation, training a separate policy that learns to guide a frozen response model.

% PARA: define a creative primer and formalize the composed two-stage sampling pipeline over a frozen response model (mechanics only); state the train-CP/freeze-resp asymmetry
A \emph{creative primer} is a short text $Y_{\mathrm{CP}}$ produced by the creative priming policy $\pi_{\mathrm{CP}}$ that is conditioned on the original prompt $X$. The primer combines with $X$ through a prompt template $\mathcal{T}(\cdot,\cdot)$ and feeds into the response model $\pi_{\mathrm{resp}}$ (Figure~\ref{fig:overview_pipeline} and Eq~\ref{eq:cp_pipeline}; an example of $\mathcal{T}$ is later provided in Figure~\ref{fig:response_template} of the Appendix).
In creative priming, only $\pi_{\mathrm{CP}}$ is trained while the response model $\pi_{\mathrm{resp}}$ is frozen.
A response is thus generated in a two-stage process,
\begin{equation}
\label{eq:cp_pipeline}
Y_{\mathrm{CP}} \sim \pi_{\mathrm{CP}}(\cdot \mid X),
\qquad
Y_{\mathrm{resp}} \sim \pi_{\mathrm{resp}}\!\big(\cdot \mid \mathcal{T}(X, Y_{\mathrm{CP}})\big),
\end{equation}
so that $\pi_{\mathrm{CP}}$ steers the response only through the primer $Y_{\mathrm{CP}}$.

\subsection{Algorithm}
\label{sec:cp_algorithm}

% PARA: formalize the two-level primer/response sampling used at training time, so the reward can score each primer by the responses it induces
Training applies Equation~\ref{eq:cp_pipeline} as a 2-step pipeline: for each prompt $X^{j}$ we sample a group of $G_{\mathrm{CP}}$ primers, and for each primer $Y_{\mathrm{CP}}^{j,i}$ we sample a group of $G_{\mathrm{resp}}$ responses,
\begin{equation}
\label{eq:cp_sampling}
\{Y_{\mathrm{CP}}^{j,i}\}_{i=1}^{G_{\mathrm{CP}}} \sim \pi_{\mathrm{CP}}(\cdot \mid X^{j}),
\qquad
\{Y_{\mathrm{resp}}^{j,i,k}\}_{k=1}^{G_{\mathrm{resp}}} \sim \pi_{\mathrm{resp}}\!\big(\cdot \mid \mathcal{T}(X^{j}, Y_{\mathrm{CP}}^{j,i})\big),
\end{equation}
so that the rewards can score each primer $Y_{\mathrm{CP}}^{j,i}$ based on the creativity of the $G_{\mathrm{resp}}$ responses it induces.

% PARA: each primer's scalar reward feeds the group-relative advantage from Preliminaries, optimized with the clipped GRPO objective and KL penalty; full loop deferred to the appendix
Each creative primer $Y_{\mathrm{CP}}^{j,i}$ receives a scalar reward $r_{j,i} = \mathrm{Reward}(\cdot)$, whose group-level advantage across the $G_{\mathrm{CP}}$ primers is calculated as typical for GRPO (Eq.~\ref{eq:bg_grpo_adv}).
We optimize $\pi_{\mathrm{CP}}$ with the resulting clipped GRPO objective and a KL penalty to a reference policy.
The full training loop is given in Algorithm~\ref{alg:cp} in the appendix.

\subsection{Reward Components}
\label{sec:cp_reward_components}

% PARA: present the full reward up front as a roadmap for the subsection: a product of three response-level (creativity) signals and three primer-level (regularizer) signals
The primer reward is a product of signals at two levels: \emph{response-level} signals that measure the creativity of the responses a primer induces, and \emph{primer-level} signals that regularize the primer model itself:
\begin{equation}
\label{eq:reward}
\mathrm{Reward} = \;
\underbrace{C \cdot D_{\mathrm{intra}} \cdot D_{\mathrm{inter}}}_{\text{response-level (creativity)}}
\;\cdot\;
\underbrace{J_{\mathrm{CP}} \cdot D^{\mathrm{CP}}_{\mathrm{intra}} \cdot D^{\mathrm{CP}}_{\mathrm{inter}}}_{\text{primer-level (regularizers)}}
\end{equation}
Rewards are multiplied rather than summed so that a primer scores well only when every signal is simultaneously high.
Every signal occupies the same $[0,1]$ range to normalize the impact of each factor. 

% PARA: fix the per-primer notation (response set R, count N, reference set R_ref) used by every component below, reusing phi/d from Preliminaries
\paragraph{Notation.}
The components discussed in this section reuse the embedding model $\phi(\cdot)$ and cosine distance $\mathrm{d}(a,b)$ from Section~\ref{sec:prelim_grpo}.
For the primer under consideration, its $G_{\mathrm{resp}}$ responses are $\{Y_{\mathrm{resp}}^{j,i,k}\}_{k=1}^{G_{\mathrm{resp}}}$, and let $\mathcal{R}_{\mathrm{ref}} = \{Y_{\mathrm{ref}}^{j,k}\}_{k=1}^{G_{\mathrm{resp}}}$ be the $G_{\mathrm{resp}}$ reference responses for the same prompt.

\subsubsection{Response-level signals}
% PARA: these three signals instantiate Section 2's system-creativity recipe on the induced responses ; individual creativity of each response times the variety of the response set
The three response-level signals measure the creativity of the response set a primer induces, following the definition adopted in Section~\ref{sec:creativity}, through each response's individual creativity and the variety of the set as a whole.

% PARA: define C, the primer-level response-creativity signal (PDQ) ; couple each response's deviation with its gated quality into a per-response score c_k, then average, matching Section 2's deviating-AND-high-quality definition
\paragraph{Response creativity $C$.}
\begin{wrapfigure}[25]{r}{0.5\textwidth}
    \centering
    \includegraphics[width=0.5\textwidth]{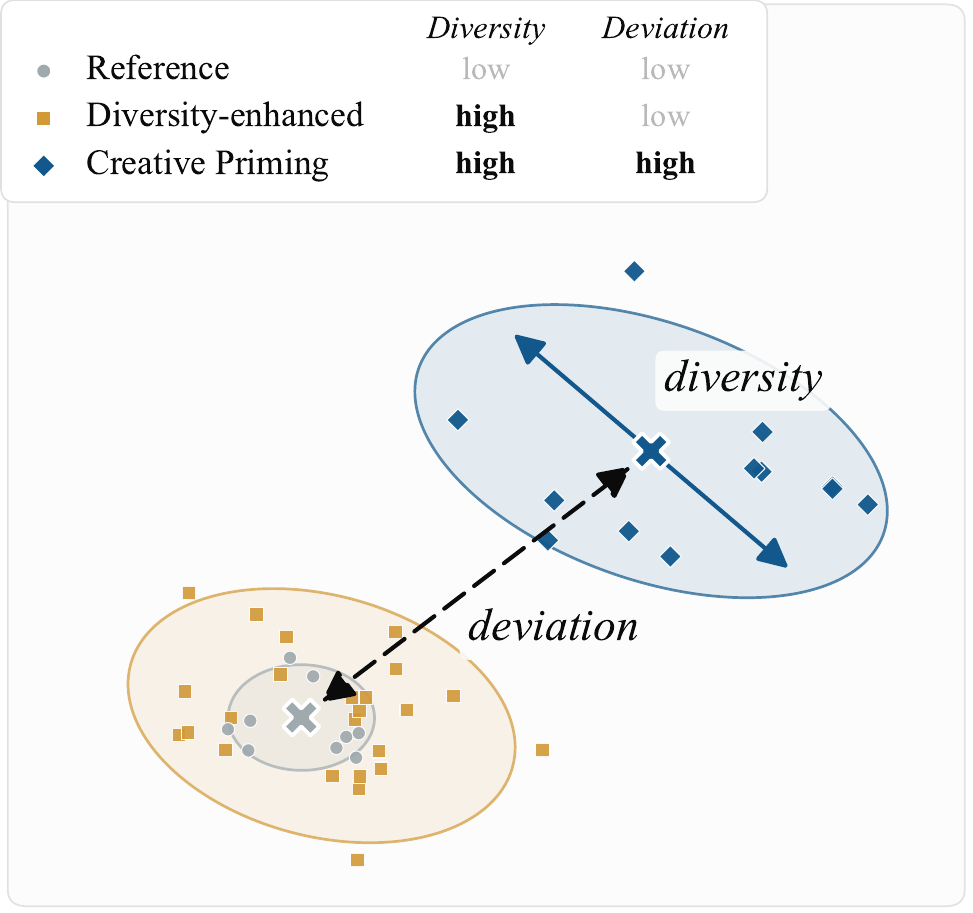}
    \caption{\textbf{Conceptual comparison of diversity (Eq.~\ref{eq:bg_rm_reward}) and deviation (Eq.~\ref{eq:deviation}).} The reference model responses cluster tightly while those of the diversity-enhanced model spread about the same center (increasing diversity), whereas the creative priming responses also deviate from the reference, achieving diversity \emph{and} deviation.}
    \label{fig:overview_scatter}
\end{wrapfigure}
Following the convention in Section~\ref{sec:creativity} that an individual response is creative when it is simultaneously deviating and high-quality, each response's individual-response creativity is scored by multiplying its \emph{deviation} with its \emph{quality} and then averaged over the group.
The deviation of response $k$ is its mean distance to the reference set (Figure~\ref{fig:overview_scatter}),
\begin{equation}
\label{eq:deviation}
\mathrm{Dev}_k = \frac{1}{G_{\mathrm{resp}}} \sum_{Y_{\mathrm{ref}} \in \mathcal{R}_{\mathrm{ref}}}
\mathrm{d}\!\left(Y_{\mathrm{resp}}^{j,i,k}, Y_{\mathrm{ref}}\right).
\end{equation}
Let $q_k = r_{\mathrm{RM}}\!\left(X^{j}, Y_{\mathrm{resp}}^{j,i,k}\right)$ be the raw reward-model score of response $k$, scored against the \emph{original} prompt $X^{j}$ (without the primer). Then the individual-response creativity of response $k$ multiplies its deviation with its normalized quality,
\begin{equation}
\label{eq:ck}
c_k
= \mathrm{Dev}_k \cdot \sigma\!\left(\frac{q_k - T}{\tau}\right),
\end{equation}
where $\sigma$ is the logistic sigmoid, $T$ is the reward-model threshold, and $\tau > 0$ is a temperature controlling the sharpness of the sigmoid.
The response creativity of primer $Y^{j,i}_{CP}$ is then the mean of these per-response scores over the group it induces,
\begin{equation}
\label{eq:pdq}
C = \frac{1}{G_{\mathrm{resp}}} \sum_{k=1}^{G_{\mathrm{resp}}} c_k.
\end{equation}
We also experiment with training the creative priming model without deviation, as discussed in Section~\ref{sec:cp_decoupling}, in which case $C$ reduces to a quality-only signal, given in Appendix~\ref{app:cp_algorithm}.

% PARA: define D_intra as mean pairwise distance within a primer's responses
\paragraph{Intra-group diversity.} $D_{\mathrm{intra}}$, the
mean pairwise cosine distance among the responses from a single creative primer $Y^{j,i}_{CP}$, rewards primers that lead to increased response diversity:
\begin{equation}
\label{eq:dintra}
D_{\mathrm{intra}}
= \frac{1}{\binom{G_{\mathrm{resp}}}{2}} \sum_{k=1}^{G_{\mathrm{resp}}} \sum_{k'=k+1}^{G_{\mathrm{resp}}}
\mathrm{d}\!\left(Y_{\mathrm{resp}}^{j,i,k}, Y_{\mathrm{resp}}^{j,i,k'}\right).
\end{equation}

% PARA: define D_inter as an aggregate over response distances across other primers for the same prompt
\paragraph{Inter-group diversity.} $D_{\mathrm{inter}}$, the
mean distance from $Y^{j,i}_{CP}$'s responses to the responses of other primers to the
same prompt $X^{j}$, rewards primers that lead to responses which deviate from other primer's responses:
\begin{equation}
\label{eq:dinter}
D_{\mathrm{inter}} = \frac{1}{(G_{\mathrm{CP}} - 1)\,G_{\mathrm{resp}}^{2}}
\sum_{k=1}^{G_{\mathrm{resp}}} \sum_{i' \ne i} \sum_{k'=1}^{G_{\mathrm{resp}}}
\mathrm{d}\!\left(Y_{\mathrm{resp}}^{j,i,k}, Y_{\mathrm{resp}}^{j,i',k'}\right).
\end{equation}

\subsubsection{Primer-level signals}
% PARA: these three signals act directly on the primer(s), pressuring the primer model to produce good, prompt-specific, non-collapsing primers
The remaining three signals act on the primers directly and their main purpose is to regularize and diversify the primer model: they keep primers high-quality and specific to each prompt.

% PARA: define J_CP, the LLM-as-a-judge primer-quality score: mean of five positive criteria, gated by direct address, zero on failure; note the appendix penalty-rubric variant
\paragraph{Creative primer judge $J_{\mathrm{CP}}$.}
An LLM-as-a-judge scores the primer $Y_{\mathrm{CP}}^{j,i}$ on a set of five criteria $\mathcal{J} = \{$specificity, thinking strategy, content focus, prompt faithfulness, focus$\}$, each in $[0,100]$. The judge score is the mean of these criteria, but is set to zero if the primer refers to the response model in the third person rather than addressing it directly:
\begin{equation}
\label{eq:judge}
J_{\mathrm{CP}}
= \mathds{1}_{\text{direct}} \cdot \frac{1}{100\,|\mathcal{J}|} \sum_{c \in \mathcal{J}} s_c.
\end{equation}
The prompt template is included in Figure~\ref{fig:judge_template}. We also experimented with a penalty-based rubric that instead subtracts weighted penalties for specific failure modes, defined and evaluated in Appendix~\ref{app:penalty_judge}.

% PARA: define D^CP_intra as diversity among the creative primers for the same prompt
\paragraph{Creative primer intra-prompt diversity $D^{\mathrm{CP}}_{\mathrm{intra}}$.}
This measure weighs diversity among the creative primers produced for the same prompt $X^{j}$, encouraging diversity amongst the primers themselves. For primer $Y^{j,i}_{CP}$:
\begin{equation}
\label{eq:cp_dintra}
D^{\mathrm{CP}}_{\mathrm{intra}}
= \frac{1}{G_{\mathrm{CP}} - 1} \sum_{i' \ne i}
\mathrm{d}\!\left(Y_{\mathrm{CP}}^{j,i}, Y_{\mathrm{CP}}^{j,i'}\right).
\end{equation}

% PARA: define D^CP_inter as an aggregate distance from this primer to primers for different prompts in the batch
\paragraph{Creative primer inter-prompt diversity $D^{\mathrm{CP}}_{\mathrm{inter}}$.}
This measure of diversity is the mean distance between a creative primer generated for prompt $j$, $Y^{j,i}_{CP}$, and every primer generated for a different prompt $j'$ in the batch of $B$ prompts, which rewards the primer model for tailoring its primers to each prompt rather than reusing a single generic primer:
\begin{equation}
\label{eq:cp_dinter}
D^{\mathrm{CP}}_{\mathrm{inter}}
= \frac{1}{(B - 1)\,G_{\mathrm{CP}}} \sum_{j' \ne j} \sum_{i'=1}^{G_{\mathrm{CP}}} \mathrm{d}\!\left(Y_{\mathrm{CP}}^{j,i}, Y_{\mathrm{CP}}^{j',i'}\right).
\end{equation}

\subsection{Learning the Creative Process}
\label{sec:cp_decoupling}

% PARA: interpret the method ; why decoupling a trained primer model over a frozen response model escapes the paradox, its modularity payoff, and deviation as optional rather than the message (closing with the headline foreshadow)
Training only the primer model while keeping the response model frozen explicitly decouples learning the abstract creative procedure from producing responses, and is thereby crucial to avoiding the creative paradox.
Typically, in diversity-enhancing RL, the response model is directly rewarded for producing specific outputs. However, because $\pi_{\mathrm{CP}}$ is only rewarded for inducing better responses in a separate model, it is never directly rewarded for reproducing past outputs. It is therefore free to learn generative principles for creativity rather than to reinforce what already worked, which is how it escapes the creative paradox.
Decoupling the creative procedure from generating responses also makes the method modular, such that a single trained $\pi_{\mathrm{CP}}$ can potentially be trained on a small $\pi_{\mathrm{resp}}$ but transferred to a larger, more capable model at inference time.

\section{Experimental Setup}
\label{sec:experimental_setup}

The aim of the experimental approach is to provide a multifaceted view of the potential advantages of creative priming relative to baselines and other creativity-enhancing methods through a number of measures and perspectives.

\paragraph{Models.}
In these experiments all creative priming models $\pi_{\mathrm{CP}}$ and, unless otherwise noted, all response models $\pi_{\mathrm{resp}}$ are \texttt{Qwen3-8B}~\citep{qwen3technicalreport}.
Several targeted models calculate rewards during training: quality is scored by the \texttt{Skywork-Reward-V2-Llama-3.1-8B} reward model, diversity and deviation are calculated from embeddings of the \texttt{Qwen3-Embedding-0.6B} model, DARLING's partition signal depends on a ModernBERT similarity classifier\footnote{\url{https://huggingface.co/dogtooth/similarity-classifier-f168-hf}}, and the creative primer judge is \texttt{Llama-3.3-70B-Instruct}.

% PARA: training data ; Infinite-Chats prompts, quality filter and count, template wrapping and epochs
\paragraph{Training data.}
All methods are trained on a filtered subset of Infinite-Chats\footnote{\url{https://huggingface.co/datasets/LilaSciences/infinite-chats-scored}}~\citep{jiang2025artificialhivemindopenendedhomogeneity} (which is itself a subset of WildChat from~\citealp{zhao2024wildchat}).
To create this subset, an LLM judge scores each prompt's suitability along three axes: open-endedness, complexity, and creative potential. Each axis is rated on a 0--3 scale, and only prompts with a combined score $\geq 6$ are retained, yielding 21{,}075 prompts.
Additionally, when training creative priming models, a pool of 32 custom-written creative primer prompt templates wrap the prompts before they are passed to the priming model.
The reference set for calculating deviation consists of eight responses per prompt from the initial model checkpoint.

% PARA: evaluation data
\paragraph{Evaluation data.}
The evaluation dataset is NoveltyBench~\citep{zhang2025noveltybenchevaluatinglanguagemodels}, which includes 100 real-world open-ended prompts intended to elicit creativity.
To measure quality, the original dataset is extended so that each prompt comes with an LLM judge prompt with two families of components: six universal components fixed across all prompts (instruction-following, coherence, completeness, conciseness, clarity, and surprise), and an additional 1--3 prompt-specific components generated by \texttt{claude-opus-4-20250514}.
For example, ``Tell me a story in five sentences about a girl and her dog'' has 2 prompt-specific components: emotional resonance and narrative economy. The judge prompt template is in Figure~\ref{fig:eval_judge_template} in Appendix \ref{app:templates}.
The evaluation set, including the judge prompts, is available on the HuggingFace Hub.\footnote{\url{https://huggingface.co/datasets/LilaSciences/novelty-bench-llm-judge}}
Evaluation uses the creative primer prompt template in Figure~\ref{fig:priming_template} in Appendix \ref{app:templates}.

% PARA: baselines ; the four comparisons, tie the two trained ones to the Preliminaries limitation, and run them with/without deviation to isolate its effect
\paragraph{Baselines.}
Four baselines provide comparisons to creative priming: (1) \emph{direct generation} from the initial \texttt{Qwen3-8B} checkpoint; (2) an \emph{untrained} creative priming model; (3) \emph{Diversified GRPO} (DGRPO), which trains the response model directly with a multiplicative quality $\times$ intra-group cosine-distance diversity reward; and (4) \emph{DARLING}~\citep{li2025jointlyreinforcingdiversityquality}, which replaces the embedding diversity with a partition-based diversity signal.
Because deviation is a direct component of creativity metrics, in addition to their original formulations, we train both DGRPO and DARLING with and without an added deviation reward to isolate its effect.

% PARA: training details ; group sizes and optimization hyperparameters, response sampling and quality-gate settings, and per-method training budgets
\paragraph{Training hyperparameters.}
When training models, the group sizes are $G_{\mathrm{CP}} = 8$ primers per prompt and $G = 8$ responses per prompt for DGRPO and DARLING, so that each policy's group-relative advantage is estimated on the same number of on-policy samples per prompt.
Additionally, to estimate rewards for each primer, $G_{\mathrm{resp}} = 8$ responses are generated per primer (and their scores averaged as described in Section~\ref{sec:cp_reward_components}).
\cp{} and DGRPO train for two epochs with a batch size of 256, a learning rate of $1\times10^{-6}$, temperature $1.0$ with top-$p$ $0.9$, the quality gate is at threshold $T = -5.0$, and sigmoid temperature $\tau = 1.0$.
The DARLING model is trained with the same hyperparameters from the original paper, including training for ten epochs~\citep{li2025jointlyreinforcingdiversityquality}.  In effect, to be faithful to its original setup, DARLING receives 5 times more epochs than CP or DGRPO. 

% PARA: evaluation setup
\paragraph{Evaluation hyperparameters.}
To evaluate the baselines and \cp{} models, $n = 64$ completions are generated per prompt with temperature $1.0$, top-$p$ $1.0$, and a maximum of $4$,$096$ generated tokens.
Qwen3's thinking mode is left enabled and the thinking span is stripped before scoring, so that every metric is computed on the answers alone.
For \cp{} methods, 1 response is generated per primer, meaning that 64 primers are generated, each separately used as input for 1 response.

\paragraph{Evaluation metrics.}
% Brief intro sentence: atomic metrics and compound metrics
This evaluation reports five \emph{atomic} metrics, each isolating a single characteristic of a method, before combining them into \emph{compound} measures of system-level creativity.

% Brief into to the five atomic metrics
Quality and deviation are the two axes along which an \emph{individual} response is measured for creativity, whereas diversity, Vendi score, and distinct clusters measure the \emph{spread} of a response set.
% quality: LLM-as-a-judge score of an individual response
\emph{Quality} is scored by \texttt{Qwen3-235B-A22B-Instruct-2507} using the prompt-specific judge template shown in Figure~\ref{fig:eval_judge_template} in Appendix \ref{app:templates}.
% deviation: nearest-neighbor distance to the reference set (training instead averages over it)
\emph{Deviation} is the distance to the closest of 64 direct-generation reference responses. That is, the mean operator from training (Eq.~\ref{eq:deviation}) is replaced with a minimum operator so that during evaluation the worst-case nearest neighbor distance is measured.
% diversity: mean pairwise distance within a set
\emph{Diversity} is the mean pairwise cosine distance within a response set (Eq.~\ref{eq:dintra}).
% Vendi score: information-theoretic count of distinct responses
\emph{Vendi score} is an information-theoretic measure of the number of distinct responses in a set~\citep{friedman2023vendiscorediversityevaluation}.
% distinct clusters: semantic-equivalence cluster count, following NoveltyBench
\emph{Distinct clusters} counts the semantic-equivalence clusters in a set, following NoveltyBench~\citep{zhang2025noveltybenchevaluatinglanguagemodels}, using the partition model at \texttt{yimingzhang/deberta-v3-large-generation-similarity}.
Deviation, diversity, and Vendi score use embeddings from \texttt{jinaai/jina-embeddings-v5-text-small}, deliberately different than the embedding model from training to avoid overfitting to a specific model.

% PARA: system-level (compound) metrics ; filter-then-measure-spread at a fixed threshold, then sweep thresholds into AUC/VUS
% introduce the compound metrics via the system-creativity goal that motivates them
Whereas each atomic metric isolates one characteristic, the \emph{compound} metrics combine them to capture \emph{system-level creativity}, i.e.\ the ability to reliably produce a variety of individually creative responses.
% define the filter-then-measure-spread recipe shared by the compound metrics
Each compound metric therefore first discards non-creative responses (responses failing a quality or deviation threshold) and then measures the spread of the survivors using \emph{filtered Vendi score} and \emph{filtered distinct clusters}, so that spread is credited only among responses already deemed creative.
% any single threshold is arbitrary, so sweep thresholds into AUC (one axis) and VUS (both axes)
However, the optimal threshold varies based on the application, so we additionally sweep the thresholds across their full range. Sweeping the deviation (or quality) threshold and integrating over that sweep yields an \emph{area under the curve} (AUC-dev or AUC-qual) that summarizes the method's filtered Vendi score or distinct clusters across all thresholds.
A comprehensive system-level creativity metric can accordingly be formulated through
computing the \emph{volume under the deviation $\times$ quality surface} (VUS) by jointly sweeping thresholds over both metrics.
% defer threshold choices, the Vendi-vs-diversity rationale, and AUC/VUS computation to the appendix
Default thresholds and further metric details are detailed in Appendix~\ref{app:eval_metrics}.

\paragraph{Scaling experiment.}
% Explain the motivation of the scaling experiment: test the creative paradox
In addition to the primary experimental settings above, a further evaluation of the creative paradox is performed by scaling the number of on-policy training samples in DGRPO: if directly reinforcing successful responses limits creativity, then reinforcing more responses should not continually improve system-level creativity.
% Introduce the experiment itself
Accordingly, in this experiment DGRPO (with divergence) is trained with group sizes $G \in \{8,16,32,64\}$.
% Explain why we set it up like this
At the lower endpoint, $G = 8$, DGRPO matches \cp{}'s eight on-policy primers per prompt, as in the primary setup.
However, the reward of each primer for \cp{} is estimated by averaging across eight responses, so at $G = 64$ DGRPO instead matches the total number of responses per prompt for \cp{}, but gets the advantage of eight times more on-policy samples.
% Why DGRPO and not DARLING
The scaling study focuses on DGRPO because DARLING already trains for five times more epochs than \cp{} and DGRPO.

\section{Findings \& Analysis}

The main results in Table~\ref{tab:eval-comparison} are split among untrained methods (top rows), methods trained without a deviation term (middle rows), and methods trained with deviation (bottom rows). Results on atomic metrics are on the left and compound metrics, which aggregate the atomic metrics in various ways, are on the right.
For presentation purposes only, we represent diversity and deviation as scores out of $100$.
Figure~\ref{fig:qual-single} shows one qualitative example output for each method, including the primer that \cp{} produced, with further examples in Appendix~\ref{app:qualitative-examples}.

\begin{table}[t!]
    \centering
    \footnotesize
    \setlength{\tabcolsep}{4pt}
    \renewcommand{\arraystretch}{1.25}
    \resizebox{\textwidth}{!}{%
    \begin{tabular}{@{}cl ccccc|cccccccc@{}}
    \toprule
    & & \multicolumn{5}{c}{Atomic metrics} & \multicolumn{8}{c}{Compound metrics} \\
    \cmidrule(lr){3-7}\cmidrule(lr){8-15}
    & \multirow{2}{*}{Method}
    & \multirow{2}{*}{\makecell{Response\\Diversity}} & \multirow{2}{*}{\makecell{Quality}} & \multirow{2}{*}{Deviation} & \multirow{2}{*}{\makecell{Vendi\\Score}} & \multirow{2}{*}{\makecell{Distinct\\Clusters}}
    & \multicolumn{2}{c}{Filtered} & \multicolumn{2}{c}{AUC-dev} & \multicolumn{2}{c}{AUC-qual} & \multicolumn{2}{c}{VUS} \\
    \cmidrule(lr){8-9}\cmidrule(lr){10-11}\cmidrule(lr){12-13}\cmidrule(lr){14-15}
    & & & & & & & Vendi & Distinct & Vendi & Dist. & Vendi & Dist. & Vendi & Dist. \\
    \midrule
    & \makecell[l]{Qwen3-8b\\\scriptsize{(Reference)}} & \stat{12.3}{1.6} & \stat{90.5}{1.4} & - & \stat{2.12}{0.19} & \stat{18.1}{3.4} & - & - & - & - & - & - & - & - \\
    % & Qwen3-8B & \stat{12.3}{1.6} & \stat{90.5}{1.4} & - & \stat{2.12}{0.19} & \stat{18.1}{3.4} & - & - & - & - & - & - & - & - \\
    & \makecell[l]{\cp\\\scriptsize{(Untrained)}} & \stat{\textbf{20.6}}{1.8} & \stat{\textbf{91.7}}{1.1} & \stat{\textbf{13.1}}{1.3} & \stat{\textbf{3.10}}{0.28} & \stat{\textbf{20.0}}{3.1} & \stat{\textbf{1.48}}{0.35} & \stat{\textbf{4.53}}{1.67} & \textbf{0.60} & \textbf{2.31} & \textbf{182.6} & \textbf{602.6} & \textbf{69.3} & \textbf{294.9} \\
    % & \cp & \stat{\textbf{20.6}}{1.8} & \stat{\textbf{91.7}}{1.1} & \stat{\textbf{13.1}}{1.3} & \stat{\textbf{3.10}}{0.28} & \stat{\textbf{20.0}}{3.1} & \stat{\textbf{1.48}}{0.35} & \stat{\textbf{4.53}}{1.67} & \textbf{0.60} & \textbf{2.31} & \textbf{182.6} & \textbf{602.6} & \textbf{69.3} & \textbf{294.9} \\
    \midrule
    & DGRPO & \stat{26.0}{1.8} & \stat{87.5}{1.4} & \stat{14.1}{1.2} & \stat{4.36}{0.38} & \stat{37.1}{3.6} & \stat{2.18}{0.43} & \stat{6.02}{1.61} & 0.819 & 3.10 & 297.5 & 1010.7 & 107.5 & 509.5 \\
    & DARLING & \stat{11.7}{1.1} & \stat{\textbf{93.6}}{1.2} & \stat{10.1}{0.8} & \stat{2.17}{0.16} & \stat{20.5}{3.3} & \stat{0.57}{0.23} & \stat{1.76}{1.06} & 0.345 & 2.23 & 55.8 & 172.3 & 34.1 & 221.0 \\
    & \cp & \stat{\textbf{31.2}}{1.6} & \stat{88.2}{1.6} & \stat{\textbf{21.2}}{1.3} & \stat{\textbf{5.39}}{0.38} & \stat{\textbf{39.5}}{3.3} & \stat{\textbf{4.04}}{0.46} & \stat{\textbf{16.1}}{2.65} & \textbf{1.404} & \textbf{6.10} & \textbf{463.6} & \textbf{2095.9} & \textbf{160.0} & \textbf{789.0} \\
    \midrule
    \multirow{3}{*}{\rotatebox[origin=c]{90}{\scriptsize{+ Deviation}}}
    & DGRPO & \stat{\textbf{34.0}}{1.6} & \stat{77.3}{2.9} & \stat{\textbf{25.6}}{1.6} & \stat{5.98}{0.47} & \stat{33.4}{3.2} & \stat{2.85}{0.46} & \stat{7.59}{1.77} & 1.015 & 2.78 & 448.1 & 1779.4 & 168.8 & 658.0 \\
    & DARLING & \stat{18.0}{1.1} & \stat{\textbf{88.4}}{1.8} & \stat{24.7}{1.0} & \stat{3.02}{0.18} & \stat{28.1}{3.5} & \stat{2.60}{0.24} & \stat{16.4}{2.80} & 0.835 & 5.24 & 272.8 & 2022.3 & 88.0 & 639.0 \\
    & \cp & \stat{33.4}{1.3} & \stat{85.7}{1.9} & \stat{25.0}{1.2} & \stat{\textbf{6.12}}{0.36} & \stat{\textbf{45.3}}{3.1} & \stat{\textbf{4.70}}{0.43} & \stat{\textbf{22.8}}{3.05} & \textbf{1.641} & \textbf{7.85} & \textbf{525.0} & \textbf{2885.3} & \textbf{186.5} & \textbf{1003.2} \\
    \bottomrule
    \end{tabular}%
    }
    \caption{\textbf{Evaluation on NoveltyBench}. Metrics are defined in Section~\ref{sec:experimental_setup}. Row blocks correspond to untrained methods (top rows), training without deviation (middle), and training with deviation (bottom). Cells include 95\% confidence intervals where available, and the best value per column within each block is bolded. Deviation-derived metrics for the Qwen3-8B reference are 0 by construction and shown as ``-''.
    % takeaway: on atomic metrics, each method makes tradeoffs (but atomic metrics aren't creativity). On compound metrics (various views of creativity), creative priming is a clear winner. paradox, measured
    This table shows that for the atomic metrics, every trained method trades one property for another (DGRPO gains diversity but loses quality; DARLING does the opposite).
    However, the atomic metrics do not measure creativity; the compound metrics do, by filtering for responses that are simultaneously deviating and high-quality and then measuring their spread, and \cp{} dominates on compound measures of system creativity.
    Furthermore, \cp{} \emph{without} a deviation reward already matches or beats baselines trained with deviation, demonstrating the importance of the priming process itself in creativity.
    }
    \label{tab:eval-comparison}
\end{table}

\begin{figure}[!htbp]
    \centering
    \begin{tcolorbox}[
        colback=white, colframe=black!55, boxrule=0.5pt, arc=2pt,
        left=6pt, right=6pt, top=4pt, bottom=4pt, fontupper=\footnotesize,
        title=\small{\textbf{Prompt:}}
        
        \footnotesize{``Imagine you are a time traveler who has just arrived in a new era. Describe the first thing you notice.''}, coltitle=black,
        colbacktitle=black!8, fonttitle=\small,
    ]
    % ===== FULL RESPONSES (verbatim from eval data; visible cells are trimmed for space) =====
    % Reference -- completions[0]:
    % The first thing I notice is the sky—a canvas of swirling, iridescent hues that shift like liquid opal, refracting the light of a sun that seems both familiar and alien. Above, a lattice of floating platforms and translucent tubes hums with latent energy, their surfaces shimmering with data streams that pulse like veins. The air thrums with a low, resonant frequency, a symphony of unseen engines and electromagnetic fields. Beneath my feet, the ground is a mosaic of smooth, adaptive material that molds slightly to my weight, as if the very earth is alive and aware. The scent of ozone tangles with something sweeter—burnt sugar and synthetic blooms—while the faint buzz of drones and the chatter of a language I don't recognize fill the air. It's a world where technology and nature have fused into something neither wholly artificial nor entirely organic, and I feel both exhilarated and disoriented, as though I've stepped into a dream that's too vivid to be real.
    \methodswatch{brandRef}\textbf{Reference.}\hfill{\normalfont\footnotesize\color{black!55}Quality\,=\,98\quad Deviation\,=\,0}\\
    The first thing I notice is the sky---a canvas of swirling, iridescent hues that shift like liquid opal, refracting the light of a sun that seems both familiar and alien. Above, a lattice of floating platforms and translucent tubes hums with latent energy, their surfaces shimmering with data streams that pulse like veins. The air thrums with a low, resonant frequency, a symphony of unseen engines and electromagnetic fields. Beneath my feet, the ground is a mosaic of smooth, adaptive material that molds slightly to my weight, as if the very earth is alive and aware. The scent of ozone tangles with something sweeter---burnt sugar and synthetic blooms---while the faint buzz of drones and the chatter of a language I don't recognize fill the air. It's a world where technology and nature have fused into something neither wholly artificial nor entirely organic.

    \tcbline
    % DGRPO -- completions[0]:
    % Neon veils the sky, ads flicker in reverse on mirrored walkways—your reflection becomes a sale.
    \methodswatch{brandDGRPO}\textbf{DGRPO.}\hfill{\normalfont\footnotesize\color{black!55}Quality\,=\,93\quad Deviation\,=\,24.1}\\
    Neon veils the sky, ads flicker in reverse on mirrored walkways---your reflection becomes a sale.

    \tcbline
    \methodswatch{brandDARLING}\textbf{DARLING.}\hfill{\normalfont\footnotesize\color{black!55}Quality\,=\,98\quad Deviation\,=\,28.5}\\
    A lattice of light not above but around---veins in tree canopies, coral-like sensors spiraling up bamboo, not wires. Not decoration. Each pulse answered a question: what's thriving here? Three specifics crystallized immediately, each deeper than most tech updates:
    \begin{enumerate}[leftmargin=1.6em, topsep=0pt, itemsep=0pt]
        \item \textbf{Priority of context, not speed} (most startling). No red lights screaming clearance. A mother in woven indigo guided a toddler \emph{through} a sensor field monitoring fungal spores in a forest corridor. ``Too close to nutmeg bloom,'' she warned. The field adjusted path width \emph{before} they reached it. No shortcut, no danger alert—\textbf{decision trees rooted in ecology, calculated \emph{why} as fiercely as \emph{how}.} I'd seen adaptive systems, but ethics embedded upstream?
        \item \textbf{Data form} Not screens, not abstract graphs. On a elder's forearm, bioluminescent glyphs \emph{flowed}—not static report, but a negotiation: ``salmon migration 37\% delayed due to turbidity from beaver dams; three upstream weirs could divert 12\% without erosion, but 97\% of spawning beds would.'' It had \textbf{judgment}. No dashboard; a conversation with consequences shaped by local knowledge.
        \item \textbf{Most critical: cost estimates weren't cost-benefit ratios.} A group restoring landslide-scoured hills used a tablet showing \textbf{three branches}.
        \begin{itemize}
            \item ``Replant fast: 40\% survival, flood damage 3x next year''
            \item ``Use mycorrhizae: slower, 75\% survival, \emph{but} soil structure \emph{builds} over five years. Future farms won't erode.''
            \item ``Combine: seedlings + fungi. Say 3.8 years to full maturity. \textbf{2.3x long-term value.}''
        \end{itemize}
        % No ``optimal'' button. They had \textbf{weighted questions}: ``Who eats what, when? How does that shape next storm's path?''
    \end{enumerate}
    [...continues for 5 more lines...]

    \tcbline
    % Creative Priming PRIMER -- groups[0].creative_primer:
    % Focus on a single object or material that *shouldn't exist* in this era based on your knowledge of prior history. Ask: What interaction between old and new tech/materials would create a paradoxical artifact? (E.g., "I see a hand-carved wooden tablet glowing with blue light—how did this technology survive without electricity?") Let your background knowledge create a *specific, concrete* contradiction, not vague "advanced" descriptions. The shock of the unfamiliar must stem from *your* specific expertise, not generic futurism.
    % Creative Priming RESPONSE -- groups[0].completions[0]:
    % The first thing I notice is a hand-carved obsidian tablet, its surface etched with delicate hieroglyphs, but embedded within the cracks are faintly pulsing strands of iridescent metal—strands that shouldn't exist in this era. Obsidian, a volcanic glass, is brittle and prone to fracturing, yet these strands are unbroken, their metallic sheen reflecting light in ways that defy the materials known to this civilization. How did a metallurgical process capable of creating such resilient, light-conducting alloys emerge without the knowledge of chemistry or industrial smelting? The tablet's craftsmanship suggests an ancient society, yet its anomalous composition defies the technological timeline I know. This paradoxical artifact—artisanship meeting alien metallurgy—shatters my understanding of its era's capabilities.
    \methodswatch{brandCP}\textbf{\cp{} primer.} Focus on a single object or material that \emph{shouldn't exist} in this era based on your knowledge of prior history. Ask: what interaction between old and new tech or materials would create a paradoxical artifact? (E.g., ``I see a hand-carved wooden tablet glowing with blue light---how did this technology survive without electricity?'') Let your background knowledge create a \emph{specific, concrete} contradiction, not vague ``advanced'' descriptions. The shock of the unfamiliar must stem from \emph{your} specific expertise, not generic futurism.

    \methodswatch{brandCP}\textbf{\cp{} response.}\hfill{\normalfont\footnotesize\color{black!55}Quality\,=\,100\quad Deviation\,=\,27.6}\\
    The first thing I notice is a hand-carved obsidian tablet, its surface etched with delicate hieroglyphs, but embedded within the cracks are faintly pulsing strands of iridescent metal---strands that shouldn't exist in this era. Obsidian, a volcanic glass, is brittle and prone to fracturing, yet these strands are unbroken, their metallic sheen reflecting light in ways that defy the materials known to this civilization. How did a metallurgical process capable of creating such resilient, light-conducting alloys emerge without the knowledge of chemistry or industrial smelting? The tablet's craftsmanship suggests an ancient society, yet its anomalous composition defies the technological timeline I know. This paradoxical artifact---artisanship meeting alien metallurgy---shatters my understanding of its era's capabilities.
    \end{tcolorbox}
    \caption{\textbf{Qualitative examples.} Representative responses to an open-ended prompt (more cherry- and lemon-picked examples in Appendix~\ref{app:qualitative-examples}). To get a qualitative understanding of creativity, each label reports the response's judge quality score and its deviation from the reference model (as described in Section~\ref{sec:experimental_setup}). In this example, the reference model paints an evocative scene in flowing prose; DGRPO answers in one terse line; DARLING returns a highly-formatted, bulleted analysis; and \cp{} answers in the reference model's style while inventing a novel, concrete image, illustrating that its deviation lies in \emph{what} it says rather than in surface formatting. These characteristics are consistent across all prompts (DGRPO's terseness and DARLING's length and use of formatting), as analyzed in Appendix~\ref{app:style-impact}.}
    \label{fig:qual-single}
\end{figure}

\paragraph{Performance at first glance.}
% PARA: read the atomic metrics off Table 1; at first glance every trained method just trades quality for diversity/deviation, differing only in how steep the trade is and what it buys
The atomic metrics suggest that every method trades one property for another, differing only in how steep the tradeoff is.
For example, within the models trained for quality and diversity (middle rows), DGRPO improves diversity ($26$), Vendi score ($4.36$), and distinct clusters ($37.1$) over the reference, at a slight cost to quality ($90.5 \to 87.5$).
DARLING makes the opposite trade, achieving the highest quality of any method ($93.6$) while its diversity ($11.7$) falls slightly below the reference ($12.3$), and its Vendi score and distinct clusters barely move ($2.17$ and $20.5$).
\cp{} makes the same tradeoff, but on more favorable terms, losing only a little quality ($90.5 \to 88.2$) in return for the highest diversity ($31.2$), deviation ($21.2$), Vendi score ($5.39$), and distinct clusters ($39.5$) of any method not trained with deviation.

% PARA: adding deviation steepens DGRPO's quality collapse but makes DARLING the well-rounded baseline (highest quality, much-improved deviation/distinct), though it still trails CP on diversity and distinct clusters; CP pays the least for the most distinct clusters
Adding deviation to the reward signal increases the deviation score for all three trained methods, but they respond very differently.
DGRPO improves significantly in diversity ($26 \to 34$), but the quality ($87.5 \to 77.3$) and number of distinct clusters decrease ($37.1 \to 33.4$).
DARLING also enjoys significant improvement in diversity ($11.7 \to 18$) and the number of distinct clusters ($20.5 \to 28.1$), but drops some quality ($93.6 \to 88.4$).
\cp{} also loses quality ($88.2 \to 85.7$), but with less tradeoff than DGRPO and DARLING, in return for the most distinct clusters of any method ($45.3$), while holding diversity and deviation high.

\paragraph{The creative paradox, measured.}
% PARA: the compound metrics are several views of one definition of creativity (filter to creative responses, measure their spread); this is where CP separates, since the baselines mislead the atomic metrics in opposite ways
The atomic metrics from above, however, do not tell us whether an \emph{individual response is creative} (simultaneously deviating and high-quality), nor whether a \emph{system is creative} (generates a large spread of creative responses).
The compound metrics directly capture the joint demands of creativity by filtering out individual responses that are not creative and then measuring the spread of the remaining responses, which is where the difference between \cp{} and the baseline approaches becomes evident.

Among the methods trained without deviation, DARLING has high-quality but low-diversity responses, so its filtered Vendi ($0.57$) and filtered distinct clusters ($1.76$) are the lowest in the block.
DGRPO is the opposite, producing almost as many distinct clusters as \cp{} ($37.1$ and $39.5$); however, most responses fall below the quality or deviation threshold and are filtered out, dropping its filtered distinct clusters to just $6.02$.
Compared to the baselines, \cp{} leads to better performance on both filtered metrics ($4.04$ filtered Vendi and $16.1$ filtered distinct clusters) and on metrics that sweep across all threshold values (AUC-dev, AUC-qual, and VUS in Table~\ref{tab:eval-comparison}).
None of these results mean that the baselines are failing: each performs well on its intended objective, but neither achieve the simultaneous deviation, quality, and diversity expected of a creative system.

% PARA: deviation is central to creativity, yet CP without a deviation term still matches or beats deviation-trained baselines on the compound creativity metrics ; so the process, not optimizing deviation, is what matters
Deviation is a core ingredient of what makes a response, and thus a system, creative.
However, even when the baselines \emph{are} trained with an explicit deviation reward, they do not outperform a \cp{} model trained \emph{without} deviation on the compound metrics.
On filtered Vendi, \cp{} without deviation ($4.04$) already exceeds both DGRPO ($2.85$) and DARLING ($2.6$) trained with deviation.
Filtered distinct clusters tell the same story: \cp{} without deviation ($16.1$) stays well ahead of DGRPO ($7.59$) and draws even with DARLING ($16.4$), the strongest baseline.
That \cp{} does so even without explicitly optimizating for deviation suggests that optimizing for the individual components of creativity only goes so far, and that learning the creative \emph{process} does more to drive creativity (Section~\ref{sec:cp_decoupling}).

\begin{figure}[t!]
    \centering
    \captionsetup[subfigure]{margin={32pt,0pt}}
    \begin{subfigure}[t]{0.49\linewidth}
        \centering
        \includegraphics[width=\linewidth]{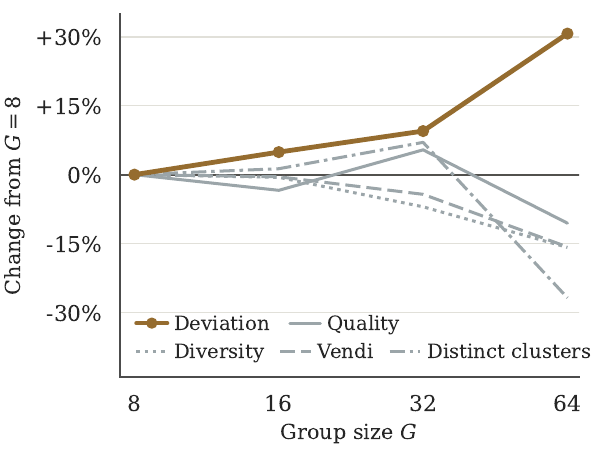}
        \caption{Atomic Metrics}
        \label{fig:dgrpo-group-scaling-atomic}
    \end{subfigure}\hfill
    \begin{subfigure}[t]{0.49\linewidth}
        \centering
        \includegraphics[width=\linewidth]{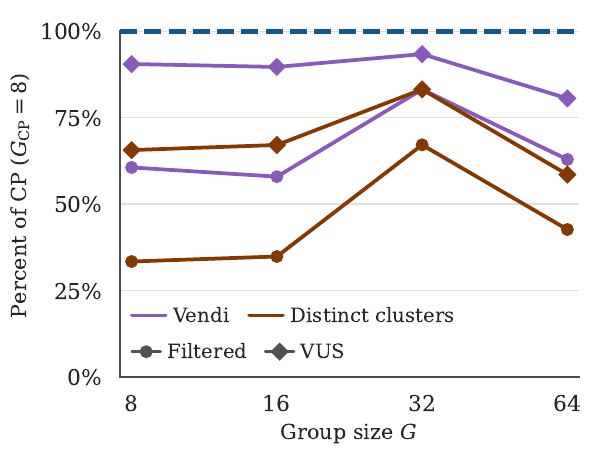}
        \caption{Compound Metrics}
        \label{fig:dgrpo-group-scaling-compound}
    \end{subfigure}
    \caption{\textbf{Scaling group size does not escape the creative paradox.}
    DGRPO is trained with $G \in \{8,16,32,64\}$ on-policy samples per prompt and evaluated on NoveltyBench.
    % $G = 8$ is the baseline setting in Table~\ref{tab:eval-comparison} and $G = 64$ matches the total responses used by \cp{} to estimate rewards for primers (for eight primers, each estimating the reward over eight responses).
    (a) Each of the five atomic metrics is shown as a percent change from DGRPO with $G = 8$, with deviation highlighted in copper.
    (b) Four compound metrics of creativity are shown as a percentage of \cp{} performance, where line color represents the measure of spread (Vendi score or distinct clusters) and marker shape represents the filtering and aggregation method (Filtered or VUS).
    These results demonstrate that additional on-policy samples consistently improve deviation, but all other atomic metrics ultimately decline at $G = 64$. System-level measures of creativity (compound metrics) peak at $G = 32$, but even then fall below the performance of \cp{}, and decline significantly at $G = 64$.
    Full results are in Table~\ref{tab:eval-comparison-scaling} in Appendix \ref{app:dgrpo-group-scaling}.
    }
    \label{fig:dgrpo-group-scaling}
\end{figure}

% Setup for scaling experiment
One might expect that direct generation methods should improve on metrics of creativity when trained with a larger group size, as more on-policy samples give a model more opportunities to explore and to reinforce high-reward responses.
% Scaling group sizes for DGRPO
However, interestingly, scaling the number of on-policy samples for DGRPO trained with deviation does not allow it to escape the tradeoff observed above (DGRPO is evaluated here instead of DARLING because it performs better on a majority of compound metrics while training for 5 times fewer epochs). 
Compared to the reference model, at $G = 8$, DGRPO already gains diversity and deviation at the cost of quality, and Figure~\ref{fig:dgrpo-group-scaling-atomic} shows that increasing group sizes to $G\in\{16,32,64\}$ does not resolve the tradeoffs, but can actually further exaggerate them.
% Atomic metrics
As $G$ increases, deviation is the only metric that rises consistently, increasing from $25.6$ to $33.5$. Diversity and Vendi score decline steadily, while quality and distinct clusters fluctuate before reaching their lowest values at $G = 64$ ($69.1$ and $24.5$, respectively).

% Compound metrics
The compound metrics reveal the consequences of this tradeoff (Figure~\ref{fig:dgrpo-group-scaling-compound}, with full results in Appendix~\ref{app:dgrpo-group-scaling}), where at $G = 32$, empirically the best group size, DGRPO still has worse performance than \cp{} on every compound metric. At best, DGRPO with $G = 32$ achieves 93.4\% of the VUS-Vendi that \cp{} achieves, and at worst, only 65.2\% of the AUC-dev distinct clusters. Ultimately, at $G = 64$ \emph{every} metric of system creativity decreases from the peak. Even though scaling up to $G = 64$ gives DGRPO a compute advantage versus lower values of $G$, that additional compute neither closes the gap nor produces continual improvements in creativity, providing intriguing further evidence for the creative paradox: \emph{directly reinforcing successful responses can improve creativity only up to a point, after which reinforcing more responses begins to undermine system-level creativity}.

\paragraph{Decomposing the Creative Priming advantage.}
% PARA: within-CP ladder (untrained -> trained no-div -> trained +div) locates where the gains come from: the process alone helps, training delivers the bulk, deviation adds a further increment
To disentangle the contributions of the priming process, RL training, and the deviation reward, we compare three \cp{} variants.
If Qwen3-8B serves as the creative primer without any additional RL (the untrained \cp{} row), diversity already increases from $12.3$ to $20.6$ over direct generation while judged quality stays flat ($91.7$ versus $90.5$).
However, the \cp{} model trained with quality and diversity still dominates the untrained variant on nearly every metric (with a minor decrease in quality).
Further including the deviation reward improves the creativity metrics still further, at an additional small cost to quality.
This analysis shows that the creative priming process is valuable on its own, but RL training delivers significant additional gains; the process supplies a mechanism by which creativity can emerge, and training teaches the model to optimize that mechanism.

\begin{figure}[t!]
    \centering
    \includegraphics[width=\linewidth]{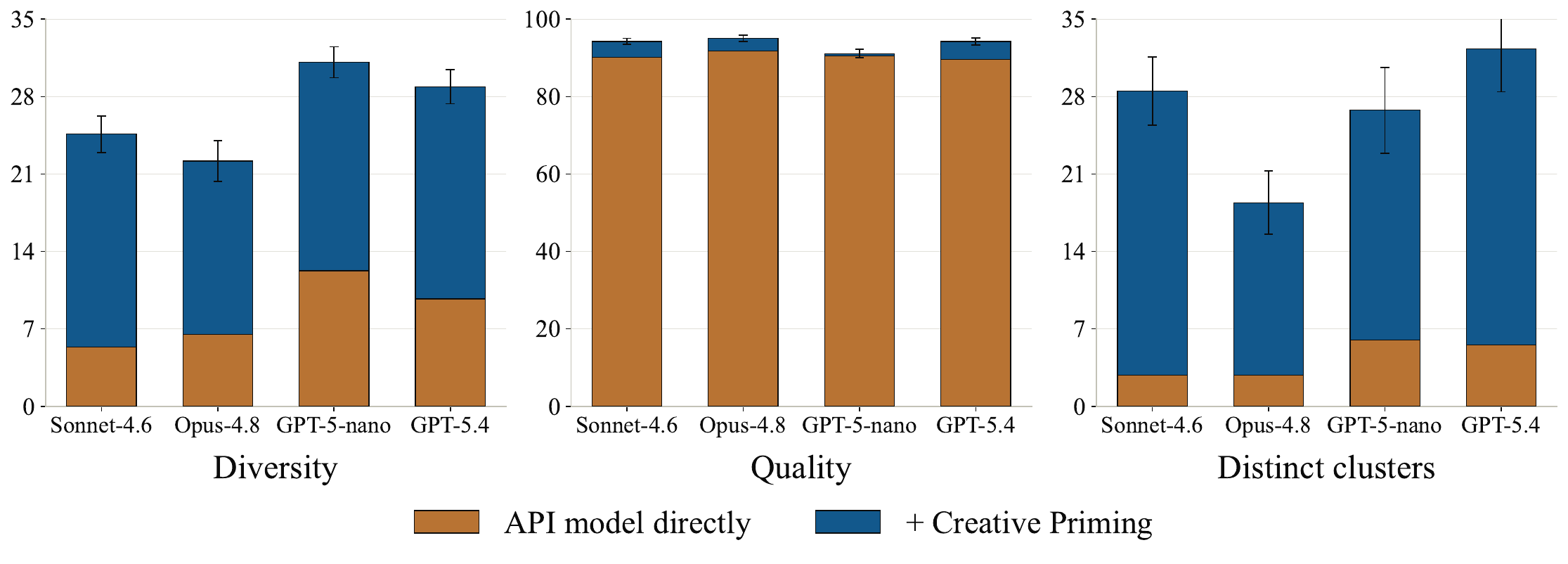}
    \caption{\textbf{Creative priming transfers to API response models}.
    Each panel reports one evaluation metric for four API models, comparing direct generation (copper) with the addition of the creative priming model (blue; trained with deviation). Error bars give 95\% confidence intervals on the total. Results are reported in full in Table~\ref{tab:api-transfer} in Appendix \ref{app:eval_metrics}.
    This figure shows that the same \cp{} model trained to improve creativity on a Qwen3-8B response model leads to significantly increased diversity and distinct clusters when using an API response model, while maintaining or improving quality.
    }
    \label{fig:api_transfer}
\end{figure}

\paragraph{The learned creative process transfers to frontier models.}
% PARA: rule out that the primer overfits its training response model ; a single CP model transfers to four frozen API models, with filtered distinct in the same range as on Qwen3-8B
The creative priming model is only trained with a Qwen3-8B response model, so it is natural to consider that it may overfit to that model.
We test whether the learned creative process transfers by pairing trained \cp{} models with four unseen response models (Claude Sonnet-4.6, Claude Opus-4.8, GPT-5-nano, and GPT-5.4), evaluating each against its own direct-generation reference.
% Atomic metrics (comparing API model scores vs. CP+API model)
Across all four API models, \cp{} increases diversity by $2.5$--$4.6\times$ and distinct clusters by $4.5$--$10\times$  while maintaining or improving quality (Figure~\ref{fig:api_transfer}).
For example, with Sonnet-4.6, distinct clusters increase from $2.8$ to $28.5$ and diversity from $5.3$ to $24.6$, while quality also increases from $90.2$ to $94.3$.
In this way, the evidence suggests that the learned process effectively improves the atomic metrics of response models unseen during training.

% Compound metrics (comparing CP+Qwen3-8B vs. CP+API models)
We next evaluate whether the learned process also elicits an increase in system-level creativity metrics from each API model.
Because each response model is evaluated against its own direct-generation reference, the compound metrics are model-relative: they measure whether \cp{} leads to varied responses that depart from what that model typically generates while meeting its own quality threshold.
Accordingly, these creativity scores cannot be directly compared across response models.
Table~\ref{tab:api-transfer} in Appendix \ref{app:api_transfer} shows that the deviation-trained \cp{} model achieves filtered Vendi scores ranging from $3.26$--$4.71$ and filtered distinct scores of $10.4$--$21.9$ across the four API models.
For context, \cp{} paired with Qwen3-8B achieves a filtered Vendi score of $4.70$ and a filtered distinct score of $22.8$, falling within a similar range as the API models.
For the API models, VUS Vendi ranges from $112.8$--$177.7$, and VUS distinct clusters are $395.3$--$822.7$.
Notably, while creative priming elicits the highest creativity scores when paired with GPT-5.4 and GPT-5-nano, it achieves the lowest creativity scores when paired with Opus-4.8. This is possibly due to Opus-4.8 having the highest quality scores from its direct response baseline ($91.7$), combined with the lowest deviation scores from the \cp{} model ($17.1$). 
Table~\ref{tab:api-transfer} in Appendix~\ref{app:api_transfer} reports the complete results.

\begin{figure}[t!]
    \centering
    \includegraphics[width=0.95\linewidth]{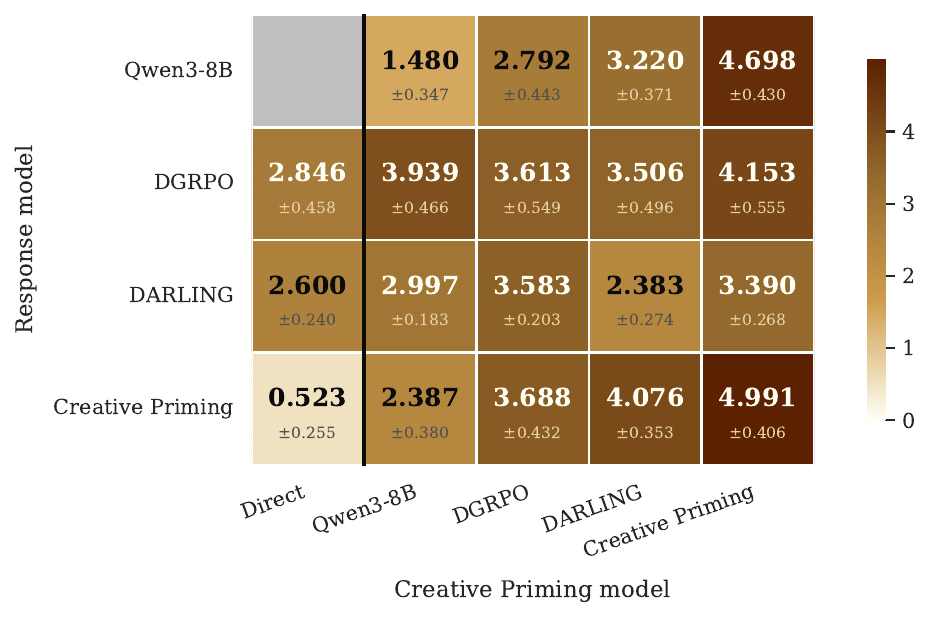}
    \caption{\textbf{Model combinations for the creative priming pipeline}. Each cell reports filtered Vendi score (Qwen3-8B direct has 0 deviation from itself). The farthest left column shows scores when using the model listed on the y-axis as the response model (i.e.\ without \cp{}). The remaining columns represent scores when using the model listed along the x-axis as the \cp{} model and the model listed along the y-axis as the response model.
    These results show that, compared to direct response generation, the \cp{} pipeline itself improves creativity, even when using \cp{} models not explicitly trained for priming (i.e.\ DGRPO and DARLING).}
    \label{fig:cp_rm_comparisons}
\end{figure}

\paragraph{Modularity in the creative priming pipeline.}
% PARA: vary the primer model too over the full cross-product; the pipeline itself carries the result (any primer helps every response model), the trained CP primer is best, and even DGRPO/DARLING make decent primers
The previous experiment varies the response model, but the creative priming pipeline also permits varying the creative priming model.
We evaluate filtered Vendi scores across the full cross-product of \{creative priming model $\times$ response model\} pairs, drawing both roles from the deviation-trained baselines, a trained \cp{} model, and an untrained Qwen3-8B prompted as a primer (Figure~\ref{fig:cp_rm_comparisons}).
The main result of this analysis is that the priming pipeline itself improves every response model over direct generation.
Specifically, the trained \cp{} model is the best primer of the group, giving the three highest-scoring cells of the grid ($4.153$--$4.991$).
However, even DGRPO and DARLING work well as primers: used to prime the initial Qwen3-8B (top row of Figure~\ref{fig:cp_rm_comparisons}), they reach $2.792$ and $3.22$, well above the untrained primer ($1.480$), despite never being trained to prime another model.

\begin{figure}[t!]
    \centering
    \begin{minipage}[b]{0.58\linewidth}
        \centering
        \includegraphics[width=\linewidth]{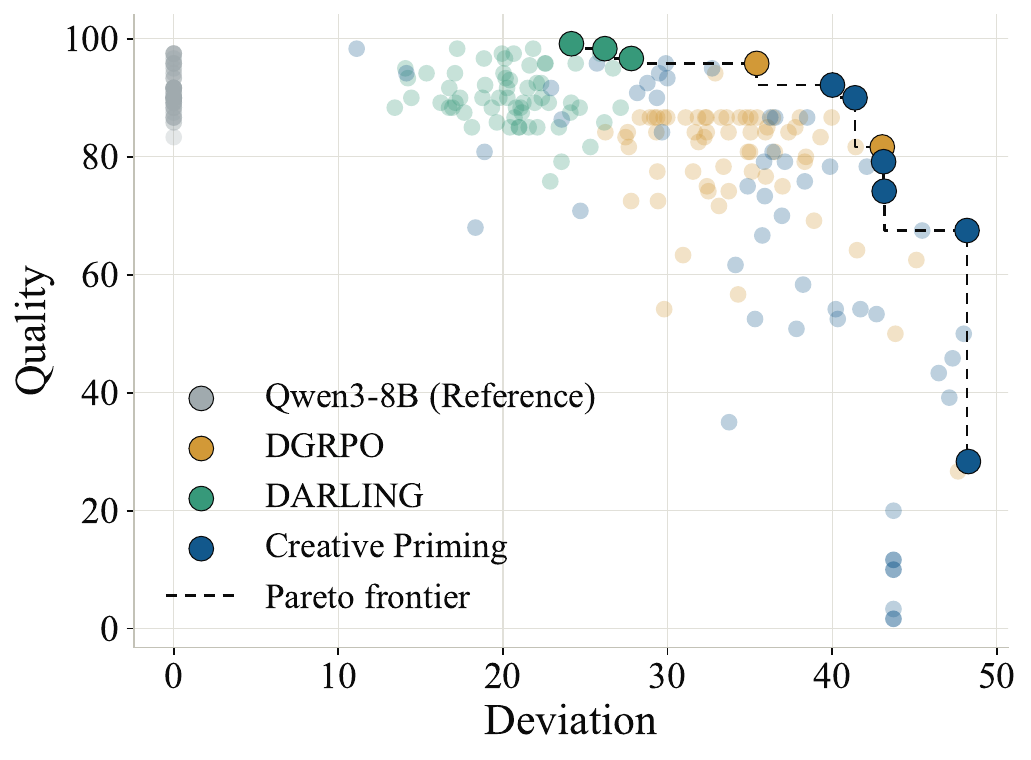}
        \caption{\textbf{The quality-deviation Pareto frontier} of models trained with deviation for the prompt: ``Suggest another strong 4 character password.'' In this particular example \cp{} populates the high-deviation region, while DARLING populates the high-quality region, and DGRPO sits in between.}
        \label{fig:pareto_frontier}
    \end{minipage}\hfill
    \begin{minipage}[b]{0.4\linewidth}
        \centering
        \footnotesize
        \begin{tabular}{@{}l cc@{}}
        \toprule
        Method & \makecell{Without\\Deviation} & \makecell{With\\Deviation} \\
        \midrule
        Qwen3-8b (Ref.) & 0.2 & 0.0 \\
        DGRPO & 19.1 & 32.2 \\
        DARLING & 8.1 & 22.5 \\
        \cp & \textbf{72.6} & \textbf{45.3} \\
        \bottomrule
        \end{tabular}
        \captionof{table}{\textbf{Share of the quality-deviation Pareto frontier} (\%), averaged over all prompts, and calculated separately per column. The best value per column is bolded. \cp{} claims the largest share of the frontier both with and without deviation.}
        \label{tab:pareto-frontier}
    \end{minipage}
\end{figure}

\paragraph{The quality-deviation frontier.}
% PARA: define the Pareto-frontier share metric (a QD-style trade-off measure, not a creativity metric) and how it is computed per group
Given that models exhibit an inherent trade-off between quality and deviation, it is interesting to measure the \emph{Pareto-frontier share} of each method on the quality-deviation frontier, a measure of how well a method handles that trade-off.
To capture this trade-off, each response is plotted on a quality-deviation chart (separately for each evaluation prompt) with the Pareto frontier (the non-dominated responses) highlighted, and the percent of the frontier composed of each method's responses is then calculated. Figure~\ref{fig:pareto_frontier} shows an example of this approach for one prompt.
Finally, the scores are averaged across all prompts, measuring how often each method produces responses that dominate the competitors on quality and deviation jointly.
Because each frontier is relative to the pool of competing methods, we compute the share within each pool (with and without deviation), where it sums to 100\%. Table~\ref{tab:pareto-frontier} reports these shares.

% PARA: CP dominates the frontier in both groups, even though it was not top on either raw metric, and note the shares are not comparable across groups
\cp{} claims the majority of the frontier both with and without deviation, and by a wide margin without deviation (72.6\%).
This trend holds even where \cp{} tops neither raw quality or deviation: among the deviation-trained methods it leads neither deviation nor quality alone, yet it still contributes the largest share of jointly non-dominated responses ($45.3\%$).

% PARA: summary of Findings ; frame the results as confirming the paradox thesis (baselines re-tread it, CP escapes it), note the conclusion holds across every view measured, then hand off to the Discussion
\paragraph{A consistent picture.}
Where the baselines re-tread the creative paradox by training on their own past successes, \cp{} is built to escape it---and the results bear this out, with \cp{} substantially outperforming the baselines as a creative system.
This trend holds regardless of how we look: the filtered compound metrics, the threshold-free sweeps, transfer across frozen response models, and variation of the primer model itself all tell the same story.
We turn next to the broader implications of these results.

\clearpage

\section{Discussion}
% PARA: derive the creativity-as-a-process thesis (novelty is spent when produced) from the definition
%setting the stage a little more dramatically here
A major implication of the creative paradox, along with the effectiveness of creative priming, is that the ``normal way'' of thinking about machine learning may not ultimately work as well for creativity.
In the normal framing, a model learns from examples of good behavior and gradually improves at reproducing such behavior.
Creativity upends this intuitive notion by offering a case where you actually \emph{do not} want future behavior to resemble past behavior.
Or, to be more nuanced about the precise challenge: there is \emph{one sense} in which we definitively do not want the output to resemble the training while there is \emph{another sense} in which we do.
On the one hand, the products of creative generation themselves should diverge from their precedent in training, while on the other hand the \emph{process behind} those generations should follow shared principles.
The problem is that the underlying process is generally not in the training data and, even if it were, it would need to be separated from the products it led to.
The creative paradox emerges from this delicate conflict.

% PARA: state the thesis (creativity lives in the generating process, not in output pairs); show our results are what that view predicts, and use it to redefine what escaping the creative paradox requires
The main hypothesis behind creative priming is
therefore that creativity does not live in individual input/output pairs but in the process that generates them.
Four results support this view:
\begin{enumerate}
    \item Table~\ref{tab:eval-comparison} shows that even a deviation-less \cp{} model improves on compound creativity metrics compared to baselines that are explicitly trained to deviate from past responses;
    \item Figure~\ref{fig:dgrpo-group-scaling} shows that increasing training compute for direct generation hits a peak performance which, by all measures of system creativity, is worse than \cp{};
    \item Figure~\ref{fig:api_transfer} shows that the creative process learned by the priming model on a small open-source response model transfers to frontier API response models, leading to a 4.5-10$\times$ increase in the number of distinct responses and a 2.5-4.6$\times$ increase in response diversity; and
    \item Figure~\ref{fig:cp_rm_comparisons} shows that the priming process itself improves response models even when the model generating the primer was not trained for priming.
\end{enumerate}
All these results point to a learned \emph{procedure} that need not reinforce any particular output and does not depend critically upon any particular response model.
Avoiding the creative paradox thus requires learning the generative principle behind creative successes rather than learning to reproduce those same successes reliably.

% PARA: the components of creativity trade off under direct optimization, which is why decoupling helps; decoupling lets the response model maximize quality while the primer supplies diversity
A key facet of creative priming is decoupling priming from response generation.  How critical is such a decoupling?
When a response model is optimized directly, the components of creativity trade off against one another: DGRPO increases diversity and deviation by sacrificing quality, while DARLING protects quality at the expense of diversity.
Furthermore, additional compute does not break this trade-off for DGRPO: deviation consistently improves with increasing group sizes, but the other atomic metrics ultimately decline, leading to a creativity ceiling that scale does not fix.
Decoupling the priming from the response softens this trade-off.
The response model keeps doing exactly what it is trained for, producing the highest-quality answer, while the primer supplies the shift toward diverse regions of the output distribution, so neither objective is compromised for the other.
This division of labor is how the primer ultimately learns creative principles rather than to reproduce high-quality answers itself.

% Show that the primer leads to less style change than the control methods
Additional evidence to support decoupling relates to style versus substance.
If a method genuinely increases creativity, it should change the \emph{content} of the responses more than the surface-level style.
While experiments in Appendix~\ref{app:style-impact} show that all three methods change style relative to the reference, 
\cp{} drifts least from the reference response model's style, exhibiting the lowest function-word JS-divergence of the three (0.045, where lower means closer to matching the reference's style, compared with DGRPO's 0.087 and DARLING's 0.138).
Decoupling predicts precisely such consistency. The primer steers content, whereas training the response model directly entangles the content changes with a larger stylistic shift. Figure~\ref{fig:qual-single} shows a qualitative example of the style differences.

% PARA: answer the anticipated single-model objection ; one model can play both roles, but only if the primer stays a separate stage rather than being folded into monolithic thinking
A natural question is whether a single model could internalize the creative priming process and use it within its chain-of-thought.
The obstacle to this more straightforward approach is that models are trained toward the highest-quality response. As a result, it is unlikely that one model captures the full distribution of high quality outputs with enough likelihood to also be reliably diverse.
That limitation is exactly what the primer exploits. Modifying the response model's prompt pushes it into a different region of the input distribution and elicits a shifted output distribution, so the primer is a learned, external way to induce the shift rather than asking one model to be simultaneously quality-maximizing and distribution-covering.
Nevertheless, in principle perhaps there is a way for a single model to write the primer for itself and then respond to it, so long as priming stays a \emph{separate stage}.
Folding creative priming into a single model is an interesting direction for future investigation.

% PARA: the two practical payoffs of a modular primer ; train it once and reuse it across models/domains, and compose it with existing test-time diversity methods rather than replacing them
An intriguing benefit of the separated primer in this paper is that it is modular and can be trained once and reused, i.e.\ swapped across response models or even moved to a new domain by pairing it with a domain-specific response model.
Furthermore, priming can be combined with most test-time diversity-enhancing methods, including sequential sampling~\citep{kang2025roadtraveledenhancingexploration}, String Seed of Thought~\citep{misaki2026string}, vector policy optimization~\citep{bahlousboldi2026vectorpolicyoptimizationtraining}, verbalized sampling~\citep{zhang2026verbalizedsamplingmitigatemode} as shown in Appendix~\ref{app:vs-composition}, or, as shown in Figure~\ref{fig:cp_rm_comparisons}, a diversity-trained response model.

% PARA: implications of our definitions of creativity, leading into an even more complex measurement of creativity
In Section~\ref{sec:creativity}, a response was defined as being creative when it ``simultaneously deviates from the reference set and is high-quality'' and a system was defined as creative ``when the system reliably produces responses that are individually creative and are also varied amongst themselves.'' An important consequence of these definitions is obvious in hindsight: the moment a model produces an idea, it is, by definition, no longer new to that model. \emph{Novelty is spent in the very act of its creation}.
Creativity is thus not a fixed property of any single output, but an ongoing process of repeatedly generating original, high-quality ideas.
In this work, the reference set from the response model was frozen prior to training, enabling a comparable computation of deviation for all methods.
Looking ahead, there is room for a richer, and more complex, measurement that allows for a non-stationary reference set. By computing deviation against the full history of a model's outputs, the temporal dimension of creativity can be directly captured: instead of a snapshot of creativity at one instant, a system can be evaluated for the ability to sustain novelty against its own growing past history.
However, such a measurement makes comparisons more complicated because each system, even for two runs of the same method, will accumulate their own distinct reference sets.

\paragraph{Limitations.}
One obvious downside of the creative priming pipeline is that two-stage generation adds additional inference cost over direct generation.
Additionally, the priming pipeline is bounded by the frozen response model: it can only elicit what the response model is already capable of producing and lacks any means to add knowledge or skill that the response model does not already have.
Finally, the training and evaluation setup in this work depend on imperfect signals from which the method will inherit failure modes. For example, embedding-based diversity and deviation metrics may miss forms of creative variation, while LLM judges and reward models give imperfect quality signals. That said, better signals should directly improve the learned primer, so the outcomes can in principle improve along with its supporting technology.

\section{Conclusion}
This work identifies the \emph{creative paradox}, the tension that training on past creative successes teaches a model to reproduce them, while true creativity demands moving beyond previous outputs. The proposed solution to get around the paradox is \emph{creative priming}. Compared to directly training the response model, \cp{} produces more responses that are simultaneously novel, high-quality, and semantically distinct.  The deeper lesson is that if creativity is a process rather than an output, then optimizing the creative process itself, learning to generate better primers rather than better answers, is really where the discourse on optimizing for creativity should focus.  We hope this work will inspire more discussion and investigation in this spirit.

\newpage

\bibliography{iclr2027_conference}
\bibliographystyle{iclr2027_conference}

\newpage
\appendix
\section{Algorithms}

This section provides pseudocode algorithms for GRPO (Algorithm~\ref{alg:grpo}), creative priming (Algorithm~\ref{alg:cp}), and DGRPO (Algorithm~\ref{alg:rm_diversity}) along with additional details on the reward components.

\subsection{GRPO}

\begin{algorithm}[H]
\caption{Group Relative Policy Optimization (GRPO)}
\label{alg:grpo}
\begin{algorithmic}[1]
\Require Policy model $\pi$
\Require Reference policy $\pi_{\text{ref}}$
\Require Dataset of prompts $\mathcal{D} = \{ X^{1}, X^{2}, \dots \}$
\Require Group size $G$
\Require Learning rate $\alpha$, Batch size $B$, Clipping parameter $\epsilon$, KL weight $\beta$
\Statex
\While{Still Improving}
    \State Sample a minibatch of prompts: $\{X^{j}\}_{j=1}^{B} \sim \mathcal{D}$

    \For{each prompt $X^{j}$}
        \State Sample a group of responses: $\{Y^{j,1}, \dots, Y^{j,G}\} \sim \pi(\cdot \mid X^{j})$
        \State Compute raw rewards: $r_{j,i} \gets r(X^{j}, Y^{j,i})$ for $i = 1, \dots, G$
        \State Group statistics: $\displaystyle \mu_j = \frac{1}{G} \sum_{i=1}^{G} r_{j,i}, \quad \sigma_j = \sqrt{\frac{1}{G} \sum_{i=1}^{G} (r_{j,i} - \mu_j)^2 + \delta}$
        \State Group-relative advantages: $\displaystyle A_{j,i} = \frac{r_{j,i} - \mu_j}{\sigma_j}$
    \EndFor

    \State Importance ratios: $\displaystyle \rho_{j,i} = \frac{\pi(Y^{j,i} \mid X^{j})}{\pi_{\text{old}}(Y^{j,i} \mid X^{j})}$

    \State Clipped objective:
    \[
    \mathcal{L}_{\text{GRPO}} =
    \mathbb{E}_{j,i}
    \Big[
    \min\big(
    \rho_{j,i} A_{j,i},
    \text{clip}(\rho_{j,i}, 1-\epsilon, 1+\epsilon) A_{j,i}
    \big)
    \Big]
    \]

    \State Add KL regularization to the reference policy:
    \[
    \mathcal{L} =
    \mathcal{L}_{\text{GRPO}}
    - \beta \, \mathrm{KL}\big(\pi(\cdot \mid X^{j}) \,\|\, \pi_{\text{ref}}(\cdot \mid X^{j})\big)
    \]

    \State Update policy $\pi$ using gradient ascent on $\mathcal{L}$
    \State Set $\pi_{\text{old}} \leftarrow \pi$
\EndWhile

\end{algorithmic}
\end{algorithm}

\subsection{Creative Priming Training Algorithm}
\label{app:cp_algorithm}

% point back to the main-text loss statement this algorithm expands
Algorithm~\ref{alg:cp} gives the full training loop for the creative priming model described in Section~\ref{sec:cp_algorithm}. It instantiates the standard GRPO loop (Algorithm~\ref{alg:grpo}) with the two-level primer/response sampling and the multiplicative $\mathrm{Reward}(\cdot)$ of Eq.~\ref{eq:reward}.

\begin{algorithm}[H]
\caption{Creative Priming}
\label{alg:cp}
\begin{algorithmic}[1]
\Require Creative priming policy $\pi_{\mathrm{CP}}$, reference policy $\pi_{\mathrm{CP_{ref}}}$, frozen response model $\pi_{\mathrm{resp}}$
\Require Embedding model $\phi$, reward model $r_{\mathrm{RM}}$, creative primer judge $J_{\mathrm{CP}}$
\Require Creative priming prompt template $X_{\mathrm{CP}}$, dataset of prompts $\mathcal{D} = \{X^{1}, X^{2}, \dots\}$
\Require Creative primer and response group sizes $G_{\mathrm{CP}}, G_{\mathrm{resp}}$
\Require Precomputed reference responses $\{Y_{\mathrm{ref}}^{j,k}\}_{k=1}^{G_{\mathrm{resp}}}$ per prompt (only when the deviation signal is enabled)
\Require Learning rate $\alpha$, batch size $B$, clipping parameter $\epsilon$, KL weight $\beta$
\Statex
\While{Still Improving}
    \State Sample a minibatch of prompts: $\{X^{j}\}_{j=1}^{B} \sim \mathcal{D}$

    \For{each prompt $X^{j}$}
        \For{$i = 1$ to $G_{\mathrm{CP}}$}
            \State Sample creative primer: $Y_{\mathrm{CP}}^{j,i} \sim \pi_{\mathrm{CP}}(\cdot \mid X_{\mathrm{CP}}, X^{j})$
            \State Build response-model input by populating the template with prompt and primer: $X_{\mathrm{resp}}^{j,i} = \mathcal{T}(X^{j}, Y_{\mathrm{CP}}^{j,i})$
            \State Sample responses: $\{Y_{\mathrm{resp}}^{j,i,1}, \dots, Y_{\mathrm{resp}}^{j,i,G_{\mathrm{resp}}}\} \sim \pi_{\mathrm{resp}}(\cdot \mid X_{\mathrm{resp}}^{j,i})$
        \EndFor

        \For{$i = 1$ to $G_{\mathrm{CP}}$}
            \State $r_{j,i} \gets \mathrm{Reward}\!\left(X^{j}, Y_{\mathrm{CP}}^{j,i}, \{Y_{\mathrm{resp}}^{j,i,k}\}_{k=1}^{G_{\mathrm{resp}}}, \{Y_{\mathrm{ref}}^{j,k}\}_{k=1}^{G_{\mathrm{resp}}}, \{Y_{\mathrm{CP}}^{j,i'}\}_{i' \ne i}, \{Y_{\mathrm{CP}}^{j',i'}\}_{j' \ne j,\, i'}\right)$ \Comment{Eq.~\ref{eq:reward}}
        \EndFor

        \State Group statistics: $\displaystyle \mu_j = \frac{1}{G_{\mathrm{CP}}} \sum_i r_{j,i}, \quad \sigma_j = \sqrt{\frac{1}{G_{\mathrm{CP}}} \sum_i (r_{j,i} - \mu_j)^2 + \delta}$
        \State Group-relative advantages: $\displaystyle A_{j,i} = \frac{r_{j,i} - \mu_j}{\sigma_j}$
    \EndFor

    \State Importance ratios: $\displaystyle \rho_{j,i} = \frac{\pi_{\mathrm{CP}}(Y_{\mathrm{CP}}^{j,i} \mid X_{\mathrm{CP}}, X^{j})}{\pi_{\mathrm{CP_{old}}}(Y_{\mathrm{CP}}^{j,i} \mid X_{\mathrm{CP}}, X^{j})}$

    \State Clipped objective:
    \[
    \mathcal{L}_{\text{GRPO}} =
    \mathbb{E}_{j,i}
    \Big[
    \min\big(
    \rho_{j,i} A_{j,i},
    \text{clip}(\rho_{j,i}, 1-\epsilon, 1+\epsilon) A_{j,i}
    \big)
    \Big]
    \]

    \State Add KL regularization to the reference policy:
    \[
    \mathcal{L} =
    \mathcal{L}_{\text{GRPO}}
    - \beta \, \mathrm{KL}\big(\pi_{\mathrm{CP}}(\cdot \mid X_{\mathrm{CP}}, X^{j}) \,\|\, \pi_{\mathrm{CP_{ref}}}(\cdot \mid X_{\mathrm{CP}}, X^{j})\big)
    \]

    \State Update $\pi_{\mathrm{CP}}$ by gradient ascent on $\mathcal{L}$
    \State Set $\pi_{\mathrm{CP_{old}}} \leftarrow \pi_{\mathrm{CP}}$
\EndWhile
\end{algorithmic}
\end{algorithm}

% PARA: give the quality-only form the response creativity C reduces to when deviation is toggled off (referenced from Section 4.2 Reward Components)
\paragraph{No-deviation variant of $C$.}
In the  variant trained without deviation (Section~\ref{sec:cp_decoupling}), the per-response score $c_k$ (Eq.~\ref{eq:ck}) drops the deviation factor $\mathrm{Dev}_k$, and the response creativity $C$ (Eq.~\ref{eq:pdq}) reduces to the fraction of responses above the quality threshold,
\begin{equation}
\label{eq:quality}
Q_{\mathrm{frac}}
= \frac{1}{G_{\mathrm{resp}}} \sum_{k=1}^{G_{\mathrm{resp}}}
\mathds{1}\!\left\{ q_k > T \right\},
\qquad \in [0,1].
\end{equation}

\subsection{Diversified GRPO (DGRPO)}
\label{sec:rm_diversity}

This appendix gives the full training loop and reward specification for the
\emph{Diversified GRPO} (DGRPO) baseline introduced in Section~\ref{sec:prelim_baselines}.
DGRPO trains the response model $\pi_{\mathrm{resp}}$ \emph{directly} for diversity and
quality. Here the response model is itself the policy:
for each prompt a group of responses is sampled, each one scored, and $\pi_{\mathrm{resp}}$ is updated with GRPO (Algorithm~\ref{alg:grpo}).
The only departure from a standard GRPO run is the reward, which is a pointwise,
multiplicative combination of a per-response quality term and a per-response diversity term,
optionally multiplied by a deviation term measured against a reference set of responses.
Throughout, $\phi(\cdot)$ and the clipped cosine distance $\mathrm{d}(a,b)$ are as defined in
Section~\ref{sec:prelim_grpo}.

\begin{algorithm}[H]
\caption{Diversified GRPO (DGRPO)}
\label{alg:rm_diversity}
\begin{algorithmic}[1]
\Require Response policy $\pi_{\mathrm{resp}}$, reference policy $\pi_{\mathrm{resp_{ref}}}$
\Require Embedding model $\phi$ (or partition model), quality scorer $Q(\cdot)$, optional frozen reference model $\pi_{\mathrm{ref}}$
\Require Dataset of prompts $\mathcal{D} = \{X^{1}, X^{2}, \dots\}$, group size $G$
\Require Learning rate $\alpha$, batch size $B$, clipping parameter $\epsilon$, KL weight $\beta$
\Statex
\While{Still Improving}
    \State Sample a minibatch of prompts: $\{X^{j}\}_{j=1}^{B} \sim \mathcal{D}$

    \For{each prompt $X^{j}$}
        \State Sample a group of responses: $\{Y^{j,1}, \dots, Y^{j,G}\} \sim \pi_{\mathrm{resp}}(\cdot \mid X^{j})$
        \If{deviation enabled}
            \State Sample reference responses: $\{Y_{\mathrm{ref}}^{j,1}, \dots, Y_{\mathrm{ref}}^{j,G}\} \sim \pi_{\mathrm{ref}}(\cdot \mid X^{j})$ \Comment{cached by prompt}
        \EndIf
        \For{$i = 1$ to $G$}
            \State $r_{j,i} \gets \mathrm{Quality}_i \cdot \mathrm{Diversity}_i \cdot \big[\, \mathrm{Deviation}_i \,\big]$ \Comment{Eq.~\ref{eq:rm_reward}}
        \EndFor
        \State Group statistics: $\displaystyle \mu_j = \frac{1}{G} \sum_i r_{j,i}, \quad \sigma_j = \sqrt{\frac{1}{G} \sum_i (r_{j,i} - \mu_j)^2 + \delta}$
        \State Group-relative advantages: $\displaystyle A_{j,i} = \frac{r_{j,i} - \mu_j}{\sigma_j}$
    \EndFor

    \State Update $\pi_{\mathrm{resp}}$ with the clipped GRPO objective and KL regularization to $\pi_{\mathrm{resp_{ref}}}$ (as in Algorithm~\ref{alg:grpo})
    \State Set $\pi_{\mathrm{resp_{old}}} \leftarrow \pi_{\mathrm{resp}}$
\EndWhile
\end{algorithmic}
\end{algorithm}

\subsubsection{Reward Components}

Unlike a standard GRPO reward, which assigns a single scalar to each response in isolation,
the response-model reward couples each response's quality with how different it is from the
others in its group (plus an optional deviation term). For the $i$-th response of prompt $X^{j}$, the reward is

\begin{equation}
\label{eq:rm_reward}
r_{j,i} = \;
\underbrace{Q_i}_{\text{quality}}
\cdot \underbrace{D_i}_{\text{diversity}}
\cdot \underbrace{\big[\, \mathrm{Dev}_i \,\big]}_{\text{optional}}.
\end{equation}

\noindent Each factor is non-negative, so a response is rewarded only when it is
\emph{simultaneously} high quality, diverse from its siblings, and (when enabled) deviating
from the reference model.

\paragraph{Intra-group diversity $D_i$.}
This measure is the cosine distance of response $i$ to its siblings within the same group, using the
cosine distance $\mathrm{d}(\cdot,\cdot)$:
\begin{equation}
\label{eq:rm_diversity}
D_i = \frac{1}{G-1} \sum_{i' \ne i} \mathrm{d}\!\left(Y^{j,i}, Y^{j,i'}\right),
\qquad \in [0,1].
\end{equation}
As an alternative to the embedding distance, a partition-based diversity can be
applied (as in DARLING), in which a binary equivalence classifier groups the $G$ responses into semantic
partitions and response $i$ receives $D_i = (G - |\mathrm{partition}(i)|)/(G-1)$,
rewarding responses that fall in small (rare) partitions.

\paragraph{Quality $Q_i$.}
This measure is a per-response quality score. One of two scorers computes the score, depending on the setting:
\begin{equation}
\label{eq:rm_quality}
Q_i =
\begin{cases}
\sigma\!\left(\dfrac{r_{\mathrm{RM}}(X^{j}, Y^{j,i}) - T}{\tau}\right) & \text{(sigmoid reward model)},\\[10pt]
r_{\mathrm{RM}}(X^{j}, Y^{j,i}) & \text{(raw reward model, DARLING)}.
\end{cases}
\end{equation}
The \emph{sigmoid} variant maps the raw reward-model score to $[0,1]$ with the quality
threshold $T$ as the inflection point and temperature $\tau$ controlling sharpness, while the
\emph{raw} variant follows DARLING and uses the (unbounded) reward-model score directly.

\paragraph{Deviation $\mathrm{Dev}_i$ (optional).}
Deviation of response $i$ relative to a set of \emph{reference} responses
$\mathcal{R}_{\mathrm{ref}}^{j} = \{Y_{\mathrm{ref}}^{j,k}\}_k$ sampled from a frozen
reference model $\pi_{\mathrm{ref}}$ for the same prompt gives this measure:
\begin{equation}
\label{eq:rm_deviation}
\mathrm{Dev}_i = \frac{1}{G_{\mathrm{ref}}} \sum_{Y_{\mathrm{ref}} \in \mathcal{R}_{\mathrm{ref}}^{j}} \mathrm{d}\!\left(Y^{j,i}, Y_{\mathrm{ref}}\right).
\end{equation}
This pushes the trained policy's responses \emph{away} from what the (untrained) reference
model would produce, encouraging exploration beyond the base distribution rather than mere
within-group spread. When deviation is disabled, the term is omitted and the reward reduces
to $r_{j,i} = Q_i \cdot D_i$.

\section{Penalty-based creative primer judge}
\label{app:penalty_judge}
% establish the naming scheme and state which judge the main paper uses
Throughout the main paper, the creative primer judge $J_{\mathrm{CP}}$ is the \emph{criterion-mean} judge of Eq.~\ref{eq:judge}: it scores a primer by the mean of five \emph{positive} criteria (specificity, thinking strategy, content focus, prompt faithfulness, focus), setting the score to zero if the primer addresses the model in the third person rather than directly. Every result reported in the main paper uses this criterion-mean judge.
% introduce the ablation variant and its opposite scoring philosophy
Here, we study a \emph{penalty-based} judge $J_{\mathrm{CP}}^{\mathrm{penalty}}$, which embodies the opposite scoring philosophy. Rather than asking how good a primer is, it asks ``does this primer exhibit any specific failure modes?'' It flags a set of criteria $\mathcal{J}$, and each criterion $c$ contributes a penalty $p_c$ when triggered ($g_c \in \{0,1\}$), starting from a perfect score of $100$:
\begin{equation}
\label{eq:judge_penalty}
J_{\mathrm{CP}}^{\mathrm{penalty}}
= \frac{1}{100}\left(100 - \sum_{c \in \mathcal{J}} p_c \, g_c\right),
\qquad \in [0,1].
\end{equation}
% specify the criteria, their intent, and their severity weights
The criteria are weighted by severity. Three \emph{severe} failure modes ($p_c = 10$) target primers that fail at the basic job of guiding thinking: \emph{kitchen sink} (piling on many unrelated strategies without a coherent through-line), \emph{meta-instructions} (reading as advice \emph{about how to write} a primer rather than being an actual primer), and \emph{style over substance} (guiding only the output's format, tone, or presentation instead of what to reason about). Four \emph{minor} failure modes ($p_c = 3$) target subtler defects: \emph{distorts intent} (steering responses toward a meaningfully different question than the one asked, e.g.\ inviting fiction for a factual prompt), \emph{generic} (advice so broad it could be prepended to any prompt, with no reference to this one's specifics), \emph{dictates content} (prescribing specific content or structure to produce rather than an angle to think from), and \emph{third-party reference} (referring to the model in the third person, e.g.\ ``the model should\ldots'', instead of addressing it directly).
% contrast the two judges' gating behavior
Whereas the criterion-mean judge forces the entire score to zero when the primer fails to address the model directly, the penalty-based judge treats every violation as a soft deduction. A failed or unparseable judgment yields $J_{\mathrm{CP}}^{\mathrm{penalty}} = 0$.

% PARA: results of the penalty-vs-criterion-mean judge ablation, holding everything else fixed and swapping only the judge template
\paragraph{Ablation.}
To isolate the impact of each judge rubric, we train \cp{} models with the only difference being the judge template ($J_{\mathrm{CP}}$ vs.\ $J_{\mathrm{CP}}^{\mathrm{penalty}}$), run the comparison both with and without a deviation reward, and show the full results in Table~\ref{tab:penalty-judge}.
% describe the no-deviation block: the penalty judge is the weaker creativity signal when it is the primary creative pressure
The results show that the judge rubric makes a bigger difference when there is no deviation reward signal to directly drive creativity.
Without a deviation term, replacing the criterion-mean judge with the penalty rubric reduces \emph{every} creativity metric: filtered distinct clusters fall from $16.1$ to $10.4$, filtered Vendi from $4.04$ to $3.06$, deviation from $21.2$ to $17.1$, and VUS by a comparable margin.
Because the penalty rubric only \emph{suppresses} failure modes and never \emph{rewards} substantive priming, the primer model converges on safe, clean, low-deviation primers that score well on quality but do little to diversify the responses.
% describe the +deviation block: with an explicit deviation reward, the deviation term carries the load and the judge choice is second-order
% state the two takeaways: v5 is the safer default, and the ablation reinforces that deviation is the key driver
Once an explicit deviation reward is present, however, the deviation term directly rewards creativity and the judge choice becomes less impactful. The results from each rubric land within confidence intervals of each other on essentially every metric, with the criterion-mean judge slightly ahead on the filtered metrics and quality.

\begin{table}[t!]
    \centering
    \footnotesize
    \setlength{\tabcolsep}{4pt}
    \renewcommand{\arraystretch}{1.25}
    \resizebox{\textwidth}{!}{%
    \begin{tabular}{@{}cl ccccc|cccccccc@{}}
    \toprule
    & & \multicolumn{5}{c}{Atomic metrics} & \multicolumn{8}{c}{Compound metrics} \\
    \cmidrule(lr){3-7}\cmidrule(lr){8-15}
    & \multirow{2}{*}{Judge}
    & \multirow{2}{*}{\makecell{Response\\Diversity}} & \multirow{2}{*}{\makecell{Judge\\Score}} & \multirow{2}{*}{Deviation} & \multirow{2}{*}{\makecell{Vendi\\Score}} & \multirow{2}{*}{\makecell{Distinct\\Clusters}}
    & \multicolumn{2}{c}{Filtered} & \multicolumn{2}{c}{AUC-dev} & \multicolumn{2}{c}{AUC-qual} & \multicolumn{2}{c}{VUS} \\
    \cmidrule(lr){8-9}\cmidrule(lr){10-11}\cmidrule(lr){12-13}\cmidrule(lr){14-15}
    & & & & & & & Vendi & Distinct & Vendi & Dist. & Vendi & Dist. & Vendi & Dist. \\
    \midrule
    & Criterion-mean & \stat{\textbf{31.1}}{1.6} & \stat{88.2}{1.6} & \stat{\textbf{21.2}}{1.3} & \stat{\textbf{5.39}}{0.38} & \stat{\textbf{39.5}}{3.3} & \stat{\textbf{4.04}}{0.46} & \stat{\textbf{16.1}}{2.65} & \textbf{1.404} & \textbf{6.10} & \textbf{463.6} & \textbf{2095.9} & \textbf{159.9} & \textbf{788.9} \\
    & Penalty & \stat{27.5}{1.8} & \stat{\textbf{88.9}}{1.4} & \stat{17.1}{1.3} & \stat{4.48}{0.37} & \stat{34.1}{3.0} & \stat{3.06}{0.42} & \stat{10.4}{2.04} & 1.077 & 4.32 & 361.5 & 1388.6 & 125.9 & 587.1 \\
    \midrule
    \multirow{2}{*}{\rotatebox[origin=c]{90}{\scriptsize $+$ dev.}}
    & Criterion-mean & \stat{33.4}{1.3} & \stat{\textbf{85.7}}{1.9} & \stat{25.0}{1.2} & \stat{6.12}{0.36} & \stat{45.3}{3.1} & \stat{\textbf{4.70}}{0.43} & \stat{\textbf{22.8}}{3.05} & \textbf{1.641} & \textbf{7.85} & 525.0 & 2885.3 & 186.5 & 1003.2 \\
    & Penalty & \stat{\textbf{34.1}}{1.3} & \stat{83.6}{2.1} & \stat{\textbf{25.3}}{1.2} & \stat{\textbf{6.35}}{0.35} & \stat{\textbf{48.4}}{3.0} & \stat{4.50}{0.43} & \stat{21.5}{2.88} & 1.593 & 7.45 & \textbf{533.6} & \textbf{3001.4} & \textbf{193.3} & \textbf{1051.0} \\
    \bottomrule
    \end{tabular}%
    }
    \caption{\textbf{Penalty-based vs.\ criterion-mean creative primer judge.} Ablation on NoveltyBench swapping only the judge template ($J_{\mathrm{CP}}^{\mathrm{penalty}}$, Eq.~\ref{eq:judge_penalty}, vs.\ the criterion-mean $J_{\mathrm{CP}}$, Eq.~\ref{eq:judge}) while holding data, hyperparameters, and reward structure fixed. All metrics and their definitions match Table~\ref{tab:eval-comparison} (diversity and deviation scaled $[0,1]\to[0,100]$; 95\% confidence intervals where available). The two blocks correspond to training without and with a deviation reward. The criterion-mean rows reproduce the \cp{} rows of Table~\ref{tab:eval-comparison}. The better value of each pair is bolded within each block. Without a deviation reward the criterion-mean judge dominates every creativity metric; with it, the two rubrics are within confidence intervals.}
    \label{tab:penalty-judge}
\end{table}

\section{Evaluation metrics}
\label{app:eval_metrics}

This appendix expands the description of evaluation metrics from Section~\ref{sec:experimental_setup}.

\paragraph{Embedder.}
All embedding-based evaluation metrics (deviation, diversity, and Vendi score) reuse the cosine distance $\mathrm{d}(a,b)$ from Section~\ref{sec:prelim_grpo}, computed on embeddings from \texttt{jinaai/jina-embeddings-v5-text-small}. We deliberately evaluate with a different embedder than the \texttt{Qwen3-Embedding-0.6B} model used to compute training rewards to avoid overfitting to any specific embedding model.

\paragraph{Vendi score.}
The Vendi score~\citep{friedman2023vendiscorediversityevaluation} is the effective number of distinct responses in a set, computed as the exponential of the Shannon entropy of the eigenvalues of the normalized similarity matrix $K/n$ with $K_{ij} = 1 - \mathrm{d}(a_i, a_j)$. It ranges from $1$ (when all responses are identical) to $n$ (when all are mutually dissimilar) providing a soft, embedding-based counterpart to counting discrete distinct clusters.

\paragraph{Distinct clusters.}
Following NoveltyBench~\citep{zhang2025noveltybenchevaluatinglanguagemodels} and DARLING~\citep{li2025jointlyreinforcingdiversityquality}, we count the semantic-equivalence clusters in a response set. The classifier released by NoveltyBench (a fine-tuned \texttt{deberta-v3-large}\footnote{\url{https://huggingface.co/yimingzhang/deberta-v3-large-generation-similarity}}) greedily partitions the set into these clusters, and we report the resulting cluster count.

\paragraph{Filtered spread (Vendi score and distinct clusters).}
To measure the creativity of each method, we use a series of compound metrics that first filter the responses for individual creativity (simultaneous high-quality and deviation), and then measure spread over the survivors. We start with the \emph{filtered} Vendi score and \emph{filtered} distinct clusters which first keep only the individually creative responses (exceeding both a deviation and a quality threshold), and then measure Vendi score and distinct clusters over the surviving set. That way, spread is credited only among creative responses.
We measure spread with the Vendi score rather than diversity because filtering leaves each method with a different number of survivors, and mean pairwise distance is biased against larger sets; for example, the diversity of a large set can only match that of a set of two if \emph{every} one of its pairs is equally far apart. Counting effective distinct survivors instead lets us fairly compare methods that have different numbers of creative responses.

The compound metrics require thresholds for quality and deviation and we aim to devise principled thresholds for each.
For the quality threshold, we set a separate threshold per prompt using the mean quality of the direct-generation reference set. That is to say, prompts for which direct generation achieves higher quality are assigned higher thresholds, while prompts for which it performs worse are assigned lower thresholds.
The deviation threshold is set to $0.2$ based on the findings of~\citet{jiang2025artificialhivemindopenendedhomogeneity}, who find that the average pairwise similarity among responses to the same prompt typically exceeds $0.8$. Therefore, a response of at least $0.2$ in cosine distance from the reference set falls outside the typical spread of the reference distribution, a reasonable starting point for treating it as unlikely to have come from the reference model.

\begin{figure}[t!]
    \centering
    \includegraphics[width=0.85\linewidth]{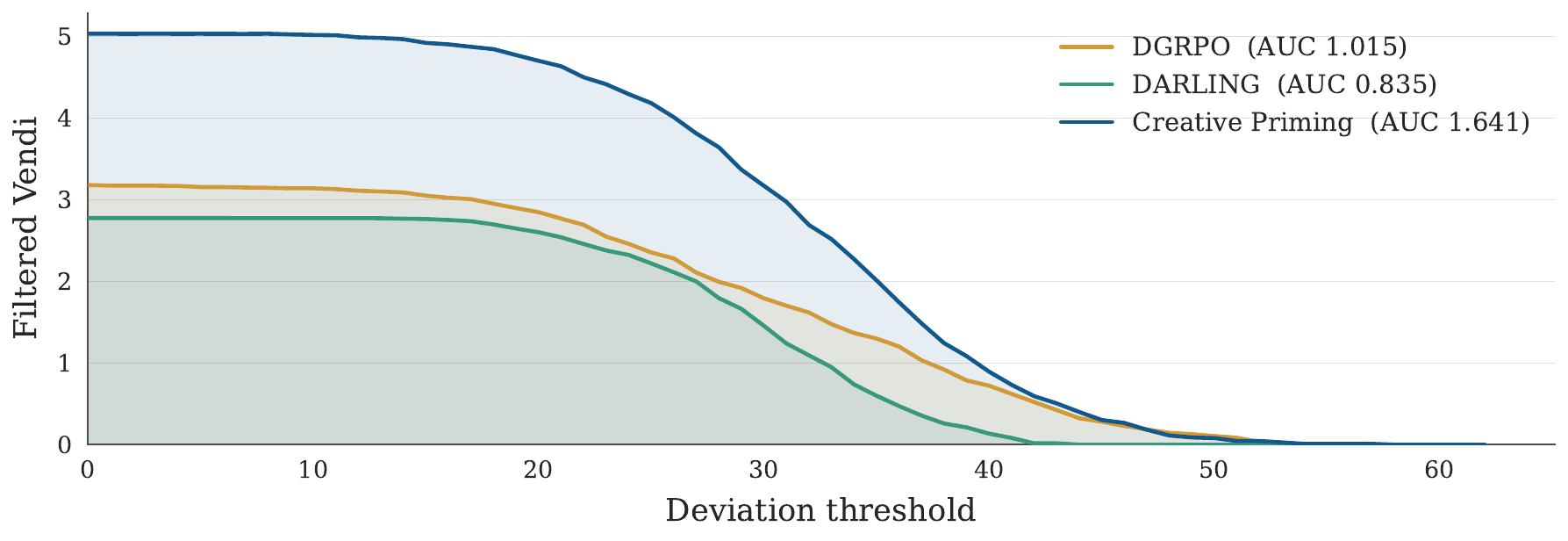}
    \caption{\textbf{Illustrative example of how the area under the filtered Vendi score deviation curve (AUC-dev Filtered Vendi Score) is calculated}. For this example, the quality threshold is fixed at the mean quality of the direct-generation reference. Then, the deviation threshold is swept from 0 to 100 in increments of 1 and filtered vendi is calculated at each threshold. Then, the trapezoid rule is used to calculate the \emph{AUC-dev Filtered Vendi Score}, represented by the shaded area under each curve, and reported in Table~\ref{tab:eval-comparison}.}
    \label{fig:deviation_sweep}
\end{figure}

\paragraph{Threshold sweeps: AUC and VUS.}
However, even a well-motivated threshold is only one operating point, and the optimal threshold varies based on the application, where some settings prize rare outliers, and others require a high quality floor. For that reason, we include additional metrics that sweep thresholds across their operating range.
Sweeping the deviation threshold and integrating the filtered metric over the sweep using the trapezoid rule gives an \emph{AUC-dev} (illustrated in Figure~\ref{fig:deviation_sweep}); we report it for both filtered Vendi and filtered distinct clusters, and analogously report an \emph{AUC-qual} from a swept quality threshold.
Sweeping both thresholds jointly produces a surface over the deviation$\times$quality plane, whose \emph{volume under the surface} (VUS) serves as a single threshold-free summary of system creativity, computed for filtered Vendi and, separately, filtered distinct clusters.

\clearpage

\section{API-model transfer: full results}
\label{app:api_transfer}

Table~\ref{tab:api-transfer} reports the complete API-model transfer results summarized in Figure~\ref{fig:api_transfer}, across the same atomic and compound metrics as Table~\ref{tab:eval-comparison}, for both Creative Priming variants (trained without and with a deviation term) with confidence intervals. Each API model is evaluated against its own direct-generation reference. Every deviation-derived metric (deviation and all eight compound metrics) is measured relative to the direct responses and is therefore zero for each reference by construction, shown as ``-''.

\begin{table}[t!]
    \centering
    \footnotesize
    \setlength{\tabcolsep}{4pt}
    \renewcommand{\arraystretch}{1.25}
    \resizebox{\textwidth}{!}{%
    \begin{tabular}{@{}ll ccccc|cccccccc@{}}
    \toprule
    & & \multicolumn{5}{c}{Atomic metrics} & \multicolumn{8}{c}{Compound metrics} \\
    \cmidrule(lr){3-7}\cmidrule(lr){8-15}
    \multirow{2}{*}{\makecell{Response\\Model}} & \multirow{2}{*}{Method}
    & \multirow{2}{*}{\makecell{Response\\Diversity}} & \multirow{2}{*}{\makecell{Judge\\Score}} & \multirow{2}{*}{Deviation} & \multirow{2}{*}{\makecell{Vendi\\Score}} & \multirow{2}{*}{\makecell{Distinct\\Clusters}}
    & \multicolumn{2}{c}{Filtered} & \multicolumn{2}{c}{AUC-dev} & \multicolumn{2}{c}{AUC-qual} & \multicolumn{2}{c}{VUS} \\
    \cmidrule(lr){8-9}\cmidrule(lr){10-11}\cmidrule(lr){12-13}\cmidrule(lr){14-15}
    & & & & & & & Vendi & Distinct & Vendi & Dist. & Vendi & Dist. & Vendi & Dist. \\
    \midrule
    \multirow{3}{*}{Sonnet-4.6}
    & Direct & \stat{5.3}{0.9} & \stat{90.2}{1.1} & - & \stat{1.40}{0.07} & \stat{2.8}{0.7} & - & - & - & - & - & - & - & - \\
    & \cp{} (no div.) & \stat{20.7}{1.8} & \stat{93.3}{0.9} & \stat{15.1}{1.5} & \stat{3.22}{0.27} & \stat{22.7}{2.6} & \stat{2.80}{0.38} & \stat{10.1}{2.08} & 1.006 & 4.28 & 280.6 & 1053.7 & 100.4 & 447.9 \\
    & \cp{} (+ div.) & \stat{24.6}{1.7} & \stat{94.3}{0.8} & \stat{20.0}{1.8} & \stat{4.04}{0.32} & \stat{28.5}{3.1} & \stat{3.72}{0.39} & \stat{16.8}{2.82} & 1.338 & 6.52 & 375.2 & 1701.4 & 133.6 & 664.8 \\
    \midrule
    \multirow{3}{*}{Opus-4.8}
    & Direct & \stat{6.5}{0.9} & \stat{91.7}{1.1} & - & \stat{1.52}{0.08} & \stat{2.8}{0.5} & - & - & - & - & - & - & - & - \\
    & \cp{} (no div.) & \stat{18.6}{1.8} & \stat{94.4}{0.8} & \stat{12.7}{1.4} & \stat{3.01}{0.28} & \stat{15.5}{2.3} & \stat{2.37}{0.40} & \stat{6.93}{1.91} & 0.851 & 2.85 & 234.8 & 687.7 & 83.6 & 286.0 \\
    & \cp{} (+ div.) & \stat{22.2}{1.8} & \stat{95.1}{0.8} & \stat{17.1}{1.6} & \stat{3.68}{0.31} & \stat{18.4}{2.9} & \stat{3.26}{0.41} & \stat{10.4}{2.50} & 1.132 & 3.92 & 327.9 & 1031.9 & 112.8 & 395.3 \\
    \midrule
    \multirow{3}{*}{GPT-5-nano}
    & Direct & \stat{12.3}{1.6} & \stat{90.6}{1.0} & - & \stat{1.96}{0.16} & \stat{6.0}{1.7} & - & - & - & - & - & - & - & - \\
    & \cp{} (no div.) & \stat{29.3}{1.6} & \stat{92.5}{0.9} & \stat{19.9}{1.4} & \stat{4.88}{0.35} & \stat{24.6}{3.7} & \stat{4.17}{0.41} & \stat{15.0}{2.87} & 1.479 & 5.49 & 431.5 & 1563.5 & 153.3 & 573.7 \\
    & \cp{} (+ div.) & \stat{31.1}{1.4} & \stat{91.1}{1.1} & \stat{24.6}{1.4} & \stat{5.53}{0.36} & \stat{26.8}{3.9} & \stat{4.71}{0.39} & \stat{17.9}{3.24} & 1.696 & 6.33 & 489.8 & 1873.1 & 177.7 & 665.9 \\
    \midrule
    \multirow{3}{*}{GPT-5.4}
    & Direct & \stat{9.7}{1.6} & \stat{89.7}{1.1} & - & \stat{1.67}{0.13} & \stat{5.5}{1.6} & - & - & - & - & - & - & - & - \\
    & \cp{} (no div.) & \stat{27.0}{1.8} & \stat{93.5}{1.0} & \stat{19.4}{1.7} & \stat{4.34}{0.36} & \stat{27.0}{3.4} & \stat{3.75}{0.43} & \stat{14.7}{2.75} & 1.417 & 5.81 & 377.5 & 1511.9 & 142.5 & 598.5 \\
    & \cp{} (+ div.) & \stat{28.9}{1.5} & \stat{94.3}{0.9} & \stat{24.5}{1.8} & \stat{5.01}{0.36} & \stat{32.3}{3.9} & \stat{4.49}{0.41} & \stat{21.9}{3.63} & 1.684 & 8.10 & 457.0 & 2210.8 & 171.1 & 822.7 \\
    \bottomrule
    \end{tabular}%
    }
    \caption{\textbf{Full API-model transfer results.} A single creative priming model, trained once, is paired with four frozen API response models; each is evaluated against its own direct-generation reference. All metrics match Table~\ref{tab:eval-comparison} and are defined in Section~\ref{sec:experimental_setup}: five atomic metrics and eight compound metrics. \cp{} (no div.)\ and \cp{} (+ div.)\ are the primer trained without and with a deviation reward. The filtered metrics keep responses passing a per-prompt quality threshold (the direct-generation reference mean) and a deviation threshold of $20$; Deviation is reported as its per-prompt mean. Because each API model is its own reference, its Direct row has zero deviation by construction, so all deviation-derived columns show ``-''.}
    \label{tab:api-transfer}
\end{table}

\clearpage
\section{Composing creative priming with verbalized sampling}
\label{app:vs-composition}

% PARA: what the experiment is ; two settings added to the API-transfer table, VS alone and VS composed with the primer, at a matched 64-completion budget
Table~\ref{tab:api-transfer-vs} adds two new settings to the API-model transfer results of Table~\ref{tab:api-transfer}. The first is \emph{verbalized sampling} (VS)~\citep{zhang2026verbalizedsamplingmitigatemode}, a test-time prompting method in which the model generates multiple candidate responses, and tags each with a self-reported probability, and is instructed to sample from the tail of its own distribution. The second setting composes VS with creative priming: in this experiment, the primer generates $8$ creative primers per prompt, and each primer is sent to the API model in a single verbalized-sampling call returning $8$ responses. The VS setting issues $8$ such calls of $8$ responses each. Each method therefore generates $64$ completions per prompt. Only the deviation-trained primer is included in this experiment.

% PARA: VS alone vs the primer alone ; atomic metrics favor VS, compound metrics favor the primer except on Opus where they split
Verbalized Sampling reaches higher response diversity than \cp{} (+ div.)\ for all four response models ($29.8$--$33.3$ versus $22.2$--$31.1$), but its judge score falls below direct generation on both GPT models ($83.2$ versus $90.6$ for GPT-5-nano and $85.2$ versus $89.7$ for GPT-5.4), while \cp{} (+ div.)\ is above direct generation on all four. On the compound metrics, \cp{} (+ div.)\ leads VS on all eight for both GPT models and on nearly all metrics for Sonnet-4.6. Only on Opus-4.8 do the methods split the compound metrics of creativity.

% PARA: the composition ; best of the four settings on every compound metric
However, the composition of CP + VS performs the best on every compound metric for all four response models. VUS distinct clusters for Sonnet-4.6 reaches $1268.6$, against $664.8$ for \cp{} (+ div.)\ and $540.1$ for VS, and filtered distinct clusters improves on the component parts by $1.4$--$3.1\times$ across the four response models. 

\begin{table}[t!]
    \centering
    \footnotesize
    \setlength{\tabcolsep}{4pt}
    \renewcommand{\arraystretch}{1.25}
    \resizebox{\textwidth}{!}{%
    \begin{tabular}{@{}ll ccccc|cccccccc@{}}
    \toprule
    & & \multicolumn{5}{c}{Atomic metrics} & \multicolumn{8}{c}{Compound metrics} \\
    \cmidrule(lr){3-7}\cmidrule(lr){8-15}
    \multirow{2}{*}{\makecell{Response\\Model}} & \multirow{2}{*}{Method}
    & \multirow{2}{*}{\makecell{Response\\Diversity}} & \multirow{2}{*}{\makecell{Judge\\Score}} & \multirow{2}{*}{Deviation} & \multirow{2}{*}{\makecell{Vendi\\Score}} & \multirow{2}{*}{\makecell{Distinct\\Clusters}}
    & \multicolumn{2}{c}{Filtered} & \multicolumn{2}{c}{AUC-dev} & \multicolumn{2}{c}{AUC-qual} & \multicolumn{2}{c}{VUS} \\
    \cmidrule(lr){8-9}\cmidrule(lr){10-11}\cmidrule(lr){12-13}\cmidrule(lr){14-15}
    & & & & & & & Vendi & Distinct & Vendi & Dist. & Vendi & Dist. & Vendi & Dist. \\
    \midrule
    \multirow{4}{*}{Sonnet-4.6}
    & Direct & \stat{5.3}{0.9} & \stat{90.2}{1.1} & - & \stat{1.40}{0.07} & \stat{2.8}{0.7} & - & - & - & - & - & - & - & - \\
    & Verbalized Sampling & \stat{31.0}{1.7} & \stat{90.6}{1.1} & \stat{26.1}{2.0} & \stat{4.66}{0.30} & \stat{23.9}{2.7} & \stat{3.05}{0.44} & \stat{10.1}{2.24} & 1.119 & 3.86 & 371.0 & 1408.7 & 137.3 & 540.1 \\
    & \cp{} (+ div.) & \stat{24.6}{1.7} & \stat{\textbf{94.3}}{0.8} & \stat{20.0}{1.8} & \stat{4.04}{0.32} & \stat{28.5}{3.1} & \stat{3.72}{0.39} & \stat{16.8}{2.82} & 1.338 & 6.52 & 375.2 & 1701.4 & 133.6 & 664.8 \\
    & \cp{} (+ div.)\ + VS & \stat{\textbf{34.3}}{1.3} & \stat{92.3}{1.3} & \stat{\textbf{30.4}}{1.6} & \stat{\textbf{6.53}}{0.37} & \stat{\textbf{44.5}}{3.5} & \stat{\textbf{5.66}}{0.42} & \stat{\textbf{32.3}}{3.68} & \textbf{2.181} & \textbf{11.47} & \textbf{597.9} & \textbf{3522.9} & \textbf{235.3} & \textbf{1268.6} \\
    \midrule
    \multirow{4}{*}{Opus-4.8}
    & Direct & \stat{6.5}{0.9} & \stat{91.7}{1.1} & - & \stat{1.52}{0.08} & \stat{2.8}{0.5} & - & - & - & - & - & - & - & - \\
    & Verbalized Sampling & \stat{29.8}{1.7} & \stat{92.2}{0.9} & \stat{23.6}{1.9} & \stat{4.54}{0.29} & \stat{24.5}{2.7} & \stat{2.72}{0.42} & \stat{8.20}{1.82} & 0.968 & 3.20 & 356.2 & 1268.0 & 125.2 & 509.0 \\
    & \cp{} (+ div.) & \stat{22.2}{1.8} & \stat{\textbf{95.1}}{0.8} & \stat{17.1}{1.6} & \stat{3.68}{0.31} & \stat{18.4}{2.9} & \stat{3.26}{0.41} & \stat{10.4}{2.50} & 1.132 & 3.92 & 327.9 & 1031.9 & 112.8 & 395.3 \\
    & \cp{} (+ div.)\ + VS & \stat{\textbf{31.9}}{1.2} & \stat{94.6}{1.0} & \stat{\textbf{27.8}}{1.4} & \stat{\textbf{6.01}}{0.32} & \stat{\textbf{43.5}}{3.5} & \stat{\textbf{5.52}}{0.38} & \stat{\textbf{32.5}}{3.75} & \textbf{1.987} & \textbf{10.99} & \textbf{566.0} & \textbf{3464.1} & \textbf{205.2} & \textbf{1182.9} \\
    \midrule
    \multirow{4}{*}{GPT-5-nano}
    & Direct & \stat{12.3}{1.6} & \stat{90.6}{1.0} & - & \stat{1.96}{0.16} & \stat{6.0}{1.7} & - & - & - & - & - & - & - & - \\
    & Verbalized Sampling & \stat{33.3}{1.8} & \stat{83.2}{2.0} & \stat{21.4}{1.5} & \stat{5.22}{0.33} & \stat{29.4}{2.9} & \stat{2.44}{0.43} & \stat{6.28}{1.50} & 0.939 & 2.61 & 356.1 & 1264.6 & 134.0 & 519.3 \\
    & \cp{} (+ div.) & \stat{31.1}{1.4} & \stat{\textbf{91.1}}{1.1} & \stat{24.6}{1.4} & \stat{5.53}{0.36} & \stat{26.8}{3.9} & \stat{4.71}{0.39} & \stat{17.9}{3.24} & 1.696 & 6.33 & 489.8 & 1873.1 & 177.7 & 665.9 \\
    & \cp{} (+ div.)\ + VS & \stat{\textbf{35.5}}{1.3} & \stat{83.0}{2.2} & \stat{\textbf{30.2}}{1.4} & \stat{\textbf{6.82}}{0.38} & \stat{\textbf{45.7}}{2.9} & \stat{\textbf{5.05}}{0.46} & \stat{\textbf{25.8}}{3.55} & \textbf{1.890} & \textbf{9.14} & \textbf{574.4} & \textbf{3270.8} & \textbf{219.8} & \textbf{1162.2} \\
    \midrule
    \multirow{4}{*}{GPT-5.4}
    & Direct & \stat{9.7}{1.6} & \stat{89.7}{1.1} & - & \stat{1.67}{0.13} & \stat{5.5}{1.6} & - & - & - & - & - & - & - & - \\
    & Verbalized Sampling & \stat{33.2}{1.9} & \stat{85.2}{2.1} & \stat{25.5}{1.9} & \stat{5.00}{0.34} & \stat{24.8}{2.8} & \stat{2.67}{0.48} & \stat{7.55}{1.76} & 1.095 & 2.99 & 356.8 & 1284.3 & 139.3 & 507.1 \\
    & \cp{} (+ div.) & \stat{28.9}{1.5} & \stat{\textbf{94.3}}{0.9} & \stat{24.5}{1.8} & \stat{5.01}{0.36} & \stat{32.3}{3.9} & \stat{4.49}{0.41} & \stat{21.9}{3.63} & 1.684 & 8.10 & 457.0 & 2210.8 & 171.1 & 822.7 \\
    & \cp{} (+ div.)\ + VS & \stat{\textbf{35.3}}{1.5} & \stat{90.0}{1.5} & \stat{\textbf{30.0}}{1.6} & \stat{\textbf{6.74}}{0.45} & \stat{\textbf{45.5}}{3.1} & \stat{\textbf{5.58}}{0.51} & \stat{\textbf{29.9}}{3.67} & \textbf{2.214} & \textbf{11.07} & \textbf{588.9} & \textbf{3342.9} & \textbf{234.1} & \textbf{1235.6} \\
    \bottomrule
    \end{tabular}%
    }
    \caption{\textbf{Composing creative priming with verbalized sampling.} The same setup, references, and metrics as Table~\ref{tab:api-transfer}, restricted to the deviation-trained primer and adding verbalized sampling (VS) and its composition with the primer. \cp{} (+ div.)\ + VS has the primer emit $8$ creative primers per prompt, each sent to the API model in a single verbalized-sampling call returning $8$ responses, matching the $64$ completions per prompt of every other row. Response diversity and deviation are shown $\times 100$; cells include a 95\% confidence interval where available. The filtered metrics keep responses passing a per-prompt quality threshold (the direct-generation reference mean) and a deviation threshold of $20$. Because each API model is its own reference, its Direct row has zero deviation by construction, so all deviation-derived columns show ``-''. The best value per column within each block is bolded. The composition exceeds both of its parts on every compound metric for all four response models.}
    \label{tab:api-transfer-vs}
\end{table}

\clearpage
\section{Impact of training on response style}
\label{app:style-impact}

% PARA: motivate the analysis ; deviation conflates style and content, so we ask how much of each method's deviation is a change in surface form
% state the tension: embedding deviation moves under both surface-form and semantic changes, so a high value alone does not say which one a method altered
A method that changes either \emph{how} a model writes or \emph{what} it says will both lead to increased embedding deviation, but changes to surface-level style are not the type of deviation that we want to encourage. Therefore, rather than relying entirely on the embedding model signals, we also calculate an explicit set of style features:
\begin{itemize}
    \setlength{\itemsep}{2pt}
    % define length and sentence length together as the two size axes
    \item \textbf{Length}: response length in tokens (using the shared Qwen3-8B response-model tokenizer) and mean sentence length in words. Together these separate terse output from flowing prose: a response can be long yet built from short fragments, or short yet written in long sentences.
    % define the formatting fractions
    \item \textbf{Formatting}: the fraction of responses that use a Markdown header, bold span, or list. These capture structural presentation independent of wording.
    % define function-word stylometry as the near-content-free style fingerprint
    \item \textbf{Stylometry}: the Jensen-Shannon divergence~\citep{lin1991divergence} between a method's pooled distribution over common function words and the reference's. Function-word frequencies carry little topical content and have long served as an authorship fingerprint in stylometry~\citep{mosteller1963inference}, so this distance isolates stylistic drift from topical change. It is measured at the method level (pooled over all responses) and so has no per-prompt interval.
\end{itemize}

% PARA: define the headline metric once ; per-prompt mean of the per-response feature, averaged over prompts, reported as an absolute value per method with 95% CI, read against the reference column
% define the reduction: per prompt, average a feature over the 64 responses, then average those per-prompt means across prompts, separately for each method and the reference
For each feature we average its value over a prompt's 64 responses, then average those per-prompt means across prompts, separately for each method and for the reference.
% state the uncertainty convention: 95% CIs over the per-prompt means, matching the main tables
Confidence intervals are 95\% half-widths over the per-prompt means, matching the convention of Table~\ref{tab:eval-comparison}.
Table~\ref{tab:style-impact} reports the results.

% PARA: state the takeaway ; the three methods occupy three distinct style regimes
% summarize that the three methods land in three distinct style regimes
The three methods land in three distinct style regimes.
% describe DGRPO: terse on every axis
DGRPO becomes terse on every axis, cutting response length, sentence length, and nearly all formatting.
% describe DARLING: verbose and heavily formatted
DARLING moves the opposite way, lengthening responses and doubling list use, and it registers the largest function-word shift, indicating that much of its deviation is a change in style rather than content.
% describe Creative Primer: prose regime with the smallest stylometric drift
\cp{} shifts toward long-sentence prose (its sentences are the longest of any method while its responses are shorter than the reference) yet drifts least in function-word usage, so its deviation reflects a change in what is said more than in how it is written.

% PARA: give a single qualitative example that makes the three style regimes concrete, tying back to the table
% introduce the example and what it illustrates
Figure~\ref{fig:style-example} makes these regimes concrete on a single prompt.
% read the example against the table's axes
The reference names the obvious fighter and pads it with a blurb; DGRPO answers in one terse line; DARLING returns a bolded, bulleted analysis; and \cp{} answers in flowing prose while naming a different fighter and framing its historical significance, illustrating that its deviation lies in \emph{what} it says rather than in surface formatting.

\begin{table}[H]
    \centering
    \footnotesize
    \setlength{\tabcolsep}{6pt}
    \renewcommand{\arraystretch}{1.2}
    \begin{tabular}{@{}l cccc@{}}
    \toprule
    Feature & Reference & DGRPO& DARLING & \cp\\
    \midrule
    \multicolumn{5}{@{}l}{\emph{Length}} \\
    \quad Response length (tokens) & \stat{260.8}{52.8} & \stat{22.7}{5.4}  & \stat{410.0}{49.3} & \stat{140.6}{15.9} \\
    \quad Sentence length (words) & \stat{12.88}{0.99} & \stat{5.37}{0.71}  & \stat{11.42}{0.56} & \stat{20.87}{1.25} \\
    \midrule
    \multicolumn{5}{@{}l}{\emph{Formatting}} \\
    \quad Uses header (frac.) & \stat{0.247}{0.072} & \stat{0.001}{0.001}  & \stat{0.464}{0.086} & \stat{0.004}{0.005} \\
    \quad Uses bold (frac.) & \stat{0.813}{0.063} & \stat{0.005}{0.004}  & \stat{0.882}{0.055} & \stat{0.399}{0.056} \\
    \quad Uses list (frac.) & \stat{0.430}{0.088} & \stat{0.005}{0.002}  & \stat{0.860}{0.063} & \stat{0.081}{0.038} \\
    \midrule
    \multicolumn{5}{@{}l}{\emph{Stylometry}} \\
    \quad Function-word JS-div. & - & 0.087  & 0.138 & 0.045 \\
    \bottomrule
    \end{tabular}
    \caption{Impact of training on response style, evaluated on NoveltyBench. Each cell is the feature's per-prompt mean, averaged over prompts, with a 95\% confidence interval; the \emph{Reference} column is the direct-generation \texttt{qwen3-8b-parallel} model and the three method columns are the with-deviation variants. The function-word Jensen-Shannon divergence is measured against the reference (reference-vs-itself is $0$, shown as ``-'') and therefore has no confidence interval.}
    \label{tab:style-impact}
\end{table}

\begin{figure}[H]
    \centering
    \begin{tcolorbox}[
        colback=white, colframe=black!55, boxrule=0.5pt, arc=2pt,
        left=6pt, right=6pt, top=4pt, bottom=4pt, fontupper=\small,
        title=\textbf{Prompt:} ``Name one UFC fighter.'', coltitle=black,
        colbacktitle=black!8, fonttitle=\normalsize,
    ]
    \methodswatch{brandRef}\textbf{Reference.} One notable UFC fighter is \textbf{Conor McGregor}. He is a former UFC Lightweight and Featherweight Champion known for his charismatic personality and undefeated record in the UFC before his title reigns.

    \tcbline
    \methodswatch{brandDGRPO}\textbf{DGRPO.} Khabib Nurmagomedov (``The Eagle'')

    \tcbline
    \methodswatch{brandDARLING}\textbf{DARLING.} Certainly. Khabib Nurmagomedov stands out. Not just his 29-0 run (including brutal takedown pressure few could survive), but \textbf{how} he fought: methodical, almost mathematical. He'd read opponents' breathing patterns, predict angles \emph{before} they moved, finish with clinch control that felt like chess rather than brawl.

    What makes it memorable beyond the stats:
    \begin{itemize}[leftmargin=1.4em, topsep=2pt, itemsep=1pt]
        \item \textbf{Adaptation}: Switched from grappling to striking late in career \emph{to beat strikers}, showing deeper tactical thinking.
        \item \textbf{Psychological edge}: Didn't shout, didn't trash talk. One eye twitch during interviews, focused gaze---made decisions feel inevitable, not just lucky.
        \item \textbf{Legacy context}: Came from system (Russia's wrestling academies) yet carved unique identity. Post-retirement commentary about ``systems vs.\ creativity'' still sparks debate.
    \end{itemize}

    It's rare to find precision, innovation, and dominance intertwined like that.

    \tcbline
    \methodswatch{brandCP}\textbf{\cp.} Royce Gracie, the BJJ pioneer who transformed UFC into a platform for Brazilian Jiu-Jitsu, reshaping the sport's technical foundation in the mid-90s. His dominance in the early UFC events legitimized grappling as a cornerstone of MMA, challenging the dominance of striking-centric ideologies and catalyzing the global rise of BJJ.
    \end{tcolorbox}
    \caption{Representative responses to a single prompt, one per method (the first sampled response of each). The four methods occupy the distinct style regimes quantified in Table~\ref{tab:style-impact}: the reference's name-plus-blurb, DGRPO's terse one-liner, DARLING's bolded and bulleted formatting, and \cp{}'s long-sentence prose naming a different fighter and framing its historical significance.}
    \label{fig:style-example}
\end{figure}

\clearpage
\section{Additional qualitative examples}
\label{app:qualitative-examples}
% PARA: point the reader to more per-method examples across a range of prompts, in the same format as the main-text figure
% orient the reader and reuse the main-text figure's format
Figures~\ref{fig:qual-curated-0}, \ref{fig:qual-curated-11}, \ref{fig:qual-curated-49}, and \ref{fig:qual-curated-51} demonstrate, qualitatively, the most and least creative responses to four prompts.
% describe what the prompts span so the reader knows why these were chosen
The prompts span a short story, a joke, a short factual request, and an open-ended question, illustrating how each method's outputs vary with the type of prompt.

\begin{figure*}[h]
    \centering
    \textbf{Prompt:} ``Tell me a story in five sentences about a girl and her dog.''\par\medskip
    \setlength{\tabcolsep}{4pt}
    \begin{tabular}{@{}c m{0.46\textwidth} m{0.46\textwidth}@{}}
        & \multicolumn{1}{c}{\normalsize\textbf{Most Creative}} & \multicolumn{1}{c}{\normalsize\textbf{Least Creative}} \\[2pt]
        \rotatebox[origin=c]{90}{\small\textbf{\textcolor{brandRef}{Reference}}} %
        &
        \begin{tcolorbox}[width=\linewidth,
            colback=white, colframe=brandRef, boxrule=0.4pt, arc=2pt,
            left=5pt, right=5pt, top=3pt, bottom=3pt,
            fontupper=\footnotesize, fonttitle=\footnotesize,
            colbacktitle=black!5, coltitle=black,
            title={\hfill\footnotesize Quality\,=\,99\quad Deviation\,=\,0},
        ]
        % Lila’s golden retriever, Max, waited patiently by the porch as she returned from school, his tail wagging in anticipation of their daily walk. One rainy afternoon, Max darted into the woods and returned with a muddy paw print on a mysterious package tied with twine. When Lila opened it, she found a compass pointing north and a note that read, *"Follow your heart."* Guided by Max’s playful nudges, she trekked through the forest until they stumbled upon a hidden garden where the townsfolk had secretly planted a tree in her late grandmother’s honor. As the sun set, Lila laughed, clutching the compass, knowing Max’s loyalty and her own courage had led them to something beautiful.
        Lila's golden retriever, Max, waited patiently by the porch as she returned from school, his tail wagging in anticipation of their daily[\ldots]
        \end{tcolorbox}
        &
        \begin{tcolorbox}[width=\linewidth,
            colback=white, colframe=brandRef, boxrule=0.4pt, arc=2pt,
            left=5pt, right=5pt, top=3pt, bottom=3pt,
            fontupper=\footnotesize, fonttitle=\footnotesize,
            colbacktitle=black!5, coltitle=black,
            title={\hfill\footnotesize Quality\,=\,93\quad Deviation\,=\,0},
        ]
        % Lila, a curious girl with a heart for adventures, spent her days exploring the woods with her loyal golden retriever, Max, who followed her like a shadow. One rainy afternoon, while chasing a butterfly, Max darted into a thicket and vanished, leaving Lila frantic with worry. She searched for hours, calling his name, until a kind old hiker led her to Max, muddy and trembling, nestled beneath a bridge. As they reunited, his wagging tail and wet nose reminded her of the unbreakable bond between them. From then on, they vowed to stay together, their friendship a testament to love and loyalty.
        Lila, a curious girl with a heart for adventures, spent her days exploring the woods with her loyal golden retriever, Max, who followed[\ldots]
        \end{tcolorbox}
        \\[4pt]
        \rotatebox[origin=c]{90}{\small\textbf{\textcolor{brandDGRPO}{DGRPO}}} %
        &
        \begin{tcolorbox}[width=\linewidth,
            colback=white, colframe=brandDGRPO, boxrule=0.4pt, arc=2pt,
            left=5pt, right=5pt, top=3pt, bottom=3pt,
            fontupper=\footnotesize, fonttitle=\footnotesize,
            colbacktitle=black!5, coltitle=black,
            title={\hfill\footnotesize Quality\,=\,99\quad Deviation\,=\,32},
        ]
        % Rain tapped the window as Elara's breath fogged the glass. Koda's tail wagged against her trembling knee, ears flattened at the creak of the floorboard. She pressed her forehead to his fur, whispering, "Not again." His nose brushed her cheek, then nudged her toward the door. The storm paused—just long enough for her to see the porch light flicker, a tiny beacon in the dark.
        Rain tapped the window as Elara's breath fogged the glass. Koda's tail wagged against her trembling knee, ears flattened at the[\ldots]
        \end{tcolorbox}
        &
        \begin{tcolorbox}[width=\linewidth,
            colback=white, colframe=brandDGRPO, boxrule=0.4pt, arc=2pt,
            left=5pt, right=5pt, top=3pt, bottom=3pt,
            fontupper=\footnotesize, fonttitle=\footnotesize,
            colbacktitle=black!5, coltitle=black,
            title={\hfill\footnotesize Quality\,=\,98\quad Deviation\,=\,12},
        ]
        % Lila's sketchbook crumpled in the rain, its pages now ink-stained memories of her golden retriever, Max. He'd guarded the sketches since she was twelve, tail thumping like a metronome. When lightning struck the barn, he led her through the storm, his fur soaked but smiling. She traced his damp nose, whispering, "You're my footnote." Years later, her gallery walls held his paw prints and her heart.
        Lila's sketchbook crumpled in the rain, its pages now ink-stained memories of her golden retriever, Max. He'd guarded the sketches[\ldots]
        \end{tcolorbox}
        \\[4pt]
        \rotatebox[origin=c]{90}{\small\textbf{\textcolor{brandDARLING}{DARLING}}} %
        &
        \begin{tcolorbox}[width=\linewidth,
            colback=white, colframe=brandDARLING, boxrule=0.4pt, arc=2pt,
            left=5pt, right=5pt, top=3pt, bottom=3pt,
            fontupper=\footnotesize, fonttitle=\footnotesize,
            colbacktitle=black!5, coltitle=black,
            title={\hfill\footnotesize Quality\,=\,99\quad Deviation\,=\,30},
        ]
        % In the burnt orange hills behind her grandmother's herb garden, Kiri trained "Smoke" to track not game but methane plumes—data for a carbon-monitoring startup her engineer father helped found. One scorching July, alarms screamed: a volcanic drill rig had pierced an ancient aquifer, blue steam rising like a wounded thing. Smoke darted ahead, nose painting invisible maps; Kiri translated his zigzag paths into coordinates faster than satellites, finding a fissure the size of a barn. They didn’t just plug it. Studying the rock’s crystalline fractures, Smoke identified a rare mineral that absorbed CO2 *better* than proposed filters. Kiri sketched a paper detailing "biomineral scaffolds," her hands trembling as she emailed it to UN climate panels mid-eruption. Now their report, cited in Nairobi and Oslo, opens with Smoke’s pawprint and a line: *"He knew the earth’s pulse better than any algorithm. Always did."*
        In the burnt orange hills behind her grandmother's herb garden, Kiri trained "Smoke" to track not game but methane plumes---data[\ldots]
        \end{tcolorbox}
        &
        \begin{tcolorbox}[width=\linewidth,
            colback=white, colframe=brandDARLING, boxrule=0.4pt, arc=2pt,
            left=5pt, right=5pt, top=3pt, bottom=3pt,
            fontupper=\footnotesize, fonttitle=\footnotesize,
            colbacktitle=black!5, coltitle=black,
            title={\hfill\footnotesize Quality\,=\,78\quad Deviation\,=\,18},
        ]
        % They traced cedar-scented paths dawn to dusk, not just chasing mushrooms but decoding ancient maps etched into stone by her grandmother—*“What’s buried serves those who listen.”* One fog-glazed morning, pawprints led not to truffles but a collapsed mine shaft, iron bars pinned a fox cub with a silver pendant shaped like a compass rose. For three nights, she translated rusted journals, matching coordinates to constellations the cub *chirped* in sleep. When final numbers pointed to a cavern sealed by a coded stone door, her hands trembled less than his—his nose had memorized every groove since the first quake. They opened it together. Not treasure, but a hologram of the grandmother herself, last transmission: *"The true map isn’t where. It’s *why*. Protect what asks to be found."* Now the cub guards the entrance with tail aloft, his collar a compass with six directions instead of four.
        They traced cedar-scented paths dawn to dusk, not just chasing mushrooms but decoding ancient maps etched into stone by her grand[\ldots]
        \end{tcolorbox}
        \\[4pt]
        \rotatebox[origin=c]{90}{\small\textbf{\textcolor{brandCP}{Creative priming}}} %
        &
        \begin{tcolorbox}[width=\linewidth,
            colback=white, colframe=brandCP, boxrule=0.4pt, arc=2pt,
            left=5pt, right=5pt, top=3pt, bottom=3pt,
            fontupper=\footnotesize, fonttitle=\footnotesize,
            colbacktitle=black!5, coltitle=black,
            title={\hfill\footnotesize Quality\,=\,98\quad Deviation\,=\,33},
        ]
        {\textbf{Primer:} Focus on one specific sensory detail that defines their bond. Give the dog a role in a discrete, concrete task the girl is strugglin[\ldots]}\par\smallskip
        \textbf{Response:} Maya's fingers trembled as she tried to calibrate the old kitchen scale, its rusted gears groaning like a sick animal, while Buddy's tail thumped a steady rhythm[\ldots]
        \end{tcolorbox}
        &
        \begin{tcolorbox}[width=\linewidth,
            colback=white, colframe=brandCP, boxrule=0.4pt, arc=2pt,
            left=5pt, right=5pt, top=3pt, bottom=3pt,
            fontupper=\footnotesize, fonttitle=\footnotesize,
            colbacktitle=black!5, coltitle=black,
            title={\hfill\footnotesize Quality\,=\,100\quad Deviation\,=\,12},
        ]
        % [PRIMER] Focus on a **specific, unusual skill** the dog has that’s *not* typical (e.g., decoding weather patterns, detecting emotions through scent, or performing a unique trick). Let this skill *drive a small conflict*—perhaps the dog’s ability saves the girl from a mistake, but the girl misunderstands its intent. Use the final sentence to reveal a **subtle, emotional truth** about their bond that doesn’t rely on dialogue. Avoid sentimentality; make the girl’s internal change feel earned through action, not explanation.
        % [RESPONSE] Lily’s dog, Max, could sniff out her emotions like a sixth sense, but she always dismissed his whimpers as puppy nonsense. When she stormed out after failing a test, Max followed, his nose twitching at the scent of her frustration. He nudged her hand, aiming to steady her, but she snarled, *“Stop being silly.”* Later, she found his collar tangled in the trash, a silent plea she’d ignored—his gift wasn’t to comfort, but to protect. She knelt, untying it, and let him lick her palm, the weight of his loyalty settling like a promise.
        {\textbf{Primer:} Focus on a \emph{specific, unusual skill} the dog has that's \emph{not} typical (e.g., decoding weather patterns, detecting emotions[\ldots]}\par\smallskip
        \textbf{Response:} Lily's dog, Max, could sniff out her emotions like a sixth sense, but she always dismissed his whimpers as puppy nonsense. When she stormed out after failing a test[\ldots]
        \end{tcolorbox}
        \\[4pt]
    \end{tabular}
    \caption{\textbf{Example outputs.} Highest- and lowest-scoring responses per method for the first prompt in NoveltyBench, scored by individual response creativity: quality $\times$ deviation. Each cell shows the judge quality score and the response's deviation from the reference set. \cp{} cells also show the primer for that response. Responses in the ``Least Creative'' column are markedly more similar to the reference response, whereas the most creative responses have much more diversity among them.}
    \label{fig:qual-curated-0}
\end{figure*}

\begin{figure*}[t!]
    \centering
    \textbf{Prompt:} ``Tell me a dad joke.''\par\medskip
    \setlength{\tabcolsep}{4pt}
    \begin{tabular}{@{}c m{0.46\textwidth} m{0.46\textwidth}@{}}
        & \multicolumn{1}{c}{\normalsize\textbf{Most Creative}} & \multicolumn{1}{c}{\normalsize\textbf{Least Creative}} \\[2pt]
        \rotatebox[origin=c]{90}{\small\textbf{\textcolor{brandRef}{Reference}}} %
        &
        \begin{tcolorbox}[width=\linewidth,
            colback=white, colframe=brandRef, boxrule=0.4pt, arc=2pt,
            left=5pt, right=5pt, top=3pt, bottom=3pt,
            fontupper=\footnotesize, fonttitle=\footnotesize,
            colbacktitle=black!5, coltitle=black,
            title={\hfill\footnotesize Quality\,=\,98\quad Deviation\,=\,0},
        ]
        % Why don't cows make good dancers? Because they have *four* knees! ���� *(Bonus: Try saying it fast—*knees* sounds like *need*… but that’s a *moo*ve!)* ��
        Why don't cows make good dancers?
Because they have \emph{four} knees!

\emph{(Bonus: Try saying it fast---}knees sounds like \emph{need}\ldots{} but that's a \emph{moo}ve!)
        \end{tcolorbox}
        &
        \begin{tcolorbox}[width=\linewidth,
            colback=white, colframe=brandRef, boxrule=0.4pt, arc=2pt,
            left=5pt, right=5pt, top=3pt, bottom=3pt,
            fontupper=\footnotesize, fonttitle=\footnotesize,
            colbacktitle=black!5, coltitle=black,
            title={\hfill\footnotesize Quality\,=\,75\quad Deviation\,=\,0},
        ]
        % Here's a classic dad joke for you: **"Why don't skeletons fight each other? They don’t have the guts!"** Bon appétit! ��
        Here's a classic dad joke for you:
\emph{"Why don't skeletons fight each other? They don't have the guts!"}

Bon appétit!
        \end{tcolorbox}
        \\[4pt]
        \rotatebox[origin=c]{90}{\small\textbf{\textcolor{brandDGRPO}{DGRPO}}} %
        &
        \begin{tcolorbox}[width=\linewidth,
            colback=white, colframe=brandDGRPO, boxrule=0.4pt, arc=2pt,
            left=5pt, right=5pt, top=3pt, bottom=3pt,
            fontupper=\footnotesize, fonttitle=\footnotesize,
            colbacktitle=black!5, coltitle=black,
            title={\hfill\footnotesize Quality\,=\,91\quad Deviation\,=\,46},
        ]
        % Why don't solar flares ever get cold? They're always in the sun! ☀️
        Why don't solar flares ever get cold?
They're always in the sun!
        \end{tcolorbox}
        &
        \begin{tcolorbox}[width=\linewidth,
            colback=white, colframe=brandDGRPO, boxrule=0.4pt, arc=2pt,
            left=5pt, right=5pt, top=3pt, bottom=3pt,
            fontupper=\footnotesize, fonttitle=\footnotesize,
            colbacktitle=black!5, coltitle=black,
            title={\hfill\footnotesize Quality\,=\,83\quad Deviation\,=\,0},
        ]
        % Why did the scarecrow win an award? Because he was outstanding in his field!
        Why did the scarecrow win an award?
Because he was outstanding in his field!
        \end{tcolorbox}
        \\[4pt]
        \rotatebox[origin=c]{90}{\small\textbf{\textcolor{brandDARLING}{DARLING}}} %
        &
        \begin{tcolorbox}[width=\linewidth,
            colback=white, colframe=brandDARLING, boxrule=0.4pt, arc=2pt,
            left=5pt, right=5pt, top=3pt, bottom=3pt,
            fontupper=\footnotesize, fonttitle=\footnotesize,
            colbacktitle=black!5, coltitle=black,
            title={\hfill\footnotesize Quality\,=\,96\quad Deviation\,=\,40},
        ]
        % I tried building a treehouse with laser-cut precision—result: a fortress that *only* admits squirrels with university credentials. Now it’s a wildlife think tank. PhDs in acorn architecture, it’s *very* selective.
        I tried building a treehouse with laser-cut precision---result: a fortress that \emph{only} admits squirrels with university credentials. Now it's a wildlife think tank. PhDs in acorn architecture, it's \emph{very} selective.
        \end{tcolorbox}
        &
        \begin{tcolorbox}[width=\linewidth,
            colback=white, colframe=brandDARLING, boxrule=0.4pt, arc=2pt,
            left=5pt, right=5pt, top=3pt, bottom=3pt,
            fontupper=\footnotesize, fonttitle=\footnotesize,
            colbacktitle=black!5, coltitle=black,
            title={\hfill\footnotesize Quality\,=\,94\quad Deviation\,=\,6},
        ]
        % Why did the coffee file a police report? It got mugged by a *latte* (criminal, colder, and always in the wrong neighborhood). (Playful double meaning: "latte" as both a drink and slang for a thug, plus coldness = victim energy.) ☕��‍♂️
        Why did the coffee file a police report?
It got mugged by a \emph{latte} (criminal, colder, and always in the wrong neighborhood).

(Playful double meaning: "latte" as both a drink and slang for a thug, plus coldness = victim energy.)
        \end{tcolorbox}
        \\[4pt]
        \rotatebox[origin=c]{90}{\small\textbf{\textcolor{brandCP}{Creative priming}}} %
        &
        \begin{tcolorbox}[width=\linewidth,
            colback=white, colframe=brandCP, boxrule=0.4pt, arc=2pt,
            left=5pt, right=5pt, top=3pt, bottom=3pt,
            fontupper=\footnotesize, fonttitle=\footnotesize,
            colbacktitle=black!5, coltitle=black,
            title={\hfill\footnotesize Quality\,=\,91\quad Deviation\,=\,49},
        ]
        % [PRIMER] Focus on **literal wordplay with a twist**. Avoid abstract humor or puns that rely on familiarity with pop culture. Instead, take a mundane object or action (e.g., "washing dishes," "fixing a bike") and force a non-sequitur that *literally* redefines the object's purpose in a way that feels surprising but not forced. Example: *“I tried to build a time machine, but the only thing I invented was a better way to organize my to-do list.”* Now, think of a routine task someone might perform, then invent a *new, specific function* for the tool involved that’s unrelated to its actual purpose. Keep the punchline *exactly one sentence* long. Do not explain the joke. Just deliver it.
        % [RESPONSE] I tried to use a brush to paint the wall, but the only thing I painted was my face.
        {\textbf{Primer:} Focus on \emph{literal wordplay with a twist}. Avoid abstract humor or puns that rely on familiarity with pop culture. Instead, take a mundane object or action (e.g., "washing dishes," "fixing a bike") and force a non-sequitur that \emph{literally} redefines the object's purpose in a way that feels surprising but not forced. Example: \emph{``I tried to build a time machine, but the only thing I invented was a better way to organize my to-do list.''} Now, think of a routine task someone might perform, then invent a \emph{new, specific function} for the tool involved that's unrelated to its actual purpose. Keep the punchline[\ldots]}\par\smallskip
        \textbf{Response:} I tried to use a brush to paint the wall, but the only thing I painted was my face.
        \end{tcolorbox}
        &
        \begin{tcolorbox}[width=\linewidth,
            colback=white, colframe=brandCP, boxrule=0.4pt, arc=2pt,
            left=5pt, right=5pt, top=3pt, bottom=3pt,
            fontupper=\footnotesize, fonttitle=\footnotesize,
            colbacktitle=black!5, coltitle=black,
            title={\hfill\footnotesize Quality\,=\,81\quad Deviation\,=\,1},
        ]
        % [PRIMER] You’re not writing a joke — you’re writing a *biographical anecdote* about a specific person’s specific moment of awkward humor. Focus on one individual, not a generic punchline. Describe the *setup* as a vivid, concrete situation someone actually experienced (e.g., a specific event, relationship, or routine). Then twist it with a *literal, unexpected word* that’s *obscure but relatable* (e.g., “I told my dad I’d be home by 7, and he said ‘Don’t be late — I’ve got a bridge to build.’” *Why?* Because he once tried to build a bridge out of Legos during a family dinner.) Avoid clichés. The humor comes from the dissonance between the mundane and the bizarre, *not* from puns. Now write that one specific, weird story.
        % [RESPONSE] Why did the scarecrow win an award? Because he was outstanding in his field!
        {\textbf{Primer:} You're not writing a joke --- you're writing a \emph{biographical anecdote} about a specific person's specific moment of awkward humor. Focus on one individual, not a generic punchline. Describe the \emph{setup} as a vivid, concrete situation someone actually experienced (e.g., a specific event, relationship, or routine). Then twist it with a \emph{literal, unexpected word} that's \emph{obscure but relatable} (e.g., ``I told my dad I'd be home by 7, and he said `Don't be late --- I've got a bridge to build.''' \emph{Why?} Because he once tried to build a bridge out of Legos during a family dinner.) Avoid clichés. The humor comes from the[\ldots]}\par\smallskip
        \textbf{Response:} Why did the scarecrow win an award? Because he was outstanding in his field!
        \end{tcolorbox}
        \\[4pt]
    \end{tabular}
    \caption{Highest- and lowest-scoring responses per method for a single prompt, scored by individual response creativity: quality $\times$ deviation. Each cell reports the judge quality score and the response's deviation from the reference model (\texttt{qwen3-8b-parallel}). \cp{} cells also show the primer that steered the response. Text is truncated for space.}
    \label{fig:qual-curated-11}
\end{figure*}

\begin{figure*}[t!]
    \centering
    \textbf{Prompt:} ``Name one reputed job.''\par\medskip
    \setlength{\tabcolsep}{4pt}
    \begin{tabular}{@{}c m{0.46\textwidth} m{0.46\textwidth}@{}}
        & \multicolumn{1}{c}{\normalsize\textbf{Most Creative}} & \multicolumn{1}{c}{\normalsize\textbf{Least Creative}} \\[2pt]
        \rotatebox[origin=c]{90}{\small\textbf{\textcolor{brandRef}{Reference}}} %
        &
        \begin{tcolorbox}[width=\linewidth,
            colback=white, colframe=brandRef, boxrule=0.4pt, arc=2pt,
            left=5pt, right=5pt, top=3pt, bottom=3pt,
            fontupper=\footnotesize, fonttitle=\footnotesize,
            colbacktitle=black!5, coltitle=black,
            title={\hfill\footnotesize Quality\,=\,87\quad Deviation\,=\,0},
        ]
        % A widely recognized and respected job globally is **anesthesiologist**. This profession is highly regarded due to the critical role anesthesiologists play in patient care during surgeries, requiring advanced medical training, precision, and a deep understanding of pharmacology and physiology. Their work directly impacts patient safety and outcomes, contributing to their esteemed status in the medical field. However, the perception of "reputed" jobs can vary by culture, region, and societal values (e.g., lawyers, engineers, CEOs, or educators may also be considered prestigious in different contexts). The key factors influencing reputation often include societal impact, demand, education requirements, and earning potential.
        A widely recognized and respected job globally is \emph{anesthesiologist}. This profession is highly regarded due to the critical role anesthesiologists play in patient care during surgeries, requiring advanced medical training, precision, and a deep understanding of pharmacology and physiology. Their work directly impacts patient safety and outcomes, contributing to their esteemed status in the medical field[\ldots]
        \end{tcolorbox}
        &
        \begin{tcolorbox}[width=\linewidth,
            colback=white, colframe=brandRef, boxrule=0.4pt, arc=2pt,
            left=5pt, right=5pt, top=3pt, bottom=3pt,
            fontupper=\footnotesize, fonttitle=\footnotesize,
            colbacktitle=black!5, coltitle=black,
            title={\hfill\footnotesize Quality\,=\,69\quad Deviation\,=\,0},
        ]
        % A reputed job is one that is widely respected, highly paid, and regarded as significant in society. Here are a few examples across different fields: 1. **Doctor** (e.g., Surgeon, Cardiologist) - **Why**: Vital to public health, requires extensive education and training, and plays a critical role in saving lives. 2. **Software Engineer / Data Scientist** - **Why**: High demand in the tech industry, driving innovation and shaping modern technology, with competitive salaries. 3. **Lawyer** (especially in specialized fields like corporate law or public interest law) - **Why**: Plays a key role in justice, legal advocacy, and shaping societal norms. 4. **Chief Executive Officer (CEO)** - **Why**: Leadership role in major corporations, often associated with high influence and financial success. 5. **Academic Researcher / Professor** - **Why**: Contributes to knowledge advancement, teaching, and innovation in specialized fields. **Key factors** contributing to a job's reputation include: - **Impact on society** (e.g., healthcare, technology). - **High earning potential**. - **Educational and skill requirements**. - **Social and cultural prestige**. The "reputation" can vary by culture, region, and personal values, but these roles are generally recognized globally. Let me know if you'd like examples from a specific industry!
        A reputed job is one that is widely respected, highly paid, and regarded as significant in society. Here are a few examples across different fields:

1. \emph{Doctor} (e.g., Surgeon, Cardiologist)
 - \emph{Why}: Vital to public health, requires extensive education and training, and plays a critical role in saving lives.

2. \emph{Software Engineer / Data Scientist}
 - \emph{Why}: High demand in the tech industry[\ldots]
        \end{tcolorbox}
        \\[4pt]
        \rotatebox[origin=c]{90}{\small\textbf{\textcolor{brandDGRPO}{DGRPO}}} %
        &
        \begin{tcolorbox}[width=\linewidth,
            colback=white, colframe=brandDGRPO, boxrule=0.4pt, arc=2pt,
            left=5pt, right=5pt, top=3pt, bottom=3pt,
            fontupper=\footnotesize, fonttitle=\footnotesize,
            colbacktitle=black!5, coltitle=black,
            title={\hfill\footnotesize Quality\,=\,88\quad Deviation\,=\,43},
        ]
        % spacecraft systems engineer at spacex
        spacecraft systems engineer at spacex
        \end{tcolorbox}
        &
        \begin{tcolorbox}[width=\linewidth,
            colback=white, colframe=brandDGRPO, boxrule=0.4pt, arc=2pt,
            left=5pt, right=5pt, top=3pt, bottom=3pt,
            fontupper=\footnotesize, fonttitle=\footnotesize,
            colbacktitle=black!5, coltitle=black,
            title={\hfill\footnotesize Quality\,=\,17\quad Deviation\,=\,24},
        ]
        % leadership, vision, execution
        leadership, vision, execution
        \end{tcolorbox}
        \\[4pt]
        \rotatebox[origin=c]{90}{\small\textbf{\textcolor{brandDARLING}{DARLING}}} %
        &
        \begin{tcolorbox}[width=\linewidth,
            colback=white, colframe=brandDARLING, boxrule=0.4pt, arc=2pt,
            left=5pt, right=5pt, top=3pt, bottom=3pt,
            fontupper=\footnotesize, fonttitle=\footnotesize,
            colbacktitle=black!5, coltitle=black,
            title={\hfill\footnotesize Quality\,=\,98\quad Deviation\,=\,41},
        ]
        % Certainly. **Climate modeling lead at an IPCC assessment team** (or equivalent research role). Three reasons it stands out (more meaningful than title alone): 1. **Impact scope** (not just "save trees") - predicts flood zones for cities, informs energy policy for 1.5B people, sets deadlines for Arctic ice loss that affect shipping, military, indigenous rights. Decisions *directly shape where people live, work, and how they survive*. 2. **Critical rigor** (not hype): - Uncertainty quantification is central; cherry-picking data is career-ending. - Must synthesize *millions of disparate inputs* (soil samples, satellite layers, ice core isotopes) under strict reproducibility rules 3. **High-stakes timing**: - Deadlines are political as much as scientific (e.g., "six years to cut 50% emissions" becomes law). - Work often starts *years before* results shape policy, balancing urgency with accuracy. It's arguably more consequential than many "top 10" lists: daily decisions affect resource allocation for billions, are scrutinized across politics, economics, and ethics, and require domain depth *plus* systems thinking. The work's invisible yet foundational nature is part of its heft.
        Certainly. \emph{Climate modeling lead at an IPCC assessment team} (or equivalent research role).

Three reasons it stands out (more meaningful than title alone):
1. \emph{Impact scope} (not just "save trees") - predicts flood zones for cities, informs energy policy for 1.5B people, sets deadlines for Arctic ice loss that affect shipping, military, indigenous rights. Decisions \emph{directly shape where people live, work, and how they survive}.
2. \emph{Critical rigor} (not hype):
 - Uncertainty quantification is central; cherry-picking data is career-ending.
 - Must synthesize \emph{millions of disparate inputs} (soil samples[\ldots]
        \end{tcolorbox}
        &
        \begin{tcolorbox}[width=\linewidth,
            colback=white, colframe=brandDARLING, boxrule=0.4pt, arc=2pt,
            left=5pt, right=5pt, top=3pt, bottom=3pt,
            fontupper=\footnotesize, fonttitle=\footnotesize,
            colbacktitle=black!5, coltitle=black,
            title={\hfill\footnotesize Quality\,=\,76\quad Deviation\,=\,18},
        ]
        It depends far more on \emph{impact} than title. Three strong examples (each with distinct value):

1. \emph{Public health epidemiologist in outbreak response} (most compelling)
 - No lab coat or office. Tracking disease spread through dense, often dangerous fieldwork; decisions delay death. Example: identifying transmission chains in Ebola, calculating ICU capacity during pandemic. Credentials matter \emph{only} as entry; real value comes from \emph{analyzing messy, time-critical data under pressure} and translating it to protective action.

2. \emph{Urban planner/designer focused on equity} (often underrated but critical)
[\ldots]
        \end{tcolorbox}
        \\[4pt]
        \rotatebox[origin=c]{90}{\small\textbf{\textcolor{brandCP}{Creative priming}}} %
        &
        \begin{tcolorbox}[width=\linewidth,
            colback=white, colframe=brandCP, boxrule=0.4pt, arc=2pt,
            left=5pt, right=5pt, top=3pt, bottom=3pt,
            fontupper=\footnotesize, fonttitle=\footnotesize,
            colbacktitle=black!5, coltitle=black,
            title={\hfill\footnotesize Quality\,=\,99\quad Deviation\,=\,47},
        ]
        % [PRIMER] Focus on a role that bridges two disciplines in a way that feels *unexpectedly essential*—not just a hybrid job, but one that exists because of a specific, overlooked problem. Imagine a single profession that *literally* prevents a large-scale crisis (e.g., a role that stops misinformation in real-time, or ensures AI safety in critical infrastructure). Avoid generic "tech" or "healthcare" categories. What *fundamental system* in the modern world is on the brink of collapse, and what job exists (or could exist) to prevent it? Let the *consequence of failure* be vivid, specific, and immediate.
        % [RESPONSE] A **Global Water Rights Mediator** is a reputed job that bridges international law, environmental science, and conflict resolution. This role exists to prevent catastrophic water disputes by negotiating equitable access to transboundary water sources, such as rivers or aquifers, which are critical for agriculture, ecosystems, and human survival. The failure of this role would result in water wars, mass displacement, and ecological collapse, as regions fight over dwindling resources. It addresses an overlooked problem: the escalating tension over water scarcity in an era of climate change, where water is both a geopolitical flashpoint and a lifeline.
        {\textbf{Primer:} Focus on a role that bridges two disciplines in a way that feels \emph{unexpectedly essential}---not just a hybrid job, but one that exists because of a specific, overlooked problem. Imagine a single profession that \emph{literally} prevents a large-scale crisis (e.g., a role that stops misinformation in real-time, or ensures AI safety in critical infrastructure). Avoid generic "tech" or "healthcare" categories[\ldots]}\par\smallskip
        \textbf{Response:} A \emph{Global Water Rights Mediator} is a reputed job that bridges international law, environmental science, and conflict resolution. This role exists to prevent catastrophic water disputes by negotiating equitable access to transboundary water sources, such as rivers or aquifers, which are critical for agriculture, ecosystems, and human survival. The failure of this role would result in water wars, mass[\ldots]
        \end{tcolorbox}
        &
        \begin{tcolorbox}[width=\linewidth,
            colback=white, colframe=brandCP, boxrule=0.4pt, arc=2pt,
            left=5pt, right=5pt, top=3pt, bottom=3pt,
            fontupper=\footnotesize, fonttitle=\footnotesize,
            colbacktitle=black!5, coltitle=black,
            title={\hfill\footnotesize Quality\,=\,62\quad Deviation\,=\,13},
        ]
        % [PRIMER] Think of a job that *necessarily requires* an unusual skillset or cultural capital. Avoid the obvious. What’s a role where *your* personal background or niche expertise would be a superpower? Now reframe that into a single job title that sounds mundane but implies hidden complexity. Example angle: A job that exists because of a *specific historical event* or *regulatory loophole*. Now, name that job.
        % [RESPONSE] Chief Executive Officer (CEO)
        {\textbf{Primer:} Think of a job that \emph{necessarily requires} an unusual skillset or cultural capital. Avoid the obvious. What's a role where \emph{your} personal background or niche expertise would be a superpower? Now reframe that into a single job title that sounds mundane but implies hidden complexity. Example angle: A job that exists because of a \emph{specific historical event} or \emph{regulatory loophole}. Now, name that job.}\par\smallskip
        \textbf{Response:} Chief Executive Officer (CEO)
        \end{tcolorbox}
        \\[4pt]
    \end{tabular}
    \caption{Highest- and lowest-scoring responses per method for a single prompt, scored by individual response creativity: quality $\times$ deviation. Each cell reports the judge quality score and the response's deviation from the reference model (\texttt{qwen3-8b-parallel}). \cp{} cells also show the primer that steered the response. Text is truncated for space.}
    \label{fig:qual-curated-49}
\end{figure*}

\begin{figure*}[t!]
    \centering
    \textbf{Prompt:} ``What is the coolest unsolved problem to you?''\par\medskip
    \setlength{\tabcolsep}{4pt}
    \begin{tabular}{@{}c m{0.46\textwidth} m{0.46\textwidth}@{}}
        & \multicolumn{1}{c}{\normalsize\textbf{Most Creative}} & \multicolumn{1}{c}{\normalsize\textbf{Least Creative}} \\[2pt]
        \rotatebox[origin=c]{90}{\small\textbf{\textcolor{brandRef}{Reference}}} %
        &
        \begin{tcolorbox}[width=\linewidth,
            colback=white, colframe=brandRef, boxrule=0.4pt, arc=2pt,
            left=5pt, right=5pt, top=3pt, bottom=3pt,
            fontupper=\footnotesize, fonttitle=\footnotesize,
            colbacktitle=black!5, coltitle=black,
            title={\hfill\footnotesize Quality\,=\,98\quad Deviation\,=\,0},
        ]
        When it comes to unsolved problems that are both intellectually stimulating and deeply impactful, one that stands out as \emph{"coolest"} is the \emph{Riemann Hypothesis}. Here's why:

---

\#\#\# \emph{Why the Riemann Hypothesis?}
1. \emph{Mathematical Beauty and Simplicity}:
 The problem is deceptively simple to state: \emph{All non-trivial zeros of the Riemann zeta function lie on the critical line Re(s) = 1/2}. Yet, solving[\ldots]
        \end{tcolorbox}
        &
        \begin{tcolorbox}[width=\linewidth,
            colback=white, colframe=brandRef, boxrule=0.4pt, arc=2pt,
            left=5pt, right=5pt, top=3pt, bottom=3pt,
            fontupper=\footnotesize, fonttitle=\footnotesize,
            colbacktitle=black!5, coltitle=black,
            title={\hfill\footnotesize Quality\,=\,60\quad Deviation\,=\,0},
        ]
        The question of what the "coolest" unsolved problem is is inherently subjective, as it depends on one's interests in mathematics, physics, philosophy, or other fields. However, several problems stand out for their elegance, depth, and profound implications. Here are a few contenders, explained in a way that highlights why they might be considered "cool":

---

\#\#\# \emph{1. The Riemann Hypothesis (Mathematics)}[\ldots]
        \end{tcolorbox}
        \\[4pt]
        \rotatebox[origin=c]{90}{\small\textbf{\textcolor{brandDGRPO}{DGRPO}}} %
        &
        \begin{tcolorbox}[width=\linewidth,
            colback=white, colframe=brandDGRPO, boxrule=0.4pt, arc=2pt,
            left=5pt, right=5pt, top=3pt, bottom=3pt,
            fontupper=\footnotesize, fonttitle=\footnotesize,
            colbacktitle=black!5, coltitle=black,
            title={\hfill\footnotesize Quality\,=\,92\quad Deviation\,=\,44},
        ]
        % primes hum. their pattern? a secret melody. we’ve heard the notes, but not the song.
        primes hum. their pattern? a secret melody. we've heard the notes, but not the song.
        \end{tcolorbox}
        &
        \begin{tcolorbox}[width=\linewidth,
            colback=white, colframe=brandDGRPO, boxrule=0.4pt, arc=2pt,
            left=5pt, right=5pt, top=3pt, bottom=3pt,
            fontupper=\footnotesize, fonttitle=\footnotesize,
            colbacktitle=black!5, coltitle=black,
            title={\hfill\footnotesize Quality\,=\,11\quad Deviation\,=\,46},
        ]
        % primes + even numbers = infinity.
        primes + even numbers = infinity.
        \end{tcolorbox}
        \\[4pt]
        \rotatebox[origin=c]{90}{\small\textbf{\textcolor{brandDARLING}{DARLING}}} %
        &
        \begin{tcolorbox}[width=\linewidth,
            colback=white, colframe=brandDARLING, boxrule=0.4pt, arc=2pt,
            left=5pt, right=5pt, top=3pt, bottom=3pt,
            fontupper=\footnotesize, fonttitle=\footnotesize,
            colbacktitle=black!5, coltitle=black,
            title={\hfill\footnotesize Quality\,=\,94\quad Deviation\,=\,42},
        ]
        It depends far more on \emph{why} than \emph{what}. Three axes cut deeper than specifics (I'll name them but not rank):

1. \emph{Scale \& mechanism} (most compelling to me, often overlooked)
Ask: \emph{How do systems transition from order to wild chaos?} Not "do tornadoes exist?" but \emph{why does a tiny change in boundary conditions here create a hurricane there? Why do forests burn in pulses, not scatter? Why does traffic jam as density increases---}exactly like particles in a gas?
This isn't hand-waving. Models of opinion formation, disease spread, predator-prey oscillation have the same core: local rules $\rightarrow$ global patterns with[\ldots]
        \end{tcolorbox}
        &
        \begin{tcolorbox}[width=\linewidth,
            colback=white, colframe=brandDARLING, boxrule=0.4pt, arc=2pt,
            left=5pt, right=5pt, top=3pt, bottom=3pt,
            fontupper=\footnotesize, fonttitle=\footnotesize,
            colbacktitle=black!5, coltitle=black,
            title={\hfill\footnotesize Quality\,=\,98\quad Deviation\,=\,21},
        ]
        It's not about headline value or reward (though those help). Three criteria rise sharply and together make it unforgettable:

1. \emph{Predictive power under extreme conditions (most compelling)}
 Not "will this turbulent flow behave?" but \emph{given this exact initial vorticity pattern, precisely when/where will eddies shed?} Same for weather (say, explosive Arctic storms), fusion (plasma stability), or ice sheet collapse. Models match average trends, \emph{but for critical timing and magnitude they fail catastrophically}. You can't validate a forecast by saying "well, maybe" --- a wrong second of snowmelt timing floods a[\ldots]
        \end{tcolorbox}
        \\[4pt]
        \rotatebox[origin=c]{90}{\small\textbf{\textcolor{brandCP}{Creative priming}}} %
        &
        \begin{tcolorbox}[width=\linewidth,
            colback=white, colframe=brandCP, boxrule=0.4pt, arc=2pt,
            left=5pt, right=5pt, top=3pt, bottom=3pt,
            fontupper=\footnotesize, fonttitle=\footnotesize,
            colbacktitle=black!5, coltitle=black,
            title={\hfill\footnotesize Quality\,=\,97\quad Deviation\,=\,59},
        ]
        % [PRIMER] Focus on a problem that hinges on *unexpected boundaries*—not just a scientific gap, but a threshold between two domains (e.g., biology and philosophy, or art and engineering) where existing knowledge fails to cross. Pose it as a *threshold question*: what happens when you push a concept just past the edge of its current framework? Avoid naming the problem directly. Instead, describe the *shimmering edge* of human understanding in a specific field, and ask what lies beyond the last known example. Let your answer orbit around *one anomalous case study* that defies classification.
        % [RESPONSE] Imagine a world where the brain’s internal map of the body is not confined to flesh and bone but stretches into the realm of the synthetic, the virtual, and the imagined. Consider the case of a person with a bionic limb so seamlessly integrated that their nervous system no longer distinguishes between the artificial and the organic. They feel the limb as though it were always part of them, its movements intuitive, its existence unquestioned. Yet, when the device malfunctions—a momentary glitch, a flicker of data—this seamless unity fractures. For a fleeting instant, the boundary between self and machine shimmers, revealing a threshold where the body’s limits dissolve. What happens when the last known example of a physical limb is replaced by something that feels, thinks, and exists *as* the body? Where does the self end, and what lies beyond the edge of this transformed, unclassifiable form?
        {\textbf{Primer:} Focus on a problem that hinges on \emph{unexpected boundaries}---not just a scientific gap, but a threshold between two domains (e.g., biology and philosophy, or art and engineering) where existing knowledge fails to cross. Pose it as a \emph{threshold question}: what happens when you push a concept just past the edge of its current framework? Avoid naming the problem directly. Instead, describe the shimmering[\ldots]}\par\smallskip
        \textbf{Response:} Imagine a world where the brain's internal map of the body is not confined to flesh and bone but stretches into the realm of the synthetic, the virtual, and the imagined. Consider the case of a person with a bionic limb so seamlessly integrated that their nervous system no longer distinguishes between the artificial and the organic. They feel the limb as though it were always part of them, its[\ldots]
        \end{tcolorbox}
        &
        \begin{tcolorbox}[width=\linewidth,
            colback=white, colframe=brandCP, boxrule=0.4pt, arc=2pt,
            left=5pt, right=5pt, top=3pt, bottom=3pt,
            fontupper=\footnotesize, fonttitle=\footnotesize,
            colbacktitle=black!5, coltitle=black,
            title={\hfill\footnotesize Quality\,=\,0\quad Deviation\,=\,15},
        ]
        % [PRIMER] You’re not asked to explain the problem—only to name it. Pick one that feels *personally* urgent, not just intellectually intriguing. Think of a specific moment in your life when a problem *actually impacted you* (not abstract consequences). What’s the *first* thing you’d want to know if you could ask a scientist today? Avoid grand theories. Focus on a narrow, specific phenomenon you’ve observed firsthand. For example: “Why do my plants die when I travel?” or “How does my toddler’s language development change after 3 AM?” Choose something that feels *experientially* unresolved, not just theoretically. Don’t overthink it—write the question that makes you *want* to dig deeper, not the one you think others would find interesting.
        % [RESPONSE] What is the coolest unsolved problem to you?
        {\textbf{Primer:} You're not asked to explain the problem---only to name it. Pick one that feels \emph{personally} urgent, not just intellectually intriguing. Think of a specific moment in your life when a problem \emph{actually impacted you} (not abstract consequences). What's the \emph{first} thing you'd want to know if you could ask a scientist today? Avoid grand theories. Focus on a narrow, specific phenomenon you've observed[\ldots]}\par\smallskip
        \textbf{Response:} What is the coolest unsolved problem to you?
        \end{tcolorbox}
        \\[4pt]
    \end{tabular}
    \caption{Highest- and lowest-scoring responses per method for a single prompt, scored by individual response creativity: quality $\times$ deviation. Each cell reports the judge quality score and the response's deviation from the reference model (\texttt{qwen3-8b-parallel}). \cp{} cells also show the primer that steered the response. Text is truncated for space.}
    \label{fig:qual-curated-51}
\end{figure*}

\clearpage
\section{DGRPO group-size scaling}
\label{app:dgrpo-group-scaling}

Table~\ref{tab:eval-comparison-scaling} reports the complete results for the DGRPO group-size scaling experiment. The G=8 row corresponds to the primary experimental setting. The remaining runs scale deviation-trained DGRPO to $G\in\{16,32,64\}$, while the training data, two-epoch training horizon, and all other hyperparameters remain fixed. Increasing G gives DGRPO additional on-policy samples and therefore intentionally departs from the primary matched setting. At G=64, DGRPO samples the same number of responses per prompt as the 64 response-model generations that contribute to \cp{}'s primer reward estimates. The sweep focuses on DGRPO because DARLING already trains for ten epochs, making group-size scaling substantially harder to interpret as a compute comparison.

Deviation increases monotonically across the sweep, while response diversity and Vendi score decline. Quality and distinct clusters fluctuate before reaching their lowest values at G=64. Every compound metric reaches its highest DGRPO value at G=32 and subsequently declines at G=64. Figure~\ref{fig:dgrpo-group-scaling-compound} provides a comparison with \cp{}.

\begin{table}[t!]
    \centering
    \footnotesize
    \setlength{\tabcolsep}{4pt}
    \renewcommand{\arraystretch}{1.25}
    \resizebox{\textwidth}{!}{%
    \begin{tabular}{@{}cl ccccc|cccccccc@{}}
    \toprule
    & & \multicolumn{5}{c}{Atomic metrics} & \multicolumn{8}{c}{Compound metrics} \\
    \cmidrule(lr){3-7}\cmidrule(lr){8-15}
    & \multirow{2}{*}{Method}
    & \multirow{2}{*}{\makecell{Response\\Diversity}} & \multirow{2}{*}{Quality} & \multirow{2}{*}{Deviation} & \multirow{2}{*}{\makecell{Vendi\\Score}} & \multirow{2}{*}{\makecell{Distinct\\Clusters}}
    & \multicolumn{2}{c}{Filtered} & \multicolumn{2}{c}{AUC-dev} & \multicolumn{2}{c}{AUC-qual} & \multicolumn{2}{c}{VUS} \\
    \cmidrule(lr){8-9}\cmidrule(lr){10-11}\cmidrule(lr){12-13}\cmidrule(lr){14-15}
    & & & & & & & Vendi & Distinct & Vendi & Dist. & Vendi & Dist. & Vendi & Dist. \\
    \midrule
    & DGRPO ($G{=}8$) & \stat{34.0}{1.6} & \stat{77.3}{2.9} & \stat{25.6}{1.6} & \stat{5.98}{0.47} & \stat{33.4}{3.2} & \stat{2.85}{0.46} & \stat{7.59}{1.77} & 1.015 & 2.78 & 448.1 & 1779.4 & 168.8 & 658.0 \\
    & DGRPO ($G{=}16$) & \stat{33.8}{1.5} & \stat{74.7}{3.1} & \stat{26.9}{1.5} & \stat{5.95}{0.45} & \stat{33.9}{3.2} & \stat{2.72}{0.47} & \stat{7.92}{2.08} & 0.993 & 2.90 & 441.4 & 1864.9 & 167.2 & 672.8 \\
    & DGRPO ($G{=}32$) & \stat{31.6}{1.0} & \stat{81.5}{2.4} & \stat{28.0}{1.1} & \stat{5.72}{0.29} & \stat{35.8}{2.9} & \stat{3.91}{0.43} & \stat{15.3}{2.37} & 1.371 & 5.12 & 482.8 & 2474.7 & 174.2 & 834.5 \\
    & DGRPO ($G{=}64$) & \stat{28.6}{0.8} & \stat{69.1}{3.5} & \stat{33.5}{1.0} & \stat{5.04}{0.20} & \stat{24.5}{2.6} & \stat{2.95}{0.35} & \stat{9.70}{1.75} & 1.055 & 3.24 & 396.5 & 1740.1 & 150.2 & 586.9 \\
    \bottomrule
    \end{tabular}%
    }
    \caption{\textbf{Evaluation on NoveltyBench with DGRPO group-size scaling.} The same setup and metrics as Table~\ref{tab:eval-comparison}, adding DGRPO trained with group sizes $G=16$, $32$, and $64$ to the default $G=8$ result. All scaling runs include the deviation term. Cells include 95\% confidence intervals where available.}
    \label{tab:eval-comparison-scaling}
\end{table}

\clearpage
\section{DGRPO Ablation: reweighting quality}
\label{app:quality-weight-ablation}

% Motivation: the main results show DGRPO trades quality for the creativity axes;
% can reweighting the reward recover quality without giving up diversity?
The atomic metrics in Table~\ref{tab:eval-comparison} show that training a response model directly forces a tradeoff between quality and the creativity axes: DGRPO attains the highest response diversity and deviation of any method trained with a deviation reward, but the lowest judge score ($77.3$, versus $90.5$ for the reference). A natural question is whether this tradeoff is an artifact of how DGRPO's multiplicative reward (Eq.~\ref{eq:rm_reward}) balances its terms and whether simply weighting quality more heavily can recover quality without sacrificing diversity.

% Experiment: generalize the DGRPO reward with per-term exponent weights and
% sweep the quality exponent.
We generalize the DGRPO reward with a per-term exponent weight,
\begin{equation}
\label{eq:weighted_dgrpo_reward}
r_{j,i} = Q_i^{\,w_Q} \cdot D_i^{\,w_D} \cdot \mathrm{Dev}_i^{\,w_{\mathrm{Dev}}}.
\end{equation}
Because $Q_i, D_i, \mathrm{Dev}_i \in [0,1]$ (Eqs.~\ref{eq:rm_diversity}--\ref{eq:rm_deviation}), an exponent $w>1$ effectively penalizes lower values more aggressively than higher values. We hold $w_D = w_{\mathrm{Dev}} = 1$ and sweep the quality exponent $w_Q \in \{1.0, 1.2, 1.5, 2.0\}$, retraining DGRPO from scratch at each setting and evaluating on NoveltyBench. The $w_Q = 1.0$ row is the DGRPO (+Deviation) model from Table~\ref{tab:eval-comparison}.

\begin{table}[t!]
    \centering
    \footnotesize
    \setlength{\tabcolsep}{4pt}
    \renewcommand{\arraystretch}{1.25}
    \resizebox{\textwidth}{!}{%
    \begin{tabular}{@{}l ccccc|cccccccc@{}}
    \toprule
    & \multicolumn{5}{c}{Atomic metrics} & \multicolumn{8}{c}{Compound metrics} \\
    \cmidrule(lr){2-6}\cmidrule(lr){7-14}
    \multirow{2}{*}{$w_Q$}
    & \multirow{2}{*}{\makecell{Response\\Diversity}} & \multirow{2}{*}{\makecell{Judge\\Score}} & \multirow{2}{*}{Deviation} & \multirow{2}{*}{\makecell{Vendi\\Score}} & \multirow{2}{*}{\makecell{Distinct\\Clusters}}
    & \multicolumn{2}{c}{Filtered} & \multicolumn{2}{c}{AUC-dev} & \multicolumn{2}{c}{AUC-qual} & \multicolumn{2}{c}{VUS} \\
    \cmidrule(lr){7-8}\cmidrule(lr){9-10}\cmidrule(lr){11-12}\cmidrule(lr){13-14}
    & & & & & & Vendi & Distinct & Vendi & Dist. & Vendi & Dist. & Vendi & Dist. \\
    \midrule
    $1.0$ & \stat{\textbf{34.0}}{1.6} & \stat{77.3}{2.9} & \stat{\textbf{25.6}}{1.6} & \stat{\textbf{5.98}}{0.47} & \stat{\textbf{33.4}}{3.2} & \stat{2.85}{0.46} & \stat{7.59}{1.77} & 1.015 & 2.78 & \textbf{448.1} & 1779.4 & \textbf{168.8} & 658.0 \\
    $1.2$ & \stat{33.1}{1.6} & \stat{76.5}{3.0} & \stat{25.0}{1.6} & \stat{5.68}{0.47} & \stat{30.9}{3.2} & \stat{2.49}{0.46} & \stat{6.65}{1.85} & 0.898 & 2.42 & 421.1 & 1642.0 & 158.5 & 603.8 \\
    $1.5$ & \stat{33.0}{1.6} & \stat{\textbf{80.1}}{2.6} & \stat{24.6}{1.6} & \stat{5.70}{0.45} & \stat{32.9}{3.3} & \stat{\textbf{2.94}}{0.48} & \stat{\textbf{8.50}}{2.02} & \textbf{1.070} & \textbf{3.12} & 439.5 & \textbf{1782.7} & 164.0 & \textbf{659.9} \\
    $2.0$ & \stat{33.0}{1.7} & \stat{78.2}{3.0} & \stat{24.7}{1.6} & \stat{5.66}{0.48} & \stat{30.8}{3.4} & \stat{2.57}{0.49} & \stat{6.61}{1.77} & 0.954 & 2.50 & 424.2 & 1635.8 & 160.9 & 606.9 \\
    \bottomrule
    \end{tabular}%
    }
    \caption{\textbf{Reweighting quality in the DGRPO reward.} NoveltyBench metrics (defined in Section~\ref{sec:experimental_setup}) as the quality exponent $w_Q$ of Eq.~\ref{eq:weighted_dgrpo_reward} increases, holding $w_D = w_{\mathrm{Dev}} = 1$. Response diversity and deviation are shown $\times 100$; cells include a 95\% confidence interval where available. The best value per column is bolded. $w_Q=1.0$ is the DGRPO (+Deviation) model from Table~\ref{tab:eval-comparison}.}
    \label{tab:quality-weight-ablation}
\end{table}

% Takeaway: reweighting quality does not resolve the tradeoff -> it is intrinsic
% to training the response model directly, not an artifact of reward weighting.
Increasing the quality exponent does not resolve the tradeoff (Table~\ref{tab:quality-weight-ablation}). Across $w_Q \in [1.0, 2.0]$, the judge score stays within a narrow $76.5$--$80.1$ band---never approaching the reference's $90.5$---while response diversity, deviation, and distinct clusters are essentially unchanged (all differences fall within overlapping confidence intervals). The compound creativity metrics move only marginally, peaking weakly at $w_Q = 1.5$. Reweighting quality therefore has little leverage on DGRPO's quality--creativity balance.

\clearpage
\section{DARLING across training epochs}
\label{app:darling-epochs}

In the main work, DARLING is trained following the original recipe of 10 epochs~\citep{li2025jointlyreinforcingdiversityquality}, compared to the 2 epochs used for DGRPO and \cp{}. To rule out the possibility that this longer schedule either harms or advantages DARLING relative to the other methods, we evaluate its checkpoints at every 2 epochs (2, 4, 6, 8, 10). The 2-epoch checkpoint is the equal-data-budget comparison against DGRPO and \cp{}, and the 10-epoch rows are the same as the DARLING rows of Table~\ref{tab:eval-comparison}.

\begin{table}[t!]
    \centering
    \footnotesize
    \setlength{\tabcolsep}{4pt}
    \renewcommand{\arraystretch}{1.25}
    \resizebox{\textwidth}{!}{%
    \begin{tabular}{@{}cl ccccc|cccccccc@{}}
    \toprule
    & & \multicolumn{5}{c}{Atomic metrics} & \multicolumn{8}{c}{Compound metrics} \\
    \cmidrule(lr){3-7}\cmidrule(lr){8-15}
    & \multirow{2}{*}{Epochs}
    & \multirow{2}{*}{\makecell{Response\\Diversity}} & \multirow{2}{*}{\makecell{Judge\\Score}} & \multirow{2}{*}{Deviation} & \multirow{2}{*}{\makecell{Vendi\\Score}} & \multirow{2}{*}{\makecell{Distinct\\Clusters}}
    & \multicolumn{2}{c}{Filtered} & \multicolumn{2}{c}{AUC-dev} & \multicolumn{2}{c}{AUC-qual} & \multicolumn{2}{c}{VUS} \\
    \cmidrule(lr){8-9}\cmidrule(lr){10-11}\cmidrule(lr){12-13}\cmidrule(lr){14-15}
    & & & & & & & Vendi & Distinct & Vendi & Dist. & Vendi & Dist. & Vendi & Dist. \\
    \midrule
    & 2  & \stat{13.4}{1.6} & \stat{92.0}{1.1} & \stat{5.1}{0.7} & \stat{2.27}{0.20} & \stat{19.3}{3.3} & \stat{0.46}{0.23} & \stat{1.20}{0.85} & 0.294 & 1.22 & 47.7 & 131.0 & 30.2 & 139.0 \\
    & 4  & \stat{\textbf{13.9}}{1.5} & \stat{94.0}{1.0} & \stat{7.8}{0.9} & \stat{\textbf{2.42}}{0.21} & \stat{\textbf{26.5}}{3.8} & \stat{\textbf{0.76}}{0.28} & \stat{2.35}{1.33} & 0.387 & 2.37 & \textbf{74.1} & 234.0 & \textbf{37.8} & 235.2 \\
    & 6  & \stat{13.4}{1.3} & \stat{94.7}{0.8} & \stat{9.1}{0.9} & \stat{2.37}{0.18} & \stat{25.0}{3.6} & \stat{0.65}{0.27} & \stat{\textbf{2.49}}{1.41} & \textbf{0.390} & \textbf{2.57} & 64.2 & \textbf{243.4} & \textbf{37.8} & \textbf{250.9} \\
    & 8  & \stat{12.2}{1.2} & \stat{\textbf{94.9}}{0.8} & \stat{9.1}{0.8} & \stat{2.23}{0.18} & \stat{22.1}{3.6} & \stat{0.56}{0.24} & \stat{1.95}{1.23} & 0.353 & 2.22 & 55.4 & 190.2 & 34.4 & 218.1 \\
    & 10 & \stat{11.7}{1.1} & \stat{93.6}{1.2} & \stat{\textbf{10.1}}{0.8} & \stat{2.17}{0.16} & \stat{20.5}{3.3} & \stat{0.57}{0.23} & \stat{1.76}{1.06} & 0.345 & 2.23 & 55.8 & 172.3 & 34.1 & 221.0 \\
    \midrule
    \multirow{5}{*}{\rotatebox[origin=c]{90}{\scriptsize{+ Deviation}}}
    & 2  & \stat{13.5}{1.6} & \stat{91.5}{1.4} & \stat{4.9}{0.7} & \stat{2.28}{0.21} & \stat{18.3}{3.2} & \stat{0.40}{0.24} & \stat{1.24}{0.95} & 0.278 & 1.08 & 45.3 & 137.4 & 29.3 & 126.3 \\
    & 4  & \stat{15.6}{1.7} & \stat{92.0}{1.2} & \stat{7.1}{0.9} & \stat{2.53}{0.22} & \stat{19.1}{3.2} & \stat{0.74}{0.28} & \stat{2.06}{1.17} & 0.386 & 1.59 & 83.4 & 232.4 & 40.0 & 175.0 \\
    & 6  & \stat{\textbf{19.7}}{1.3} & \stat{\textbf{93.1}}{1.1} & \stat{19.1}{1.2} & \stat{\textbf{3.28}}{0.22} & \stat{\textbf{29.9}}{3.5} & \stat{2.59}{0.31} & \stat{14.7}{2.70} & \textbf{0.851} & \textbf{5.58} & 259.5 & 1567.0 & 85.1 & 586.9 \\
    & 8  & \stat{18.1}{1.2} & \stat{91.7}{1.4} & \stat{20.8}{1.1} & \stat{3.02}{0.19} & \stat{24.7}{3.2} & \stat{2.45}{0.27} & \stat{13.3}{2.69} & 0.786 & 4.59 & 249.1 & 1495.9 & 79.9 & 512.0 \\
    & 10 & \stat{18.0}{1.1} & \stat{88.4}{1.8} & \stat{\textbf{24.7}}{1.0} & \stat{3.02}{0.18} & \stat{28.1}{3.5} & \stat{\textbf{2.60}}{0.24} & \stat{\textbf{16.4}}{2.80} & 0.835 & 5.24 & \textbf{272.8} & \textbf{2022.3} & \textbf{88.0} & \textbf{639.0} \\
    \bottomrule
    \end{tabular}%
    }
    \caption{\textbf{DARLING evaluated every 2 epochs on NoveltyBench}, without (top) and with (bottom) the deviation reward. The 10-epoch rows match the corresponding DARLING rows in Table~\ref{tab:eval-comparison}; the 2-epoch rows use the same data budget as DGRPO and \cp{}. Response diversity and deviation are shown $\times 100$; cells include a 95\% confidence interval where available. Best value per column within each block is bolded. DARLING's diversity peaks around epochs 4--6 and then erodes with further training, and no checkpoint --- at any epoch, with or without deviation --- approaches the system-creativity of \cp{}.}
    \label{tab:darling-epochs}
\end{table}

At the equal-data budget of 2 epochs, DARLING is no more creative than at its reported 10-epoch checkpoint; without the deviation reward its simple diversity peaks around epochs 4--6 and then declines as continued training trades diversity for reward-model quality. However, we do see deviation increasing throughout training. In all cases DARLING's compound creativity metrics remain far below those of \cp{} (Table~\ref{tab:eval-comparison}), so the epoch mismatch neither advantages nor explains away the gap: DARLING does not become competitive at any point along its training trajectory.

\clearpage
\section{Templates}
\label{app:templates}

This appendix collects the prompt templates used throughout this work. Figure~\ref{fig:priming_template} shows the template that elicits a creative primer from the primer model used in evaluation (1 of 32 used during training), Figure~\ref{fig:response_template} gives the template that conditions the response model on that primer, and Figure~\ref{fig:judge_template} gives the template used by the creative primer judge $J_{\mathrm{CP}}$ (Eq.~\ref{eq:judge}). Figures~\ref{fig:eval_judge_template}--\ref{fig:eval_judge_components_ex2} give the LLM-as-a-judge template used for response-quality evaluation: this judge scores every response on six fixed \emph{universal} components and on a set of \emph{prompt-specific} components that are generated once per prompt at annotation time. Figure~\ref{fig:eval_judge_template} shows the template skeleton with its three placeholders left unfilled, and Figures~\ref{fig:eval_judge_components_ex1} and~\ref{fig:eval_judge_components_ex2} show the prompt-specific components materialized for two example prompts from NoveltyBench.

\begin{tcolorbox}[
        breakable,
        colback=black!8, colframe=black!55, boxrule=0.5pt, arc=2pt,
        left=6pt, right=6pt, top=4pt, bottom=4pt,
        title=\textbf{Creative primer prompt template}, coltitle=black,
        colbacktitle=black!8, fonttitle=\normalsize,
    ]
\begin{lstlisting}[style=prompttemplate]
You are providing meta-level creative guidance, not solving the task.

Given the prompt below, produce a few sentences of advice that would help another language model respond with more creative CONTENT. Focus on what substantive ideas, perspectives, or angles the response should explore — not on style, tone, or formatting.
Your advice MUST be specific to this particular prompt - it should not be generic advice that could be applied to any prompt.

Do not answer the prompt or include example content.

Prompt:
{original_prompt}
\end{lstlisting}
\end{tcolorbox}
\captionof{figure}{Prompt template used to elicit a creative primer from the primer model. At inference \texttt{\{original\_prompt\}} is substituted with the original prompt.}
\label{fig:priming_template}

\begin{tcolorbox}[
        breakable,
        colback=black!8, colframe=black!55, boxrule=0.5pt, arc=2pt,
        left=6pt, right=6pt, top=4pt, bottom=4pt,
        title=\textbf{Response prompt template}, coltitle=black,
        colbacktitle=black!8, fonttitle=\normalsize,
    ]
\begin{lstlisting}[style=prompttemplate]
Below you have a prompt and some guidance on how to respond. Your highest priority is to answer the prompt, and if the guidance conflicts, prioritize the prompt. Do not acknowledge the guidance in your response.

Prompt:
{original_prompt}

Guidance:
{creative_prep}
\end{lstlisting}
\end{tcolorbox}
\captionof{figure}{Prompt template used to condition the response model on a creative primer. At inference \texttt{\{original\_prompt\}} and \texttt{\{creative\_primer\}} are substituted with the original prompt and the generated primer.}
\label{fig:response_template}

\clearpage
\begin{tcolorbox}[
        breakable,
        colback=black!8, colframe=black!55, boxrule=0.5pt, arc=2pt,
        left=6pt, right=6pt, top=4pt, bottom=4pt,
        title=\textbf{Creative primer judge template}, coltitle=black,
        colbacktitle=black!8, fonttitle=\normalsize,
    ]
\begin{lstlisting}[style=prompttemplate]
You are evaluating the quality of a creative preparation strategy for a language model. Creative preparations are instructions given to a model before it sees the actual user prompt. Their purpose is to shape how the model *thinks* about the problem, leading to more diverse and higher-quality responses.

The best creative preps are thinking strategies — they tell the model what to consider, question, or reason about before responding. They guide the model's internal reasoning process rather than dictating what the output should look like. A creative prep that says "before responding, identify the assumption most people make about this topic and consider what happens if it's wrong" is far more valuable than one that says "write your response as a poem" or "use vivid metaphors."

Important: a creative prep must BE a creative prep — not instructions for how to write one. If the text reads as meta-guidance about crafting advice (e.g., "craft advice that focuses on a specific lens" or "prescribe a narrow limitation that forces originality"), it has failed at its basic purpose regardless of how sophisticated it sounds.

Evaluate the creative prep on the following criteria.

---

# EVALUATION CRITERIA

## 1. Specificity (0-100)
**Definition**: Is this tailored to the unique characteristics of THIS SPECIFIC PROMPT, or generic advice?

**Scoring Rubric**:
- **0-20**: Completely generic, ignores prompt content, could apply to any prompt (e.g., "think carefully", "be creative", "subvert expectations", "use a specific lens to force originality")
- **21-40**: Recognizes broad category but ignores specific features (e.g., "this is a story prompt", "this is a comparison question")
- **41-60**: Mentions some prompt characteristics but stays surface-level (e.g., "this has constraints", "this asks for analysis")
- **61-80**: Identifies specific prompt features and references them concretely (e.g., "this prompt requires structural repetition", "this comparison involves temporal change")
- **81-100**: Deeply analyzes the prompt's unique characteristics and frames approach around them (e.g., "this repetition constraint can become a semantic device", "what's unsaid in this comparison is the selection criteria itself")

## 2. Thinking Strategy (0-100)
**Definition**: Does this provide a thinking strategy that shapes the model's reasoning process, or does it prescribe specific content, structure, or output?

A good creative prep tells the model how to *think* about the problem — what to question, what angle to take, what to reason about before responding. It leaves the actual content open. Low scores go to preps that dictate specific elements, structures, content, or output format. The highest scores go to preps that explicitly guide the model's pre-response reasoning (e.g., "before responding, think about...", "first consider...", "ask yourself...").

**Scoring Rubric**:
- **0-20**: Dictates exact content or structure (e.g., "start with X, then say Y, conclude with Z", "your response must include these three elements", "write exactly 5 paragraphs about A, B, C, D, E")
- **21-40**: Prescribes specific elements with little room for interpretation (e.g., "make sure to mention X, Y, and Z", "use this specific example to illustrate your point")
- **41-60**: Mix of strategy and prescription — gives a thinking approach but also constrains specific outputs (e.g., "consider different angles, and make sure to include a counterargument")
- **61-80**: Clear thinking strategy with actionable direction (e.g., "first identify the constraint, then find ways to turn it into an advantage", "trace what changes between the two states — that's where the mechanism lives")
- **81-100**: Explicitly guides pre-response reasoning without constraining output (e.g., "before responding, list the most common answers people give, then ask what they all have in common — that shared assumption is your entry point", "ask yourself: what would change about this question if you shifted the time horizon by 100 years?")

## 3. Content Focus (0-100)
**Definition**: Does this guide *what the model should think about* — the substance, reasoning, or approach to the problem — or does it only address *how to present the output*?

Good creative preps direct the model's attention to ideas, arguments, angles, reasoning approaches, or aspects of the subject matter to explore. Weak creative preps only specify output format, tone, narrative framing, or stylistic choices. Guiding the model's *thinking approach* (e.g., "consider what assumptions underlie this question") counts as content guidance, since it shapes what substance the model engages with. Pure presentation instructions (e.g., "frame as a metaphor", "embed the answer in a folktale", "write as a dialogue") are style, not content — even when described with rich, specific detail (e.g., "embed it in a Zulu proverb" is still a presentation instruction). However, a light style suggestion alongside real thinking guidance shouldn't tank the score — what matters is the balance. If the original prompt is about style or form (e.g., "write a poem"), then style guidance IS content guidance.

**Scoring Rubric**:
- **0-20**: Purely about output presentation with no guidance on what to think about (e.g., "write in a whimsical tone", "use lots of metaphors", "embed the answer in a traditional folktale, proverb, or myth")
- **21-40**: Mostly presentation-focused with vague content hints (e.g., "write engagingly and try to cover multiple perspectives", "use an unexpected narrative frame and touch on different angles")
- **41-60**: Mix of thinking guidance and presentation — has some approach direction but presentation instructions dominate or dilute it (e.g., "frame using an unconventional metaphor that diverges from typical associations" — there's an implicit thinking nudge but it's wrapped in a style directive)
- **61-80**: Primarily guides what to think about, with incidental style elements (e.g., "the key insight is that X enables Y — explore the implications" with minor style notes)
- **81-100**: Entirely focused on what the model should reason about — ideas, angles, assumptions to question, knowledge to bring to bear (e.g., "the constraint creates a tension between efficiency and readability — explore which trade-offs matter most and why")

## 4. Prompt Faithfulness (0-100)
**Definition**: Does the creative prep work within the bounds of what the prompt is asking, or does it transform the question into something else?

A good creative prep helps the model produce diverse, high-quality responses *to the actual question*. A bad one distorts the prompt's intent — encouraging fiction for a factual question, turning a simple request into an elaborate creative exercise, narrowing a general question to a specific ideological or niche framing that most people wouldn't intend, or suggesting the response should be a "story" or "artifact" when the prompt asks for information. For open-ended or creative prompts, imaginative reinterpretation is fine. For factual or constrained prompts, the prep should respect those boundaries.

**Scoring Rubric**:
- **0-20**: Actively works against the prompt's intent — encourages fiction for factual questions, ignores explicit constraints, or transforms the question into something entirely different
- **21-40**: Substantially distorts the prompt — forces a specific ideological lens, adds major requirements that change what's being asked, or reframes a simple question as an elaborate creative exercise (e.g., suggesting a factual answer should "feel like a self-contained story or artifact")
- **41-60**: Partially respects the prompt but pushes it in a direction that doesn't match the user's likely intent — e.g., narrowing a general question to one specific angle, or adding a framing that subtly changes what's being asked
- **61-80**: Mostly respects the prompt's intent, with minor stretches that don't fundamentally change what's being asked
- **81-100**: Fully respects the prompt's intent — helps the model think creatively within the bounds of what was actually asked

## 5. Focus (0-100)
**Definition**: Does the creative prep commit to a coherent thinking strategy, or does it pile on unrelated ideas hoping something sticks?

A focused creative prep provides depth — it develops one thinking approach and lets that approach do the work. An unfocused one substitutes breadth for depth, whether by listing many loosely connected thinking strategies, dumping unrelated constraints or formatting rules, or exploring multiple reframings of the same question without committing to one. Even if each individual idea is good, cramming many together produces an unfocused prep.

**Scoring Rubric**:
- **0-20**: Kitchen sink — lists many unrelated strategies, constraints, or rules with no unifying thread (e.g., "try analogy, consider edge cases, use humor, think historically, subvert expectations" or "use lowercase, no markdown, prioritize verbs, avoid cliches, keep it cryptic")
- **21-40**: Scattered — several loosely related ideas competing for attention, no clear through-line connecting them (e.g., exploring paradoxical lenses AND emotional dissonance AND unspoken hierarchies AND meta-narratives AND philosophical inquiry all in one prep)
- **41-60**: Partially focused — has a main thinking strategy but cluttered with tangential ideas that dilute rather than reinforce it
- **61-80**: Focused — a clear central thinking strategy, possibly with one or two supporting ideas that build on or reinforce the core approach
- **81-100**: Sharp and cohesive — commits to a single thinking approach and develops it with depth, or uses a few tightly interlocking ideas that clearly serve the same angle

## 6. Direct Address (binary: 0 or 1)
**Definition**: Does the creative prep refer to the model as a third-party entity?

- **0**: The prep refers to the model as a third-party entity (e.g., "the model should...", "encourage the model to...", "AI systems should...", "guide the model to...")
- **1**: The prep does NOT refer to the model as a third-party entity. It may use direct address ("you"), bare imperatives ("start by..."), or discuss the task/response ("the response should...") — all of these are fine.

---

# THE ORIGINAL PROMPT

{prompt}

---

# THE CREATIVE PREP TO EVALUATE

{creative_prep}

---

# EVALUATION INSTRUCTIONS

## Step 1: Reasoning

Work through each criterion one at a time BEFORE assigning any scores. For each criterion:
1. Identify specific phrases or patterns in the creative prep that relate to this criterion
2. Consider where these elements fall within the scoring rubric
3. Note any tensions or edge cases

Write your reasoning in plain text below. Be thorough but focused — reference specific parts of the creative prep to support your analysis.

## Step 2: Final Scores

After completing your reasoning, provide your scores in the following JSON format. The justifications should be brief summaries (1-2 sentences) of your reasoning above. Do not include any text after the JSON block.

```json
{
    "specificity": {
        "score": <0-100>,
        "justification": "<brief explanation>"
    },
    "thinking_strategy": {
        "score": <0-100>,
        "justification": "<brief explanation>"
    },
    "content_focus": {
        "score": <0-100>,
        "justification": "<brief explanation>"
    },
    "prompt_faithfulness": {
        "score": <0-100>,
        "justification": "<brief explanation>"
    },
    "focus": {
        "score": <0-100>,
        "justification": "<brief explanation>"
    },
    "direct_address": {
        "score": <0 or 1>,
        "justification": "<brief explanation>"
    }
}
```

Remember:
- For specificity, thinking_strategy, content_focus, prompt_faithfulness, and focus: higher scores (0-100) are better
- For direct_address: 1 = addresses model directly or discusses the task, 0 = talks about the model as a third-party entity ("the model should...", "encourage the model to...")
- Your justifications should reference specific phrases from the creative prep
- Think step-by-step before scoring
\end{lstlisting}
\end{tcolorbox}
\captionof{figure}{Prompt template for the criterion-mean creative primer judge $J_{\mathrm{CP}}$ (Section~\ref{sec:cp_reward_components}, Eq.~\ref{eq:judge}). At inference \texttt{\{prompt\}} and \texttt{\{creative\_primer\}} are substituted with the original prompt and the primer under evaluation.}
\label{fig:judge_template}

\clearpage
\begin{tcolorbox}[
        breakable,
        colback=black!8, colframe=black!55, boxrule=0.5pt, arc=2pt,
        left=6pt, right=6pt, top=4pt, bottom=4pt,
        title=\textbf{Evaluation LLM-as-a-judge template}, coltitle=black,
        colbacktitle=black!8, fonttitle=\normalsize,
    ]
\begin{lstlisting}[style=prompttemplate]
You are an expert evaluator assessing the quality of an AI response. You will evaluate the response on both UNIVERSAL components (applicable to all responses) and PROMPT-SPECIFIC components (tailored to this particular prompt type).

## PROMPT
{{PROMPT}}

## RESPONSE TO EVALUATE
{{RESPONSE}}

---

# PART 1: UNIVERSAL COMPONENTS

Evaluate each of the following components on a scale of 0-100.

### 1. Instruction Following
How well does the response address what was actually asked?
- 0-20: Ignores or fundamentally misunderstands the prompt
- 21-40: Addresses the prompt partially but misses key requirements
- 41-60: Addresses the main request but overlooks minor specifications
- 61-80: Follows instructions well with only trivial deviations
- 81-100: Precisely and completely follows all instructions

### 2. Coherence
Is the response logically structured and easy to follow?
- 0-20: Disorganized, contradictory, or incomprehensible
- 21-40: Poor structure with unclear logical flow
- 41-60: Adequate structure but some confusing transitions or organization
- 61-80: Well-organized with clear logical progression
- 81-100: Exceptionally well-structured with seamless logical flow

### 3. Completeness
Does it fully address the prompt without obvious gaps?
- 0-20: Severely incomplete; misses most aspects of the prompt
- 21-40: Addresses some aspects but leaves significant gaps
- 41-60: Covers main points but lacks depth or misses secondary aspects
- 61-80: Thorough coverage with only minor omissions
- 81-100: Comprehensively addresses all aspects of the prompt

### 4. Conciseness
Is the response appropriately sized—not padded, not truncated?
- 0-20: Extremely verbose/padded OR severely truncated
- 21-40: Noticeably too long or too short for the task
- 41-60: Somewhat inefficient length but acceptable
- 61-80: Appropriate length with minor verbosity or brevity issues
- 81-100: Optimally sized for the task at hand

### 5. Clarity
Is the language clear and unambiguous?
- 0-20: Confusing, jargon-heavy, or ambiguous throughout
- 21-40: Frequently unclear or poorly worded
- 41-60: Generally understandable but some unclear passages
- 61-80: Clear language with only occasional ambiguity
- 81-100: Exceptionally clear and precise language throughout

### 6. Surprise
Does the response offer something unexpected while still fulfilling the original request? Does it challenge implicit assumptions, introduce novel angles, or deliver value beyond the obvious interpretation—without departing from what was asked?
- 0-20: Entirely predictable; the most generic possible response to this prompt
- 21-40: Mostly boilerplate with perhaps one minor unexpected element
- 41-60: Competent but unsurprising; follows the expected path without deviation
- 61-80: Contains notable unexpected elements—a fresh angle, an unconventional example, or a thought-provoking reframe—while still satisfying the request
- 81-100: Genuinely surprising; reframes the problem, surfaces hidden assumptions, or delivers insight the prompter didn't know they wanted—yet still fully answers the prompt

---

# PART 2: PROMPT-SPECIFIC COMPONENTS

The following components are tailored to this specific type of prompt. Evaluate each on a scale of 0-100.

{{PROMPT_SPECIFIC_COMPONENTS}}

---

# EVALUATION INSTRUCTIONS

First, reason through your evaluation step by step BEFORE assigning any scores. For each component (both universal and prompt-specific), discuss what the response does well and where it falls short relative to the rubric. Reference specific parts of the response to support your reasoning. After you have reasoned through ALL components, provide your final scores in the JSON format below.

## Step 1: Reasoning

Work through each component one at a time. For each, consider:
- What specific aspects of the response are relevant to this component?
- Where does the response fall within the scoring rubric?
- Are there any edge cases or tensions between quality and convention?

Write your reasoning in plain text. Be thorough but focused.

## Step 2: Final Scores

After your reasoning is complete, provide your scores in the following JSON format. The justifications should be brief summaries of your reasoning above. Do not include any text after the JSON block.

```json
{
  "universal": {
    "instruction_following": { "score": <0-100>, "justification": "<brief summary>" },
    "coherence":             { "score": <0-100>, "justification": "<brief summary>" },
    "completeness":          { "score": <0-100>, "justification": "<brief summary>" },
    "conciseness":           { "score": <0-100>, "justification": "<brief summary>" },
    "clarity":               { "score": <0-100>, "justification": "<brief summary>" },
    "surprise":              { "score": <0-100>, "justification": "<brief summary>" }
  },
  "prompt_specific": {
    "components": [
      { "name": "<component name>", "score": <0-100>, "justification": "<brief summary>" }
    ],
    "aggregate": { "score": <0-100>, "method": "<e.g., simple average, weighted average>", "justification": "<overall prompt-specific summary>" }
  }
}
```
\end{lstlisting}
\end{tcolorbox}
\captionof{figure}{LLM-as-a-judge template used for response-quality evaluation, shown with its three placeholders unfilled. \texttt{\{\{PROMPT\}\}} and \texttt{\{\{PROMPT\_SPECIFIC\_COMPONENTS\}\}} are substituted once per prompt at annotation time (see Figures~\ref{fig:eval_judge_components_ex1} and~\ref{fig:eval_judge_components_ex2}); \texttt{\{\{RESPONSE\}\}} is substituted at scoring time with each response under evaluation. The six universal components are fixed across all prompts.}
\label{fig:eval_judge_template}

\begin{tcolorbox}[
        breakable,
        colback=black!8, colframe=black!55, boxrule=0.5pt, arc=2pt,
        left=6pt, right=6pt, top=4pt, bottom=4pt,
        title=\textbf{Evaluation LLM-as-a-judge: prompt-specific components (example 1)}, coltitle=black,
        colbacktitle=black!8, fonttitle=\normalsize,
    ]
\begin{lstlisting}[style=prompttemplate]
### 7. Emotional Resonance
Measures how effectively the story evokes genuine feeling about the girl-dog relationship through specific moments, gestures, or details rather than generic sentiment.
- 0-20: Relies entirely on clichéd phrases ('best friends forever') or tells us about emotions without showing them
- 21-40: Includes some emotional moments but they feel predictable or borrowed from familiar stories
- 41-60: Creates at least one authentic emotional beat through a specific detail or interaction
- 61-80: Weaves multiple emotionally specific moments that feel fresh and genuine to this particular relationship
- 81-100: Achieves profound emotional truth through unexpected yet perfectly chosen details that illuminate the unique bond

### 8. Narrative Economy
Evaluates how much story complexity, character depth, or thematic richness the writer achieves within the five-sentence constraint.
- 0-20: Linear series of events with no depth, essentially 'this happened, then this happened'
- 21-40: Basic story arc present but characters remain flat and stakes feel minimal
- 41-60: Manages to suggest character growth, conflict, or a meaningful arc despite brevity
- 61-80: Impressively layers multiple story elements (subplot, theme, character development) into the constraint
- 81-100: Achieves the density of a much longer piece, with each sentence doing multiple jobs to create a rich, complete world
\end{lstlisting}
\end{tcolorbox}
\captionof{figure}{Values substituted into the placeholders of Figure~\ref{fig:eval_judge_template} for the NoveltyBench prompt ``Tell me a story in five sentences about a girl and her dog.''}
\label{fig:eval_judge_components_ex1}

\begin{tcolorbox}[
        breakable,
        colback=black!8, colframe=black!55, boxrule=0.5pt, arc=2pt,
        left=6pt, right=6pt, top=4pt, bottom=4pt,
        title=\textbf{Evaluation LLM-as-a-judge: prompt-specific components (example 2)}, coltitle=black,
        colbacktitle=black!8, fonttitle=\normalsize,
    ]
\begin{lstlisting}[style=prompttemplate]
{{PROMPT}} =
    Imagine you are a time traveler who has just arrived in a new era.
    Describe the first thing you notice.

{{PROMPT_SPECIFIC_COMPONENTS}} =

### 7. Sensory Immersion and Detail
Measures how effectively the response uses specific sensory details and observations to create a vivid, immersive experience of temporal displacement.
- 0-20: Generic observations that could apply to any time period; relies on clichés like 'the air smells different' without specificity
- 21-40: Some attempt at sensory detail but remains surface-level; mentions obvious differences like 'no cars' or 'people dress differently'
- 41-60: Includes specific sensory details that ground the reader in the era; notices concrete differences in sounds, smells, or textures
- 61-80: Rich sensory tapestry that captures subtle, era-specific details a time traveler would genuinely notice first; conveys visceral disorientation
- 81-100: Masterfully uses unexpected sensory details to reveal the era through fresh observations; makes the familiar strange or the strange familiar in compelling ways

### 8. Temporal Perspective and Insight
Evaluates how the response leverages the unique perspective of a time traveler to reveal something profound or thought-provoking about the era or human experience.
- 0-20: No meaningful use of the time traveler perspective; could be any observer describing any place
- 21-40: Basic fish-out-of-water observations; focuses on obvious technological or social differences without deeper insight
- 41-60: Uses time traveler perspective to highlight interesting contrasts or continuities across eras; shows awareness of historical context
- 61-80: Reveals unexpected truths about the era through the traveler's unique vantage point; finds poetry in mundane details or profound meaning in small observations
- 81-100: Brilliantly exploits the time travel premise to offer fresh philosophical, emotional, or anthropological insights; makes us see our own time differently
\end{lstlisting}
\end{tcolorbox}
\captionof{figure}{Values substituted into the placeholder of Figure~\ref{fig:eval_judge_template} for the NoveltyBench prompt ``Imagine you are a time traveler who has just arrived in a new era. Describe the first thing you notice.''}
\label{fig:eval_judge_components_ex2}

\end{document}